\documentclass[11pt]{article}

\usepackage{epsfig,epsf,fancybox}
\usepackage{amsmath}
\usepackage{mathrsfs}
\usepackage{amssymb}
\usepackage{graphicx}
\usepackage{color}
\usepackage{multirow}
\usepackage{paralist}
\usepackage{verbatim}
\usepackage{galois}
\usepackage{algorithm}
\usepackage{algorithmic}
\usepackage{boxedminipage}
\usepackage{booktabs}
\usepackage{accents}
\usepackage{stmaryrd}

\usepackage{subfig}

\usepackage{natbib}

\usepackage{url}
\usepackage[colorlinks,linkcolor=magenta,citecolor=blue, pagebackref=true,backref=true]{hyperref}
\renewcommand*{\backrefalt}[4]{%
    \ifcase #1 \footnotesize{(Not cited.)}%
    \or        \footnotesize{(Cited on page~#2.)}%
    \else      \footnotesize{(Cited on pages~#2.)}%
    \fi}

\newtheorem{theorem}{Theorem}[section]
\newtheorem{corollary}[theorem]{Corollary}
\newtheorem{lemma}[theorem]{Lemma}
\newtheorem{proposition}[theorem]{Proposition}

\newtheorem{definition}{Definition}[section]

\newtheorem{remark}[theorem]{Remark}
\newtheorem{assumption}[theorem]{Assumption}
\numberwithin{equation}{section}

\newcommand{\BB}{\mathbb{B}}
\newcommand{\EE}{\mathbb{E}}

\newcommand{\st}{\textnormal{s.t.}}

\newcommand{\sign}{\textnormal{sign}}

\newcommand{\x}{\mathbf x}

\newcommand{\s}{\mathbf s}

\newcommand{\g}{\mathbf g}

\newcommand{\argmin}{\mathop{\rm argmin}}

\newcommand{\PCal}{\mathcal{P}}

\newcommand{\SCal}{\mathcal{S}}

\newcommand{\YCal}{\mathcal{Y}}

\newcommand{\JCal}{\mathcal{J}}

\newcommand{\br}{\mathbb{R}}

\newcommand{\bn}{\mathbb{N}}
\newcommand{\ba}{\begin{array}}
\newcommand{\ea}{\end{array}}

\newcommand{\ZCal}{\mathcal{Z}}

\newcommand{\NCal}{\mathcal{N}}

\newcommand{\mydefn}{:=}

\begin{document}


\begin{center}

{\bf{\LARGE{On Structured Filtering-Clustering: Global Error Bound \\ [.2cm] and Optimal First-Order Algorithms}}}

\vspace*{.2in}
{\large{ \begin{tabular}{c}
Nhat Ho$^{\star, \ddagger}$ \and Tianyi Lin$^{\star, \diamond}$ \and Michael I. Jordan$^{\diamond, \dagger}$ \\
\end{tabular}
}}

\vspace*{.2in}

\begin{tabular}{c}
Department of Electrical Engineering and Computer Sciences$^\diamond$ \\
Department of Statistics$^\dagger$ \\
University of California, Berkeley \\
Department of Statistics and Data Sciences, University of Texas, Austin$^\ddagger$ \\ 
\end{tabular}

\vspace*{.2in}

\today

\vspace*{.2in}

\begin{abstract}
The filtering-clustering models, including trend filtering and convex clustering, have become an important source of ideas and modeling tools in machine learning and related fields. The statistical guarantee of optimal solutions in these models has been extensively studied yet the investigations on the computational aspect have remained limited. In particular, practitioners often employ the first-order algorithms in real-world applications and are impressed by their superior performance regardless of ill-conditioned structures of difference operator matrices, thus leaving open the problem of understanding the convergence property of first-order algorithms. This paper settles this open problem and contributes to the broad interplay between statistics and optimization by identifying a \textit{global error bound} condition, which is satisfied by a large class of dual filtering-clustering problems, and designing a class of \textit{generalized dual gradient ascent} algorithm, which is \textit{optimal} first-order algorithms in deterministic, finite-sum and online settings. Our results are new and help explain why the filtering-clustering models can be efficiently solved by first-order algorithms. We also provide the detailed convergence rate analysis for the proposed algorithms in different settings, shedding light on their potential to solve the filtering-clustering models efficiently. We also conduct experiments on real datasets and the numerical results demonstrate the effectiveness of our algorithms.
\end{abstract}

\let\thefootnote\relax\footnotetext{$^\star$ Nhat Ho and Tianyi Lin contributed equally to this work.}
\end{center}

\section{Introduction}
We are interested in the filtering-clustering models which are given by 
\begin{equation}\label{prob:main}
\min\limits_{\beta \in \br^d} \ \Phi(\beta) \mydefn f(\beta) + \lambda\left(\sum_{k=1}^n \|D_k \beta\|_p\right), 
\end{equation}
where $f: \br^d \mapsto \br$ is a strongly convex loss function and has Lipschitz continuous gradient, $D_k \in \br^{m \times d}$ are called \textit{discrete difference operator} matrices for $1 \leq k \leq n$, and $\lambda>0$ is a regularization parameter and $p \in \bn^+$ is a regularization index. Common applications of the filtering-clustering models in Eq.~\eqref{prob:main}, such as $\ell_1$-trend filtering and $\ell_2$-convex clustering, are associated with squared Euclidean loss function $f$ and poorly conditioned matrices $D_k$. Such ill-conditioning poses several challenges for first-order algorithms to achieve fast and stable convergence.

The filtering-clustering models, including trend filtering and convex clustering, cover a wide range of application problems arising from machine learning and statistics. Examples of trend filtering applications include nonparametric regression~\citep{Kim-2009-Ell1, Ryan-2014-Adaptive, Lin-2017-Sharp, Padilla-2018-adaptive, Guntuboyina-2020-Adaptive}, adaptive estimators in graphs~\citep{Wang-2016-Trend, Padilla-2018-DFS}, and time series analysis~\citep{Leser-1961-Simple}. Convex clustering has been proposed as an alternative to traditional clustering approaches,  e.g., $K$-means clustering and hierarchical clustering, and was well known for its appealing robustness and stability properties~\citep{Hocking-2011-Clusteringpath, Zhu-2014-Convex, Tan-2015-Statistical, Wu-2017-New, Radchenko-2017-Convex}.

Recent years have witnessed much progress on the statistical aspects of filtering-clustering models. Indeed, the global solutions of these models have been shown to enjoy the desirable properties~\citep{Ryan-2014-Adaptive, Zhu-2014-Convex, Tan-2015-Statistical, Wang-2016-Trend, Wu-2017-New, Radchenko-2017-Convex, Padilla-2018-DFS, Guntuboyina-2020-Adaptive}. However, investigations on the computational theory for the filtering-clustering models is relatively scarce. Indeed, there have been a flurry of optimization algorithms that were proposed for solving the filtering-clustering models in the literature. For example, the trend filtering algorithms include primal-dual interior-point method (PDIP)~\citep{Kim-2009-Ell1}, alternating direction method of multipliers (ADMM)~\citep{Ramdas-2016-Fast} and specialized Newton's method~\citep{Wang-2016-Trend}. The convex clustering algorithms include ADMM, alternating minimization algorithm (AMA)~\citep{Eric-2015-Splitting}, projected dual gradient ascent~\citep{Wang-2016-Robust} and semismooth Newton's method~\citep{Sun-2021-Convex}. Most of these existing approaches were built on first-order algorithmic frameworks and have been recognized as the benchmark in the literature due to their simplicity and ease-of-the-implementation. Practitioners employ them in real-world applications and often observe superior performance~\citep{Eric-2015-Splitting, Ramdas-2016-Fast, Wang-2016-Robust} regardless of ill-conditioned structures of difference operator matrices. This is somehow surprising since the first-order algorithms are known to suffer from the slow sublinear convergence rate for general convex problems, thus leaving open the problem of understanding the convergence property of the first-order algorithms as applied to the filtering-clustering models. To our knowledge, there is currently a paucity of computational theory that can uncover the mysterious success of these first-order algorithms and further lead to better algorithms.

\paragraph{Contributions.} In this paper, we study the computational theory concerning the filtering-clustering models. Our contributions are summarized as follows:
\begin{enumerate}
\item We analyze the structure of the filtering-clustering model in Eq.~\eqref{prob:main} and prove that a \textit{global} error bound condition holds for the dual filtering-clustering model when $p = 1$ or $p \in [2, +\infty]$. Our results are nontrivial: first of all, the filtering-clustering model is not amenable to the existing techniques developed in~\citet{Wang-2014-Iteration} since the non-smooth term in the objective function of the dual form does not a polyhedral epigraph when $p \in [2, +\infty)$. Second, the filtering-clustering model in~\eqref{prob:main} can not be formulated as an $\ell_{1, p}$-regularized problem for some $p \geq 1$. As such, our results are not a straightforward consequence of~\citet{Zhou-2015-Error}. 
\item We propose a class of first-order primal-dual optimization algorithms for solving the filtering-clustering model with an optimal linear convergence rate. There are two fundamental reasons for the non-triviality of the result: (i) The objective function of the filtering-clustering model is non-strongly convex since $D_k^\top$ are not full column rank; (ii) the gradient of the objective function of the dual filtering-clustering model is inaccessible.
\item In addition to deterministic counterparts, we propose and analyze a class of stochastic first-order primal-dual optimization algorithms for solving the filtering-clustering model in Eq.~\eqref{prob:main}. In the finite-sum setting, the proposed stochastic variance reduced first-order algorithms attain an optimal linear convergence rate. In the online setting, we conduct a similar analysis and show that our algorithm achieve the optimal rate up to log factors. Notably, our algorithms are build on celebrated gradient-based algorithms from~\citet{Allen-2017-Katyusha} and~\citet{Rakhlin-2012-Making}.
\end{enumerate}
\paragraph{Notation.} We denote $[n]$ as to the set $\{1, 2, \ldots, n\}$. For $p \in [1, +\infty]$, the notion $\|\cdot\|_p$ denotes $\ell_p$-norm and $\|\cdot\|$ denotes the Euclidean norm for a vector and the operator norm for a matrix. For all $q \geq 1$, $\BB_q = \{\alpha \in \br^m \mid \|\alpha\|_q \leq 1\}$ refers to a $\ell_q$-norm unit ball in $\mathbb{R}^m$ and $\bar{D}_q = \max_{\x, \x' \in \BB_q} \|\x - \x'\|$ refers to a diameter of $\ell_q$-norm unit ball in $\ell_2$-norm. For simplicity, we let $\BB_q^n$ denote the product of $n$ unit balls in $\ell_q$-norm. For a convex function $f$, $\partial f$ refers to the subdifferential of $f$. If $f$ is differentiable, $\partial f = \{\nabla f\}$ where $\nabla f$ is the gradient vector of $f$. For any closed set $\SCal$, we let $d(\x, \SCal)$ denote the distance between $\x$ and $\SCal$. If $\SCal$ is convex, we let $\NCal_\SCal(\x)$ denote the normal cone to $\SCal$ at $\x$. Lastly, given a tolerance $\varepsilon \in (0, 1)$, the notation $n = \Omega(b(\varepsilon))$ and $n = O(b(\varepsilon))$ stand for lower bound $n \geq c_1 \cdot b(\varepsilon)$ and upper bound $n \leq c_2 \cdot b(\varepsilon)$ where $c_1, c_2 > 0$ are independent of $1/\varepsilon$. In addition, $n = \Theta(b(\varepsilon))$ means that upper and lower bounds hold true. 

\section{Related Work}
Regarding the computation theory for deterministic first-order optimization algorithms, the best possible that we can expect is linear rate of convergence~\citep{Nesterov-2018-Lectures}. Such convergence result is commonly established under some additional assumptions on problem structure; e.g., strong convexity, global and local error bound~\citep{Pang-1987-Posteriori, Pang-1997-Error, Zhou-2017-Unified, Drusvyatskiy-2018-Error} and restricted strongly convexity~\citep{Negahban-2012-Unified, Negahban-2012-Restricted}. The former two assumptions are standard in the optimization literature and the error bound condition can be interpreted as a relaxation of strong convexity. Local error bound provides a guarantee for the \textit{asymptotic} linear convergence of feasible descent algorithms~\citep{Luo-1992-Linear, Luo-1993-Error, Tseng-2010-Approximation}, conditional gradient method~\citep{Beck-2017-Linearly}, ADMM~\citep{Hong-2017-Linear}, and proximal gradient algorithms~\citep{Drusvyatskiy-2018-Error}. Many application problems were shown to satisfy error bound conditions. For example,~\citet{Zhou-2015-Error} proved local error bound condition for $\ell_{1,p}$-norm regularized problems and~\citet{Wang-2014-Iteration} derived a clean form of global error bound for a class of nonstrongly convex problems. In addition,~\citet{Karimi-2016-Linear} established the linear convergence of proximal gradient algorithms in various non-strongly convex settings under the Polyak-Lojasiewicz condition (another relaxation of strong convexity). Very recently,~\citet{Jane-2021-Variational} have introduced a new framework based on the error bound, which enables us to provide some sufficient conditions for linear convergence and applicable approaches for calculating linear convergence rates of these first-order algorithms for a class of structured convex problems. On the other hand, the restricted strongly convexity played an important role in statistical learning literature and forms the basis for deriving optimal rate of first-order algorithms as applied to high-dimensional statistical recovery problems with sparsity-induced regularization~\citep{Agarwal-2012-Fast, Wang-2014-Optimal, Loh-2015-Regularized}. However, the aforementioned works do not contain (or imply) \textit{any global error bound analysis} for the filtering-clustering models in Eq.~\eqref{prob:main} due to the regularization term $\sum_{k=1}^n \|D_k \beta\|_p$ and thus can not explain the success of first-order algorithms, e.g., projected dual gradient ascent and ADMM, for solving the filtering-clustering models. 

Another line of relevant work focuses on first-order primal-dual optimization algorithms for convex-concave saddle-point problems; see, e.g.,~\citet{Chen-1997-Convergence, Palaniappan-2016-Stochastic, Wang-2017-Exploiting, Zhang-2017-Stochastic} and the references therein. More specifically, some works have derived the linear convergence results if the model is associated with either a strongly convex-concave structure~\citep{Chen-1997-Convergence, Palaniappan-2016-Stochastic}, together with efficient computed proximal mappings for nonsmooth terms. Nevertheless, these assumptions are not satisfied by the filtering-clustering models in Eq.~\eqref{prob:main}. An alternative way to establish the linear convergence of first-order primal-dual optimization algorithms is based on the construction of a potential function which decreases at a linear rate~\citep{Wang-2017-Exploiting, Zhang-2017-Stochastic}; however, it remains open how to construct such function for the filtering-clustering models. In addition, our algorithm is an inexact first-order primal-dual algorithm and thus resembles a generic inexact proximal algorithms~\citep{Schmidt-2011-Convergence}. However, their analysis is conducted under strong convexity and can not be extended to the filtering-clustering models. 

\section{Preliminaries}\label{sec:prelim}
We flesh out the basic filtering-clustering model in Eq.~\eqref{prob:main} and then turn to specific examples of this model. We also discuss primal and dual forms of the filtering-clustering model and demonstrate that its structure is suitable for the development of first-order primal-dual optimization algorithms.
  
\subsection{Filtering-clustering model}
The goal of this paper is to find a way to efficiently compute an optimal solution of the filtering-clustering model in Eq.~\eqref{prob:main}. Formally, we have
\begin{definition}\label{def:opt-sol}
A point $\beta^\star \in \br^d$ is an optimal solution of the filtering-clustering model in Eq.~\eqref{prob:main} if $\Phi(\beta^\star) \leq \Phi(\beta)$ for all $\beta \in \br^d$. 
\end{definition}
Since the convergence of first-order optimization algorithms to an optimal solution will depend on the gradient of the loss function $f$ in its neighborhood, it is necessary to impose Lipschtiz continuity conditions on the gradient $\nabla f$. We also impose the strong convexity on $f$ and remark that this assumption is standard in the filtering-clustering model arising from real-world application problems~\citep{Kim-2009-Ell1, Eric-2015-Splitting, Ramdas-2016-Fast}. However, we notice that $\sum_{k=1}^n \|D_k \beta\|_p$ is nonsmooth and does not admit an efficiently computed proximal mapping (even when $p=1$). Thus, the proximal gradient algorithms can not be directly applied to the filtering-clustering model in Eq.~\eqref{prob:main} with linear convergence.
\begin{definition}\label{def:grad-Lips}
$f$ is $\ell$-gradient Lipschitz if $\forall \beta, \beta' \in \br^d$, $\|\nabla f(\beta) - \nabla f (\beta')\| \leq \ell\|\beta-\beta'\|$. 
\end{definition}
\begin{definition}\label{def:strong-convex}
$f$ is $\mu$-strongly convex if $\forall \beta, \beta' \in \br^d$, $\|\nabla f(\beta) - \nabla f(\beta')\| \geq \mu \|\beta - \beta'\|$\footnote{The definition here is a consequence of the standard definition in the textbook~\citep{Nesterov-2018-Lectures}; that is, $(\beta - \beta')^\top(\nabla f(\beta) - \nabla f(\beta')) \geq \mu \|\beta - \beta'\|^2$. Nevertheless, we define it like this for simplicity.}. 
\end{definition}
In online setting, we impose unbiased and boundedness conditions on the stochastic gradient oracle that has become standard in the literature.
\begin{definition}
$g(\cdot, \xi)$ is unbiased and bounded if for $\forall \beta \in \br^d$, $\EE[g(\beta, \xi)] = \nabla f(\beta)$ and $\EE[\|g(\beta, \xi)\|^2] \leq G^2$ for a constant $G > 0$.  
\end{definition}
\begin{assumption}\label{Assumption:main}
The loss function $f$ is $\ell$-gradient Lipschitz and $\mu$-strongly convex. The stochastic gradient oracle $g(\cdot, \xi)$ is unbiased and bounded. The optimal set is nonempty. 
\end{assumption}
Under Assumption~\ref{Assumption:main}, the objective function $\Phi$ in Eq.~\eqref{prob:main} is \textit{non-smooth but strongly convex} and at least one optimal solution exists. Thus, the filtering-clustering model has a unique optimal solution $\beta^\star$. In general, a first-order optimization algorithm can not return an exact optimal solution in finite time. As such, we define an $\varepsilon$-optimal solution of the filtering-clustering model in Eq.~\eqref{prob:main}.
\begin{definition}\label{def:eps-opt}
$\beta \in \br^d$ is an $\varepsilon$-optimal solution of the filtering-clustering model in Eq.~\eqref{prob:main} if $\|\beta - \beta^\star\|^2 \leq \varepsilon$ given that $\beta^\star$ is a unique optimal solution. 
\end{definition}
With these notions in mind, one opts to develop first-order algorithms in that the required number of gradient evaluations to return an $\varepsilon$-optimal solution has the logarithmic dependence on $1/\varepsilon$ (deterministic or finite-sum setting) and the linear dependence on $1/\varepsilon$ (online setting) under Assumption~\ref{Assumption:main}.

\subsection{Specific instances}
We provide some typical examples of the filtering-clustering model in real-world application and give an overview of the existing algorithms. 

\textbf{$\ell_1$-trend filtering}~\citep{Kim-2009-Ell1, Ryan-2014-Adaptive} has been recently recognized as a benchmark approach to nonparametric regression in the area of statistical learning. The problem setup is given as follows: for $1 \leq i \leq \bar{n}$, we have a few input/response pairs $(\x_i, y_i)$ such that $y_i = f_0(\x_i) + w_i$ where $w_1, \ldots, w_{\bar{n}}$ are independent and identically distributed random variables. Given an integer $k \geq 0$, the $k$-th order $\ell_1$-trend filtering is implemented by solving the $\ell_1$-regularized least-squares problem:
\begin{equation}\label{prob:TF}
\min \limits_{\beta \in \br^{\bar{n}}} \ \frac{1}{2}\|y - \beta\|^2 + \lambda \|D^{(k+1)}\beta\|_1, 
\end{equation}
where $\lambda \geq 0$ is a regularization parameter and $D^{(k+1)} \in \br^{(\bar{n} - k - 1) \times \bar{n}}$ is a discrete difference operator of order $k+1$. For example, we have
\begin{equation*}
D^{(1)} = \begin{bmatrix}
-1 & 1 & 0 & \ldots & 0 & 0 \\ 0 & - 1 & 1 & \ldots & 0 & 0 \\ \vdots & \vdots & \vdots & \ldots & \vdots & \vdots \\ 0 & 0 & 0 & \ldots & - 1 & 1 
\end{bmatrix}. 
\end{equation*}
In general, the nonzero entries in each row of the matrix $D^{(k + 1)}$ are the $(k+1)$-th row of Pascal's triangle with alternating signs. When $k = 0$, the $\ell_1$-trend filtering model in Eq.~\eqref{prob:TF} can be interpreted as a special instance of 1-dimensional total variation denoising~\citep{Rudin-1992-Nonlinear} and fused Lasso~\citep{Tibshirani-2005-Sparsity}. Two classical algorithms for solving $\ell_1$-trend filtering model include primal-dual interior-point algorithm (PDIP)~\citep{Kim-2009-Ell1} and alternating direction of multiplier method (ADMM)~\citep{Ramdas-2016-Fast}. Despite their superior practical performance, these algorithms lack theoretical guarantees. In fact, we are not aware of any provably linearly convergent first-order algorithms that have been proposed for solving $\ell_1$-trend filtering model in Eq.~\eqref{prob:TF}. 

\textbf{Graph $\ell_1$-trend filtering}~\citep{Wang-2016-Trend} can be seen as an extension of $\ell_1$-trend filtering to graph problems. The problem setup is given as follows: let $G = (V, E)$ denote a graph consisting of a set of nodes $V = \{1, 2, \ldots, \bar{n}\}$ and undirected edges $E = (e_1, \ldots, e_{\bar{m}})$. Given $k \geq 0$ and a few inputs associated with the nodes, $y = (y_1, \ldots, y_{\bar{n}}) \in \br^{\bar{n}}$, the $k$-th order graph $\ell_1$-trend filtering is implemented by solving the following $\ell_1$-regularized least squares problem:
\begin{equation}\label{prob:GTF}
\min\limits_{\beta \in \br^{\bar{n}}} \ \frac{1}{2}\|y - \beta\|^2 + \lambda \|\Delta^{(k+1)}\beta\|_1, 
\end{equation}
where $\lambda \geq 0$ is a regularization parameter and $\Delta^{(k+1)} \in \br^{\bar{m} \times \bar{n}}$ is a discrete graph difference operator of order $k + 1$. Concretely, the explicit form of each row of $\Delta^{(1)}$ is given by: (if $e_l = (i, j)$ for $1 \leq l \leq m$)
\begin{equation*}
\Delta^{(1)}_l = (0, \ldots, \underbrace{-1}_i, \ldots, \underbrace{1}_j, \ldots, 0). 
\end{equation*}
Recursively, $\Delta^{(k + 1)}$ can be represented by
\begin{equation*}
\Delta^{(k + 1)} = \left\{\begin{array}{ll} (\Delta^{(1)})^\top \Delta^{(k)} & \textnormal{if } k \text{ is odd}, \\ \Delta^{(1)} \Delta^{(k)} & \textnormal{if } k \text{ is even}. \end{array}\right. 
\end{equation*}
When $k = 0$, the graph $\ell_1$-trend filtering model in Eq.~\eqref{prob:GTF} can be interpreted as a special instance of graph fused Lasso~\citep{Ryan-2011-The}. Two popular algorithms for solving graph $\ell_1$-trend filtering model include ADMM and specialized Newton's method~\citep{Wang-2016-Trend}. However, the theoretical convergence guarantee for these algorithms is missing, and it remains open whether a linearly convergent first-order algorithm exists for solving graph $\ell_1$-trend filtering model in Eq.~\eqref{prob:GTF} or not. 

\textbf{$\ell_2$-convex clustering} has been proposed as an effective alternative to classical clustering approaches in the literature and can be represented as a convex optimization model~\citep{Hocking-2011-Clusteringpath}. The problem setup is given as follows: given a number of inputs $\{x_1, x_2, \ldots, x_{\bar{n}}\} \subseteq \br^{\bar{n}}$, the $\ell_2$-convex clustering is implemented by solving the following $\ell_2$-regularized least square problem:
\begin{equation}\label{prob:CC}\small
\min\limits_{\{\beta_1, \ldots, \beta_{\bar{n}}\}} \frac{1}{2} \left(\sum_{i=1}^{\bar{n}} \|x_i - \beta_i\|^2\right) + \lambda \left(\sum_{1 \leq i < j \leq \bar{n}} w_{ij}\|\beta_i - \beta_j\|_2\right), 
\end{equation}
where $\lambda$ and $\{w_{ij}\}_{1 \leq i < j \leq \bar{n}}$ are both positive parameters. The benchmark first-order algorithms that have been used in practice for solving $\ell_2$-convex clustering model in Eq.~\eqref{prob:CC} include ADMM and alternating minimization algorithm (AMA)~\citep{Eric-2015-Splitting}. However, these algorithms are mostly heuristic and lack the solid theoretical guarantee. As an alternative,~\citet{Sun-2021-Convex} have proposed to use the second-order algorithm (e.g., semismooth Newton method) for solving $\ell_2$-convex clustering model and proved a linear convergence under certain condition. To the best of our knowledge, there are no provably linearly convergent first-order algorithms that have been proposed for solving $\ell_2$-convex clustering model in Eq.~\eqref{prob:CC}.

\subsection{Primal-dual filtering-clustering model}
We give an alternative form of the filtering-clustering model in Eq.~\eqref{prob:main} by starting with a convex-concave saddle-point formulation. Suppose $q = \frac{p}{p-1}$, we reformulate the filtering-clustering model equivalently as
\begin{equation}\label{prob:SP}
\min\limits_{\beta \in \br^d} \max_{\alpha \in \BB_q^n} \ f(\beta) - \lambda\alpha^\top D \beta, \quad \alpha \mydefn \begin{bmatrix} \alpha_1 \\ \vdots \\ \alpha_n \end{bmatrix}, D \mydefn \begin{bmatrix} D_1 \\ \vdots \\ D_n \end{bmatrix},
\end{equation}
where $\BB_q^n$ is the product of $n$ unit balls in $\ell_q$-norm. 

The saddle-point formulation in Eq.~\eqref{prob:SP} is different from the existing saddle-point formulation in~\citet{Lan-2017-Optimal} and~\citet{Zhang-2017-Stochastic}. To be more specific, the formulation in the previous works as applied to the filtering-clustering model in Eq.~\eqref{prob:main} is given by 
\begin{equation*}
\min_{\beta \in \br^d} \max_{\alpha \in \br^d} \ \alpha^\top\beta - f^\star(\alpha) + \lambda \left(\sum_{k=1}^n \|D_k\beta\|_p\right). 
\end{equation*}
where $f^\star: \br^d \rightarrow \br$ is the convex conjugate of the loss function $f$; see~\citet{Rockafellar-2015-Convex} for the definition. Here the proximal mapping of $\sum_{k=1}^n \|D_k\beta\|_p$ can not be efficiently computed in the filtering-clustering model~\citep{Parikh-2014-Proximal}. Therefore, the algorithms developed in~\citet{Lan-2017-Optimal} and~\citet{Zhang-2017-Stochastic} are not suitable for solving the filtering-clustering model in Eq.~\eqref{prob:main}. Second, the saddle-point formulation in Eq.~\eqref{prob:SP} also differs from the existing linearly constrained formulation in~\citet{Ramdas-2016-Fast}. In particular, the formulation in~\citet{Ramdas-2016-Fast} as applied to the filtering-clustering model in Eq.~\eqref{prob:main} is given by 
\begin{equation*}
\min\limits_{\beta \in \br^d, \alpha_k \in \br^m} \ f(\beta) + \lambda \left(\sum_{k=1}^n \|\alpha_k\|_p\right), \quad \st \ \alpha_k = D_k\beta.  
\end{equation*}
Based on the convex-concave saddle point formulation in Eq.~\eqref{prob:SP}, we obtain the dual filtering-clustering model as follows, 
\begin{equation}\label{prob:dual-main}
\min_{\alpha \in \BB_q^n} \ \bar{f}(\alpha) \mydefn f^\star(\lambda D^\top\alpha),
\end{equation}
where $f^\star: \br^d \mapsto \br$ is the convex conjugate of the loss function $f$. As such, the dual filtering-clustering model in Eq.~\eqref{prob:dual-main} is derived by, 
\begin{eqnarray*}
\lefteqn{\min_{\beta \in \br^d} \max_{\alpha \in \BB_q^n} \ f(\beta) - \lambda\alpha^\top D\beta \ \Longleftrightarrow \ \max_{\alpha \in \BB_q^n} \min_{\beta \in \br^d} \ f(\beta) - \lambda\alpha^\top D\beta} \\
& \Longleftrightarrow & \max_{\alpha \in \BB_q^n} \ -f^\star(\lambda\alpha^\top D) \ \Longleftrightarrow \ \min_{\alpha \in \BB_q^n} \ f^\star(\lambda\alpha^\top D). 
\end{eqnarray*}
Notably, the above dual filtering-clustering model is amenable to structure analysis and algorithmic design. Indeed, $f^\star$ is a smooth and strongly convex function and $\BB_q$ is a structured and bounded convex set with efficient projection when $q = 1, 2, +\infty$. In the sequel, we demonstrate that a global error bound condition is satisfied for the dual filtering-clustering model in Eq.~\eqref{prob:dual-main}, making it possible to develop a class of first-order optimization algorithms with desirable convergence guarantee in different settings. 

\section{Global Error Bound Condition}\label{sec:GEB}
We prove that the global error bound (GEB) condition holds for the dual filtering-clustering model in Eq.~\eqref{prob:dual-main}. Indeed, we investigate the special structure of the dual filtering-clustering model and use them to prove that the GEB condition holds for the dual filtering-clustering model when $q \in [1, 2] \cup \{+\infty\}$ under certain conditions, which correspond to the filtering-clustering model in Eq.~\eqref{prob:main} with $p \in \{1\} \cup [2, +\infty]$.  All the proof details are deferred to the appendix. 

\subsection{Problem structure}\label{subsec:prob}
We investigate the special structure of the dual filtering-clustering model concerning the objective function and the optimal set. Before our formal analysis, we recall some notations: 
\begin{eqnarray*}
f^\star(z) & = & \min_{\beta\in\br^d} \ f(\beta) -z^\top \beta, \\
\beta^\star(\alpha) & = & \argmin_{\beta\in\br^d} \ f(\beta) - \lambda\alpha^\top D\beta, \\
\bar{f}(\alpha) & = & f^\star(\lambda D^\top\alpha). 
\end{eqnarray*}
where $f^\star$ is the convex conjugate of the loss function $f$ and $\bar{f}$ is the objective function of the dual filtering-clustering model in Eq.~\eqref{prob:dual-main}. 
\begin{lemma} \label{Lemma:conjugate} 
Under Assumption~\ref{Assumption:main}, $f^\star$ is $\frac{1}{\mu}$-gradient Lipschitz and $\frac{1}{\ell}$-strongly convex. 
\end{lemma}
\begin{lemma}\label{Lemma:minimizer}
Under Assumption~\ref{Assumption:main}, $\beta^\star$ is a Lipschitz function over $\BB_q^n$ with parameter $\frac{\lambda\|D\|}{\mu}$. 
\end{lemma}
\begin{lemma}\label{Lemma:dual-objective}
Under Assumption~\ref{Assumption:main}, $\bar{f}$ is differentiable and $\frac{\lambda^2\|D\|^2}{\mu}$-gradient Lipschitz. 
\end{lemma}
From the above lemmas, we see that $f^\star$ and $\bar{f}$ have favorable structures from an algorithmic point of view. 

\subsection{GEB condition and ULC property}
We introduce the notion of \textit{upper Lipschitz continuity} (ULC) and use it to prove a sufficient condition for the dual filtering-clustering model in Eq.~\eqref{prob:dual-main} to satisfy the global error bound (GEB) condition. In particular, we first introduce the quantity $d(\alpha, \Omega^\star)$ that measures the distance between a point $\alpha$ and the optimal set $\Omega^\star$. Note that this quantity is generally not accessible since $\Omega^\star$ is unknown. As an alternative, we consider a function $R: \br^{mn} \rightarrow \br^{mn}$, which we refer to as \textit{residual function}. Formally, it is given by 
\begin{equation}\label{def:residue}
R(\alpha) \mydefn \PCal_{\BB_q^n}(\alpha - \nabla \bar{f}(\alpha)) - \alpha. 
\end{equation}
It is easy to verify that $R(\alpha) = 0$ if and only if $\alpha \in \Omega^\star$. Notably, the quantity $R(\alpha)$ can be computed for any $\alpha \in \br^{mn}$ given the access to $\nabla \bar{f}(\cdot)$. As such, this suggests that $\|R(\alpha)\|$ is a reasonable surrogate for characterizing the optimality of $\alpha \in \BB_q^n$. Then, it is natural to ask whether $\|R(\alpha)\|$ is related to $d(\alpha, \Omega^\star)$ or not, further motivating the GEB condition for the dual filtering-clustering model in Eq.~\eqref{prob:dual-main}. 
\begin{definition} 
The dual filtering-clustering model in Eq.~\eqref{prob:dual-main} satisfies a GEB condition if there exists $\tau > 0$ such that $d(\alpha, \Omega^\star) \leq \tau\|R(\alpha)\|$ for all $\alpha \in \BB_q^n$.
\end{definition}
The GEB condition is a relaxed notion of global strong convexity~\citep{Pang-1987-Posteriori}. After removing the constraint set $\BB_q^n$, we see that $R(\alpha) = -\nabla \bar{f}(\alpha)$ and the GEB condition is satisfied when $\bar{f}$ is strongly convex. The above definition is only given for the sake of completeness and does not give the insights on the range of $q$ in which the GEB condition holds for the dual filtering-clustering model in Eq.~\eqref{prob:dual-main}. This is because the residual function $R$ is in the abstract form and it remains elusive which value of $q$ can guarantee that $d(\alpha, \Omega^\star) \leq \tau\|R(\alpha)\|$ holds for some $\tau > 0$. ~\citet{Zhou-2015-Error} have presented an alternative approach based on the notion of ULC property of set-valued mapping and used it to identify the range of $p$ in which the local error bound condition holds for $\ell_{1,p}$-norm regularized problems. Combined with some nontrivial modifications, we show that this approach can be adopted here. 

We compute an upper bound for $\tau$ in a special example of the dual filtering-clustering model in Eq.~\eqref{prob:dual-main}. Since the graph $\ell_1$-trend filtering reduces to $\ell_1$-trend filtering if we choose a proper graph, we focus on $\ell_1$-trend filtering. Our example shows that a dimension-free upper bound for $\tau$ in some application problems.  

Consider fused Lasso~\citep{Tibshirani-2005-Sparsity} which is the $\ell_1$-trend filtering with $k=0$. For simplicity, we assume that $y = 0$ is an input data and $\lambda = 1$. Then, the problem in Eq.~\eqref{prob:TF} is given by  
\begin{equation}\small
\min \limits_{\beta \in \br^{\bar{n}}} \ \frac{1}{2}\|\beta\|^2 + \|D\beta\|_1,  \ \st \ D = \begin{bmatrix}
-1 & 1 & 0 & \ldots & 0 & 0 \\ 0 & - 1 & 1 & \ldots & 0 & 0 \\ \vdots & \vdots & \vdots & \ldots & \vdots & \vdots \\ 0 & 0 & 0 & \ldots & - 1 & 1 \end{bmatrix}. 
\end{equation}
which implies the dual filtering-clustering model: 
\begin{equation*}
\min \limits_{\alpha \in \br^{\bar{n} - 1}} \ \frac{1}{2}\|D^\top \alpha\|^2, \quad \st -1 \leq \alpha_i \leq 1, \ \forall 1 \leq i \leq \bar{n} - 1. 
\end{equation*}
It is easy to verify that 
\begin{equation*}
DD^\top \succeq \begin{bmatrix}
1 & 0 & 0 & \cdots & 0 & 0 \\ 
0 & 1 & 0 & \ldots & \vdots & \vdots \\ 
\vdots & \vdots & \ddots & \ldots & \vdots & \vdots \\ 
\vdots & \vdots & \vdots & \ldots & 1 & 0 \\ 
0 & 0 & 0 & \ldots & 0 & 1 
\end{bmatrix}
\end{equation*}
which implies that $\bar{f}(\alpha) = (1/2)\|D^\top \alpha\|^2$ is $1$-strongly convex. Combining it with~\citet[Corollary~3.6]{Drusvyatskiy-2018-Error}, we have $\tau \leq 4$. 
\begin{remark}
Computing an upper bound of $\tau$ is very difficult in general and has been a research topic by itself in the community. Indeed,~\citet{Drusvyatskiy-2018-Error} provided a comprehensive qualitative treatment of error bound with quantitative studies for special cases and~\citet{Wang-2014-Iteration} studied machine learning problems and algorithms using error bound condition. We are also aware of a recent work~\citep{Pena-2018-Algorithm} that proposed an algorithm to compute the error bound. Can we compute a bound of $\tau$ for special graph $\ell_1$-trend filtering? We leave it to future work. 
\end{remark}
We present our main results on the GEB condition holds for the dual filtering-clustering model when $q \in [1, 2] \cup \{+\infty\}$, which correspond to the filtering-clustering model in Eq.~\eqref{prob:main} with $p \in \{1\} \cup [2, +\infty]$. 
\begin{theorem}\label{Theorem:GEB-polyhedron}
Under Assumption~\ref{Assumption:main} and let $q \in \{1, +\infty\}$, the GEB condition holds for the dual filtering-clustering model in Eq.~\eqref{prob:dual-main}. 
\end{theorem}
Note that $\g^\star = \nabla\bar{f}(\alpha^\star)$ for all optimal solutions $\alpha^\star$ (see Proposition~\ref{Prop:set-map}),  we let $\JCal = \{j \in [n] \mid \g_j^\star \neq 0\}$ be the set of indices of nonzero coordinates. 
\begin{theorem}\label{Theorem:GEB-other}
Under Assumption~\ref{Assumption:main} and let $q \in (1, 2]$ and $\JCal = [n]$, the GEB condition holds for the dual filtering-clustering model in Eq.~\eqref{prob:dual-main}.
\end{theorem}
\begin{remark}
$\JCal = [n]$ is necessary for deriving the GEB condition when $q \in (1, 2]$ (see Appendix~\ref{app:counterexample} for an example) and also assumed in~\citet[Theorem~6]{Jane-2021-Variational} for proving the metric subregularity. As argued in~\citet[Remark~2]{Jane-2021-Variational}, this assumption is mild in terms of applications since an optimal solution often lie on the boundary of a constraint set.  
\end{remark}

\section{Algorithmic Framework}\label{sec:framework}
We propose and analyze a \textit{generalized dual gradient ascent} (GDGA) algorithm for solving Eq.~\eqref{prob:main}. It can be interpreted as an inexact forward-backward splitting algorithm~\citep{Tseng-1991-Applications, Tseng-2000-Modified} and reduces to the alternating minimization algorithm (AMA)~\citep{Eric-2015-Splitting} when specialized to convex clustering. 

\subsection{Generalized dual gradient ascent}
Our GDGA algorithm is a first-order optimization algorithm that only accesses the gradient of $f$, the matrix $D$ and the regularization parameter $\lambda$. Since $f^\star$ does not admit an analytic form in general, we design a subroutine which returns an approximation of $\nabla f^\star$ by minimizing the smooth and strongly convex objective $f(\beta) - \lambda\alpha_t^\top D\beta$ with respect to $\beta$; see Algorithm~\ref{Algorithm:GDGA}. 
\begin{algorithm}[!t]\small
\caption{Generalized Dual Gradient Ascent}\label{Algorithm:GDGA}
\begin{algorithmic}
\STATE \textbf{Input:} learning rate $\eta > 0$ and tolerance $\varepsilon > 0$.
\STATE \textbf{Initialization:} $\beta_0 \in \br^d$, $\alpha_1 \in \BB_q^n$ and the other tolerance $\hat{\varepsilon} > 0$. 
\FOR{$t = 1, 2, \ldots, T$}
\STATE $\beta_t \leftarrow \textsc{InnerLoop}(f, \lambda, D, \alpha_t, \beta_{t-1}, \hat{\varepsilon})$. 
\STATE $\alpha_{t+1} \leftarrow \PCal_{\BB_q^n} (\alpha_t - \eta \lambda D\beta_t)$.
\ENDFOR
\STATE \textbf{Return:} $\beta_{T+1}$. 
\end{algorithmic}
\end{algorithm}

Algorithm~\ref{Algorithm:GDGA} is simple, matrix-free, and amenable to distributed implementation. It outperforms the existing approaches~\citep{Kim-2009-Ell1, Wang-2016-Trend, Ramdas-2016-Fast} which require matrix decomposition and suffer from scalability. Specialized to $\ell_2$-convex clustering model, Algorithm~\ref{Algorithm:GDGA} becomes inexact AMA~\citep{Eric-2015-Splitting} and exhibits fast convergence in practice. It is also worth mentioning that the semismooth Newton method~\citep{Sun-2021-Convex} can outperform our algorithm by exploiting the special structure of Jacobian matrix in $\ell_2$-convex clustering model. However, it appears to be difficult to extend such approach to solve general filtering-clustering model due to the complicated nonsmooth term $\sum_{k=1}^n \|D_k \beta\|_p$.  Further, the subroutine can be constructed based on different algorithmic components. Indeed, $\beta_t \leftarrow \textsc{InnerLoop}(f, \lambda, D, \alpha_t, \beta_{t-1}, \hat{\varepsilon})$ is an $\hat{\varepsilon}$-minimizer of $f(\beta) - \lambda\alpha_t^\top D\beta$ satisfying that $\|\beta_t - \beta^\star(\alpha_t)\| \leq \hat{\varepsilon}$ where $\beta^\star(\alpha_t) \mydefn \argmin_{\beta \in \br^d} f(\beta) - \lambda\alpha_t^\top D\beta$. The subroutine implements some fast first-order optimization algorithms with an initial point $\beta_{t-1}$. For example, we can apply celebrated Nesterov's accelerated gradient descent~\citep{Nesterov-2018-Lectures} in the deterministic setting and optimal stochastic gradient descent~\citep{Rakhlin-2012-Making} in the online setting. If $f(\beta) = (1/2)\|\beta - y\|^2$, this subroutine can be removed since we have the analytic form of $\beta^\star(\alpha)$.  

Algorithm~\ref{Algorithm:GDGA} enjoys a solid theoretical guarantee. Indeed, we prove that it achieves linear convergence without counting the number of gradient or stochastic gradient oracles used in the subroutine. Although the proof idea is not new but follows the standard strategy~\citep{Luo-1992-Linear, Luo-1993-Error, Wang-2014-Iteration} given the global error bound condition previously established for the dual filtering-clustering models in Eq.~\eqref{prob:dual-main}, we provide the rigorous computational theory for the filtering-clustering model, shedding the light on great potential of AMA to pursue a high-accurate solution in practice.  

\subsection{Complexity of GDGA algorithmic framework}
We establish the linear convergence of the GDGA algorithmic framework without counting the number of gradient or stochastic gradient oracles used in the subroutine. We hope to remark that our results are nontrivial due to the following reasons:

First, Eq.~\eqref{prob:main} is strongly convex but nonsmooth and the term $\sum_{k=1}^n \|D_k \beta\|_p$ is not computationally favorable. Indeed, the proximal mapping of $\sum_{k=1}^n \|D_k \beta\|_p$ does not have the closed form even when $p=1$ and can not be efficiently computed in general. This makes proximal gradient algorithm~\citep{Parikh-2014-Proximal} not applicable. Second,  $f^\star$ is smooth and strongly convex. However, the objective function $\bar{f}$ of the dual filtering-clustering model in Eq.~\eqref{prob:dual-main} is non-strongly convex since a matrix $D^\top$ can be degenerate in real application, e.g., the graph $\ell_1$-trend filtering model with $\bar{m} \gg \bar{n}$. In addition, $\nabla \bar{f}$ is not available to the algorithm and necessities an efficient subroutine in both theory and practice.  Finally, the objective function of the saddle-point formulation in Eq.~\eqref{prob:SP} is strongly convex in $\beta$ but linear in $\alpha$ with a ball constraint set $\BB_q^n$. As such, we can not derive the linear convergence of a few existing primal-dual optimization algorithms as a consequence of the existing results~\citep{Palaniappan-2016-Stochastic, Wang-2017-Exploiting}. 

We present our main theorem and refer the readers to the appendix for the full version of Theorem~\ref{Theorem:GDGA-Framework}. 
\begin{theorem}\label{Theorem:GDGA-Framework}
Under Assumption~\ref{Assumption:main} and set $\eta \in (0, \min\{1, \frac{\mu}{4\|D\|^2\max\{1, \lambda^2\}}\})$ in Algorithm~\ref{Algorithm:GDGA} and $\hat{\varepsilon} > 0$ properly. Then, the number of iterations to return an $\varepsilon$-optimal solution is $T = O(\log(1/\varepsilon))$. 
\end{theorem}
\begin{corollary}\label{Corollary:GDGA-Stochastic-Framework}
Under same setting as Theorem~\ref{Theorem:GDGA-Framework} with stopping criterion in the subroutine as $\EE[\delta_t] \leq \hat{\varepsilon}$, the number of iterations required by the stochastic GDGA algorithm to return an $\varepsilon$-optimal solution is bounded by $T = O(\log(1/\varepsilon))$. 
\end{corollary}
The key ingredient here that enables an improved linear convergence rate is that a global error bound condition holds for the dual filtering-clustering model in Eq.~\eqref{prob:dual-main}. If the dual form of any optimization problems admits same structure, we can develop an algorithm with the same convergence guarantee. However, it is nontrivial to prove a global error bound for specific problems and the case of the dual filtering-clustering model is rare. As such, it is difficult to generalize our approach to generic convex optimization problems. 

We study the case with $f(\beta) = (1/2)\|\beta - y\|^2$ and demonstrate that Algorithm~\ref{Algorithm:GDGA-simplified} suffices to solve many widely used filtering-clustering models, e.g., $\ell_1$-trend filtering and $\ell_2$-convex clustering. 
\begin{theorem}\label{Theorem:GDGA-simplified}
Under Assumption~\ref{Assumption:main} and set $\eta \in (0, \min\{1, \frac{\mu}{4\|D\|^2\max\{1, \lambda^2\}}\})$ in Algorithm~\ref{Algorithm:GDGA-simplified}. Then, the number of gradient evaluations to return an $\varepsilon$-optimal solution is $N = O(\log(1/\varepsilon))$.
\end{theorem}
\begin{algorithm}[!t]\small
\caption{Simplified GDGA}\label{Algorithm:GDGA-simplified}
\begin{algorithmic}
\STATE \textbf{Input:} learning rates $\eta > 0$ and tolerance $\varepsilon > 0$. 
\STATE \textbf{Initialization:} $\beta_0 \in \br^d$ and $\alpha_1 \in \BB_q^n$.  
\FOR{$t = 1, 2, \ldots, T$}
\STATE $\beta_t = \lambda D^{\top} \alpha_t + y$. 
\STATE $\alpha_{t+1} \leftarrow \PCal_{\BB_q^n}(\alpha_t - \eta \lambda D\beta_t)$.
\ENDFOR
\STATE \textbf{Return:} $\beta_{T+1}$. 
\end{algorithmic}
\end{algorithm}
\begin{algorithm}[!t]\small
\caption{Deterministic GDGA}\label{Algorithm:GDGA-deterministic}
\begin{algorithmic}
\STATE \textbf{Input:} learning rates $\eta > 0$ and tolerance $\varepsilon > 0$. 
\STATE \textbf{Initialization:} $\beta_0 \in \br^d$ and $\alpha_1 \in \BB_q^n$. 
\FOR{$t = 1, 2, \ldots, T$}
\STATE $\beta_t \leftarrow \textsc{InnerLoopAgd}(f(\cdot), \lambda, D, \alpha_t, \beta_{t-1}, \hat{\varepsilon})$. 
\STATE $\alpha_{t+1} \leftarrow \PCal_{\BB_q^n}(\alpha_t - \eta \lambda D\beta_t)$.
\ENDFOR
\STATE \textbf{Return:} $\beta_{T+1}$. 
\end{algorithmic}
\end{algorithm}
For the deterministic setting in which $\nabla f$ is accessible by $\textsc{InnerLoop}$, we do not have the closed-form minimizer of $f(\beta) - \alpha_t^\top D\beta$ and obtain an $\hat{\varepsilon}$-minimizer by implementing $\textsc{InnerLoop}$ using AGD.  
\begin{theorem}\label{Theorem:GDGA-deterministic}
Under Assumption~\ref{Assumption:main} and set $\eta \in (0, \min\{1, \frac{\mu}{4\|D\|^2\max\{1, \lambda^2\}}\})$ and $\hat{\varepsilon} > 0$ in Algorithm~\ref{Algorithm:GDGA-deterministic}. The number of gradient evaluations to return an $\varepsilon$-optimal solution is $N = O(\sqrt{\ell/\mu}\log(1/\varepsilon))$. 
\end{theorem}
\begin{algorithm}[!t]\small
\caption{Stochastic Variance Reduced GDGA}\label{Algorithm:GDGA-finite-sum}
\begin{algorithmic}
\STATE \textbf{Input:} learning rates $\eta > 0$ and tolerance $\varepsilon > 0$. 
\STATE \textbf{Initialization:} $\beta_0 \in \br^d$ and $\alpha_1 \in \BB_q^n$. 
\FOR{$t = 1, 2, \ldots, T$}
\STATE $\beta_t \leftarrow \textsc{InnerLoopKatyusha}(\frac{1}{M}(\sum_{i=1}^M f_i(\cdot)), \lambda, D, \alpha_t, \beta_{t-1}, \hat{\varepsilon})$. 
\STATE $\alpha_{t+1} \leftarrow \PCal_{\BB_q^n}(\alpha_t - \eta \lambda D\beta_t)$.
\ENDFOR
\STATE \textbf{Return:} $\beta_{T+1}$. 
\end{algorithmic}
\end{algorithm}
\begin{figure*}[!t]
\begin{minipage}[b]{.33\textwidth}
\includegraphics[width=60mm,height=45mm]{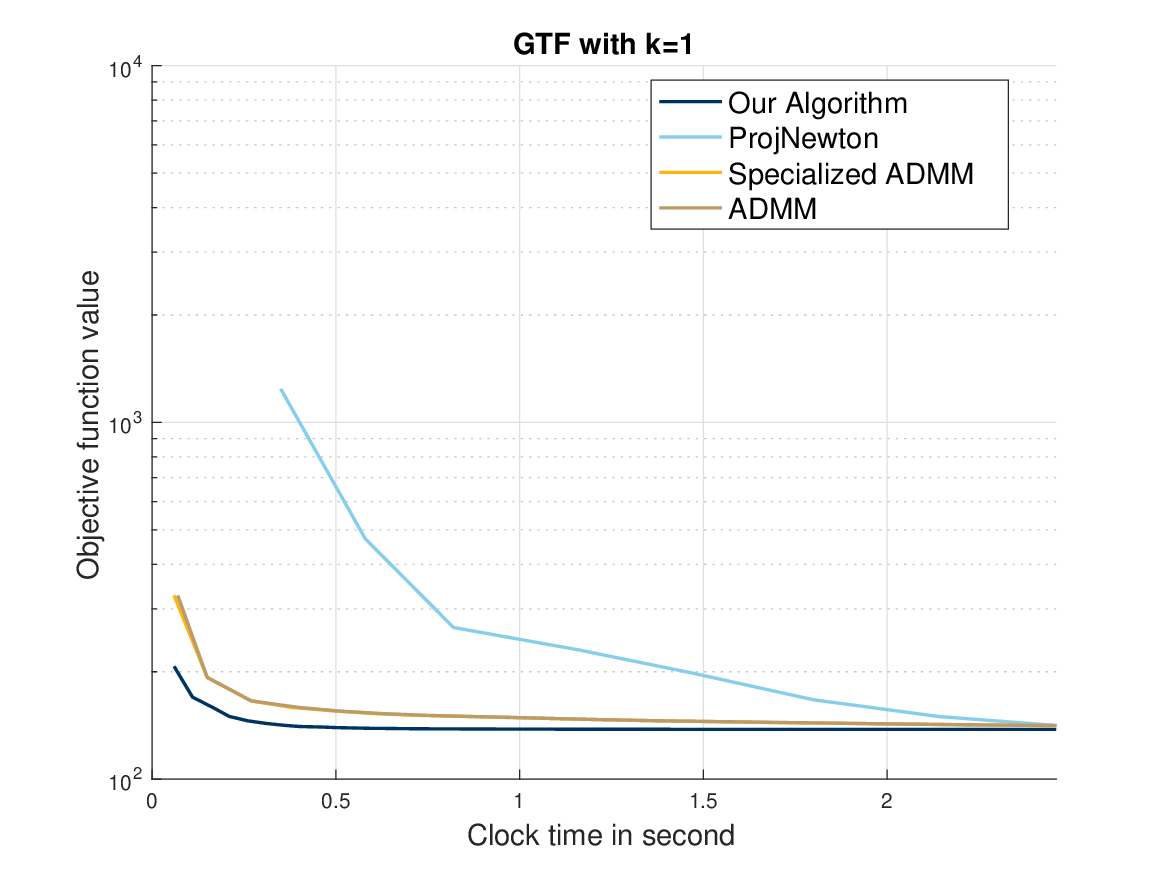}
\end{minipage}
\begin{minipage}[b]{.33\textwidth}
\includegraphics[width=60mm,height=45mm]{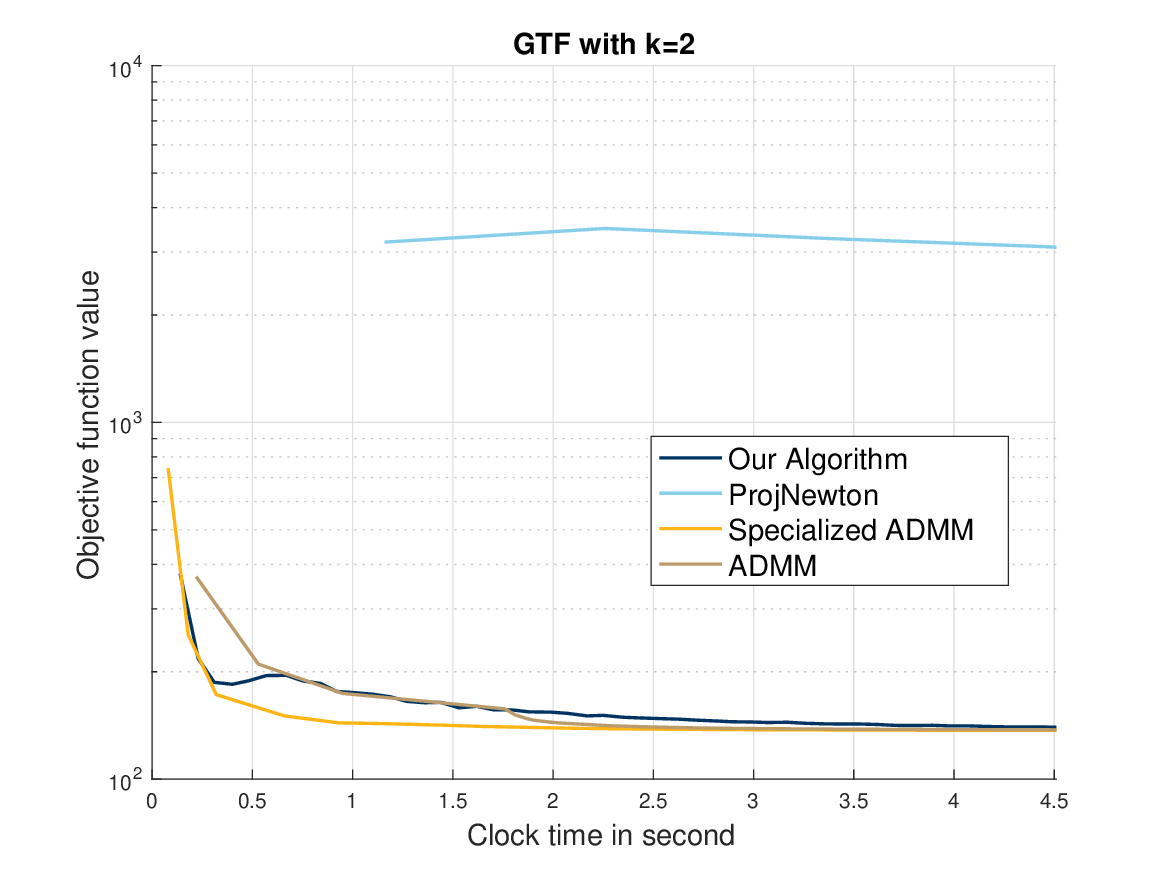}
\end{minipage}
\begin{minipage}[b]{.33\textwidth}
\includegraphics[width=60mm,height=45mm]{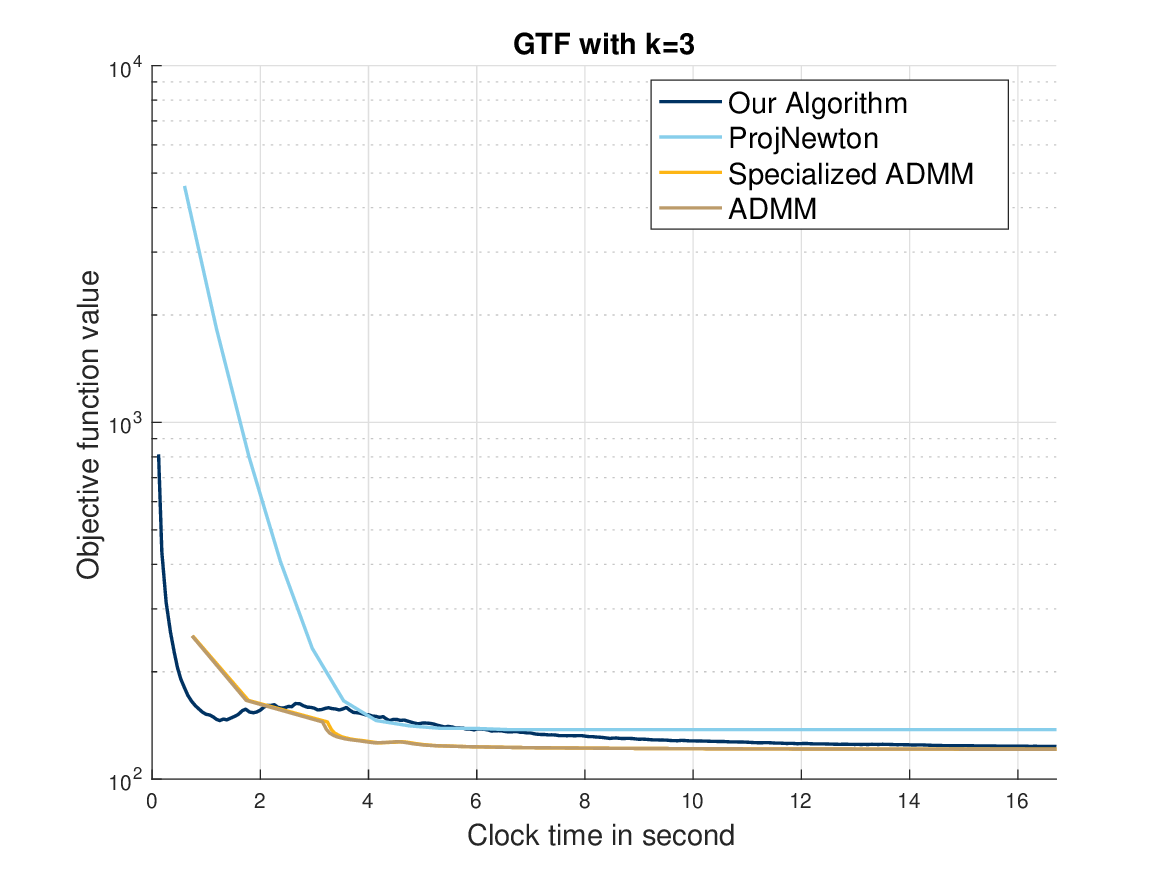}
\end{minipage} \\
\begin{minipage}[b]{.33\textwidth}
\includegraphics[width=60mm,height=45mm]{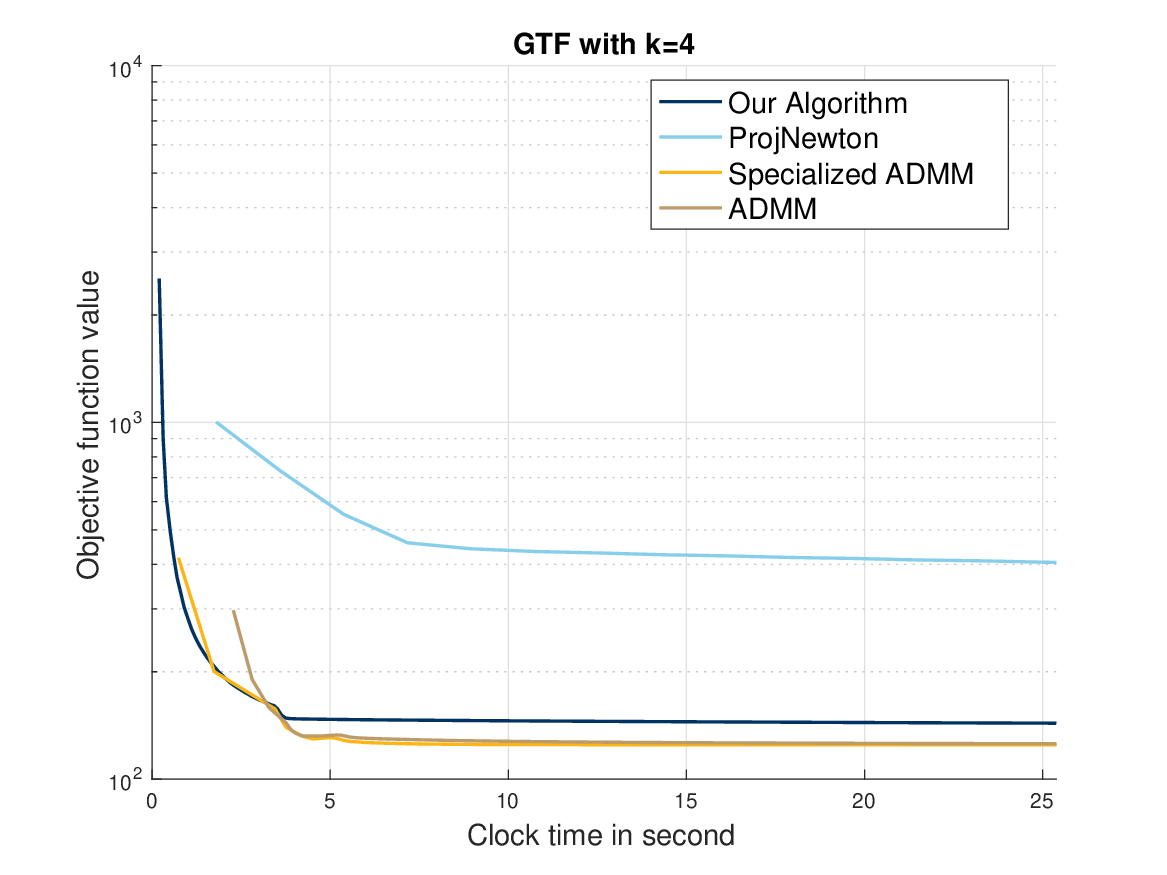}
\end{minipage}
\begin{minipage}[b]{.33\textwidth}
\includegraphics[width=60mm,height=45mm]{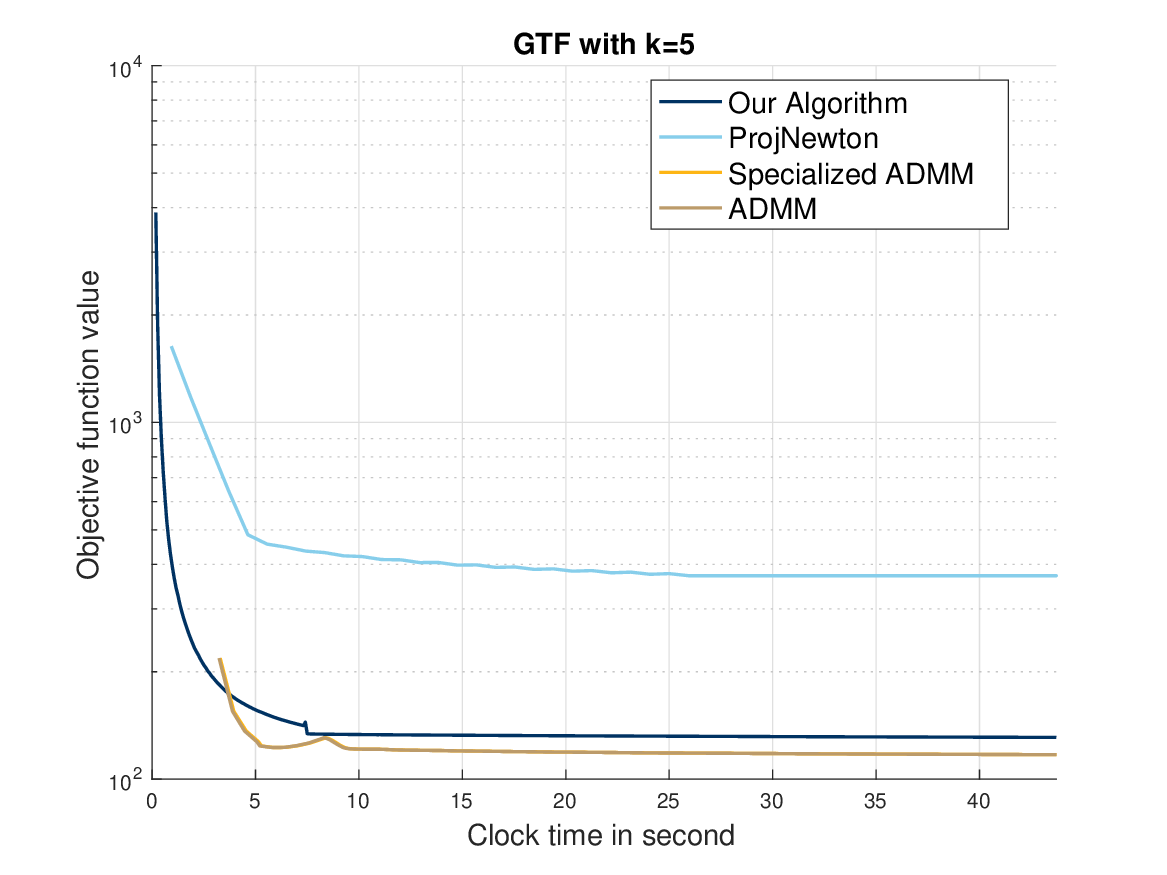}
\end{minipage}
\begin{minipage}[b]{.33\textwidth}
\includegraphics[width=60mm,height=45mm]{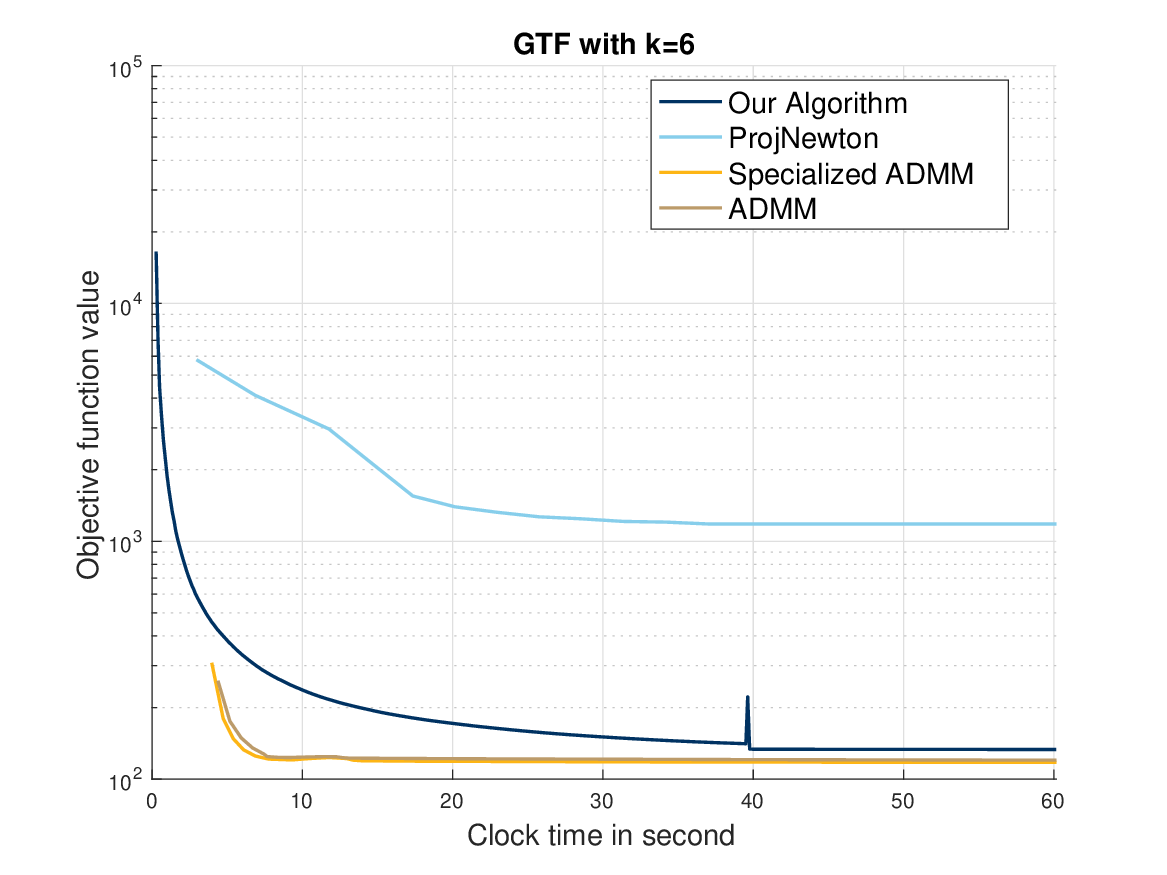}
\end{minipage}
\caption{\small{Comparison of Algorithm~\ref{Algorithm:GDGA-simplified}, ADMM, specialized ADMM, and projected Newton method on medium image.}} \label{fig:TF_small}
\end{figure*}
For the finite-sum setting in which the loss function $f$ is of the form $\frac{1}{M}\sum_{i=1}^M f_i$, we obtain an $\hat{\varepsilon}$-minimizer by implementing $\textsc{InnerLoop}$ using Katyusha.   
\begin{theorem}\label{Theorem:GDGA-finite-sum}
Under Assumption~\ref{Assumption:main} and set $\eta \in (0, \min\{1, \frac{\mu}{4\|D\|^2\max\{1, \lambda^2\}}\})$ and $\hat{\varepsilon} > 0$ in Algorithm~\ref{Algorithm:GDGA-finite-sum}. The number of component gradient evaluations to get $\varepsilon$-optimal solution is $N = O((M + \sqrt{\ell M/\mu})\log(1/\varepsilon))$. 
\end{theorem}
Theorem~\ref{Theorem:GDGA-finite-sum} guarantees the linear convergence rate for Algorithm~\ref{Algorithm:GDGA-finite-sum} which outperforms Algorithm~\ref{Algorithm:GDGA-deterministic} in terms of the required number of component gradient evaluations by shaving off $\sqrt{M}$. 
\begin{algorithm}[!t]\small
\caption{Stochastic GDGA}\label{Algorithm:GDGA-stochastic}
\begin{algorithmic}
\STATE \textbf{Input:} learning rates $\eta > 0$ and tolerance $\varepsilon > 0$. 
\STATE \textbf{Initialization:} $\beta_0 \in \br^d$ and $\alpha_1 \in \BB_q^n$. 
\FOR{$t = 1, 2, \ldots, T$}
\STATE $\beta_t \leftarrow \textsc{InnerLoopSgd}(\EE_P[F(\cdot, \xi)], \lambda, D, \alpha_t, \beta_{t-1}, \hat{\varepsilon})$. 
\STATE $\alpha_{t+1} \leftarrow \PCal_{\BB_q^n}(\alpha_t - \eta \lambda D\beta_t)$.
\ENDFOR
\STATE \textbf{Return:} $\beta_{T+1}$. 
\end{algorithmic}
\end{algorithm}

For the online setting where $f(\cdot) = \EE_P[F(\cdot, \xi)]$ and $\xi \sim P$ is the streaming data that should be processed incrementally without having access to all data, we implement the subroutine using SGD.
\begin{theorem}\label{Theorem:GDGA-stochastic}
Under Assumption~\ref{Assumption:main} and set $\eta \in (0, \min\{1, \frac{\mu}{4\|D\|^2\max\{1, \lambda^2\}}\})$ and $\hat{\varepsilon} > 0$ in Algorithm~\ref{Algorithm:GDGA-stochastic}. The number of stochastic gradient evaluations to return an $\varepsilon$-optimal solution is $N = O((1/\varepsilon)\log(1/\varepsilon))$. 
\end{theorem}
Theorem~\ref{Theorem:GDGA-stochastic} guarantees the sublinear convergence rate for Algorithm~\ref{Algorithm:GDGA-stochastic}. Our results match the lower bound of any stochastic first-order algorithm for nonsmooth and strongly convex setting up to log factors, showing that Algorithm~\ref{Algorithm:GDGA-stochastic} is the best possible that we can expect. 
\begin{remark}
We apply different subroutines when the problem has different structure so that the best possible theoretical convergence rate can be achieved. However, the choice of subroutine in practice is different and it is possible that specific subroutines are suitable for solving the particular problems.  This is beyond the scope of our paper and we leave the answers to future work.
\end{remark}

\section{Experiments}\label{sec:simulation}
We carefully conduct the experiments on $\ell_1$-trend filtering to demonstrate the effectiveness of Algorithm~\ref{Algorithm:GDGA-simplified}. The baseline approaches include ADMM, specialized ADMM~\citep{Ramdas-2016-Fast} and projected Newton method~\citep{Wang-2016-Trend} and we consider three real images with various sizes: 128 by 128 pixels (small), 256 by 256 pixels (medium) and 512 by 512 pixels (large)\footnote{These images can be found at: \url{http://sipi.usc.edu/database/database.php?volume=misc}}. All the algorithms are evaluated as the order $k$ varies in the discrete difference operator $D^{(k+1)}$ in which the evaluation metric is the objective function value. We choose $\lambda = 0.2$ in our experiment and consider the adaptive step-size $\eta_t > 0$ based on Barzilai-Borwein rule~\citep{Fletcher-2005-Barzilai}. 

We present the numerical results on different images in Figure~\ref{fig:TF_small} (see Appendix for other figures). Algorithm~\ref{Algorithm:GDGA-simplified} is comparable with ADMM and specialized ADMM, all of which significantly outperform projected Newton method. More specifically, Algorithm~\ref{Algorithm:GDGA-simplified} is the best when $k = 1$ remain effective as $k$ increases. Note that ADMM and specialized ADMM are robust since they conduct matrix decomposition which significantly alleviates the ill-conditioning of the matrix $D^{(k + 1)}$ as $k$ increases. Compared to the ADMM-type methods, our methods require more iterations to reach the same tolerance but enjoy lower per-iteration computational cost since our methods are matrix-free.  For small datasets,  the ADMM-type methods can work quite well since the matrix inversion is not an issue.  In fact, we find that the ADMM-type methods are more robust than our methods if the matrix inversion is not an issue and the hyperparameters tuning for our methods needs more work. However, the ADMM-type methods are likely to fail for large datasets since the matrix inversion becomes a computational bottleneck. In contrast, Algorithm~\ref{Algorithm:GDGA-simplified} is matrix-free and we find that the Barzilai-Borwein rule can speed up the algorithm by exploring the curvature information and alleviate the ill-conditioning. As such, our algorithms can be good alternatives to the ADMM-type methods in practice.

\section{Conclusion}\label{sec:conclu}
This paper contributes to the landscape of computational aspect of filtering-clustering model in Eq.~\eqref{prob:main} by identifying a global error bound condition, which is satisfied by a large class of dual filtering-clustering problems, and designing a class of generalized dual gradient ascent algorithm, which is \textit{optimal} first-order algorithms in the deterministic, finite-sum and online settings. Our results are new and shed the light on superior performance of several first-order optimization algorithms as applied to solve the filtering-clustering models in practice. We also conduct experiments on real datasets and the numerical results demonstrate the effectiveness of our algorithms.

\section*{Acknowledgments}
We would like to thank the AC and five reviewers for suggestions that improve the quality of this paper. This work was supported in part by the Mathematical Data Science program of the Office of Naval Research under grant number N00014-18-1-2764.

\bibliographystyle{plainnat}
\bibliography{ref}

\begin{thebibliography}{57}
\providecommand{\natexlab}[1]{#1}
\providecommand{\url}[1]{\texttt{#1}}
\expandafter\ifx\csname urlstyle\endcsname\relax
  \providecommand{\doi}[1]{doi: #1}\else
  \providecommand{\doi}{doi: \begingroup \urlstyle{rm}\Url}\fi

\bibitem[Agarwal et~al.(2012)Agarwal, Negahban, and
  Wainwright]{Agarwal-2012-Fast}
A.~Agarwal, S.~Negahban, and M.~J. Wainwright.
\newblock Fast global convergence of gradient methods for high-dimensional
  statistical recovery.
\newblock \emph{Annals of Statistics}, 40\penalty0 (5):\penalty0 2452--2482,
  2012.

\bibitem[Allen-Zhu(2017)]{Allen-2017-Katyusha}
Z.~Allen-Zhu.
\newblock Katyusha: The first direct acceleration of stochastic gradient
  methods.
\newblock In \emph{STOC}, pages 1200--1205. ACM, 2017.

\bibitem[Bauschke and Borwein(1996)]{Bauschke-1996-Projection}
H.~H. Bauschke and J.~M. Borwein.
\newblock On projection algorithms for solving convex feasibility problems.
\newblock \emph{SIAM Review}, 38\penalty0 (3):\penalty0 367--426, 1996.

\bibitem[Beck and Shtern(2017)]{Beck-2017-Linearly}
A.~Beck and S.~Shtern.
\newblock Linearly convergent away-step conditional gradient for non-strongly
  convex functions.
\newblock \emph{Mathematical Programming}, 164\penalty0 (1-2):\penalty0 1--27,
  2017.

\bibitem[Chen and Rockafellar(1997)]{Chen-1997-Convergence}
G.~H.~G. Chen and R.~T. Rockafellar.
\newblock Convergence rates in forward--backward splitting.
\newblock \emph{SIAM Journal on Optimization}, 7\penalty0 (2):\penalty0
  421--444, 1997.

\bibitem[Chi and Lange(2015)]{Eric-2015-Splitting}
E.~C. Chi and K.~Lange.
\newblock Splitting methods for convex clustering.
\newblock \emph{Journal of Computational and Graphical Statistics}, 24\penalty0
  (4):\penalty0 994--1013, 2015.

\bibitem[Drusvyatskiy and Lewis(2018)]{Drusvyatskiy-2018-Error}
D.~Drusvyatskiy and A.~S. Lewis.
\newblock Error bounds, quadratic growth, and linear convergence of proximal
  methods.
\newblock \emph{Mathematics of Operations Research}, 43\penalty0 (3):\penalty0
  919--948, 2018.

\bibitem[Fletcher(2005)]{Fletcher-2005-Barzilai}
R.~Fletcher.
\newblock On the {B}arzilai-{B}orwein method.
\newblock In \emph{Optimization and control with applications}, pages 235--256.
  Springer, 2005.

\bibitem[Gafni and Bertsekas(1984)]{Gafni-1984-Two}
E.~M. Gafni and D.~P. Bertsekas.
\newblock Two-metric projection methods for constrained optimization.
\newblock \emph{SIAM Journal on Control and Optimization}, 22\penalty0
  (6):\penalty0 936--964, 1984.

\bibitem[Guntuboyina et~al.(2020)Guntuboyina, Lieu, Chatterjee, and
  Sen]{Guntuboyina-2020-Adaptive}
A.~Guntuboyina, D.~Lieu, S.~Chatterjee, and B.~Sen.
\newblock Adaptive risk bounds in univariate total variation denoising and
  trend filtering.
\newblock \emph{Annals of Statistics}, 48\penalty0 (1):\penalty0 205--229,
  2020.

\bibitem[Hocking et~al.(2011)Hocking, Vert, Bach, and
  Joulin]{Hocking-2011-Clusteringpath}
T.~Hocking, J.-P. Vert, F.~Bach, and A.~Joulin.
\newblock Clusterpath: {A}n algorithm for clustering using convex fusion
  penalties.
\newblock In \emph{ICML}, 2011.

\bibitem[Hong and Luo(2017)]{Hong-2017-Linear}
M.~Hong and Z-Q. Luo.
\newblock On the linear convergence of the alternating direction method of
  multipliers.
\newblock \emph{Mathematical Programming}, 162\penalty0 (1-2):\penalty0
  165--199, 2017.

\bibitem[Jane et~al.(2021)Jane, Yuan, Zeng, and Zhang]{Jane-2021-Variational}
J.~Y. Jane, X.~Yuan, S.~Zeng, and J.~Zhang.
\newblock Variational analysis perspective on linear convergence of some first
  order methods for nonsmooth convex optimization problems.
\newblock \emph{Set-Valued and Variational Analysis}, pages 1--35, 2021.

\bibitem[Jourani(2000)]{Jourani-2000-Hoffman}
A.~Jourani.
\newblock Hoffman's error bound, local controllability, and sensitivity
  analysis.
\newblock \emph{SIAM Journal on Control and Optimization}, 38\penalty0
  (3):\penalty0 947--970, 2000.

\bibitem[Karimi et~al.(2016)Karimi, Nutini, and Schmidt]{Karimi-2016-Linear}
H.~Karimi, J.~Nutini, and M.~Schmidt.
\newblock Linear convergence of gradient and proximal-gradient methods under
  the {P}olyak-{L}ojasiewicz condition.
\newblock In \emph{Joint European Conference on Machine Learning and Knowledge
  Discovery in Databases}, pages 795--811. Springer, 2016.

\bibitem[Kim et~al.(2009)Kim, Koh, Boyd, and Gorinevsky]{Kim-2009-Ell1}
S-J. Kim, K.~Koh, S.~Boyd, and D.~Gorinevsky.
\newblock $\ell_1$ trend filtering.
\newblock \emph{SIAM Review}, 51\penalty0 (2):\penalty0 339--360, 2009.

\bibitem[Lan(2020)]{Lan-2020-First}
G.~Lan.
\newblock \emph{First-order and Stochastic Optimization Methods for Machine
  Learning}.
\newblock Springer Nature, 2020.

\bibitem[Lan and Zhou(2017)]{Lan-2017-Optimal}
G.~Lan and Y.~Zhou.
\newblock An optimal randomized incremental gradient method.
\newblock \emph{Mathematical Programming}, pages 1--49, 2017.

\bibitem[Leser(1961)]{Leser-1961-Simple}
C.~Leser.
\newblock A simple method of trend construction.
\newblock \emph{Journal of the Royal Statistical Society: Series B},
  23:\penalty0 91--107, 1961.

\bibitem[Lin et~al.(2017)Lin, Sharpnack, Rinaldo, and
  Tibshirani]{Lin-2017-Sharp}
K.~Lin, J.~L. Sharpnack, A.~Rinaldo, and R.~J. Tibshirani.
\newblock A sharp error analysis for the fused {L}asso, with application to
  approximate changepoint screening.
\newblock In \emph{NIPS}, 2017.

\bibitem[Loh and Wainwright(2015)]{Loh-2015-Regularized}
P-L. Loh and M.~J. Wainwright.
\newblock Regularized $m$-estimators with nonconvexity: Statistical and
  algorithmic theory for local optima.
\newblock \emph{The Journal of Machine Learning Research}, 16\penalty0
  (1):\penalty0 559--616, 2015.

\bibitem[Luo and Tseng(1992)]{Luo-1992-Linear}
Z-Q. Luo and P.~Tseng.
\newblock On the linear convergence of descent methods for convex essentially
  smooth minimization.
\newblock \emph{SIAM Journal on Control and Optimization}, 30\penalty0
  (2):\penalty0 408--425, 1992.

\bibitem[Luo and Tseng(1993)]{Luo-1993-Error}
Z-Q. Luo and P.~Tseng.
\newblock Error bounds and convergence analysis of feasible descent methods: a
  general approach.
\newblock \emph{Annals of Operations Research}, 46\penalty0 (1):\penalty0
  157--178, 1993.

\bibitem[Negahban and Wainwright(2012)]{Negahban-2012-Restricted}
S.~Negahban and M.~J. Wainwright.
\newblock Restricted strong convexity and weighted matrix completion: Optimal
  bounds with noise.
\newblock \emph{The Journal of Machine Learning Research}, 13\penalty0
  (1):\penalty0 1665--1697, 2012.

\bibitem[Negahban et~al.(2012)Negahban, Ravikumar, Wainwright, and
  Yu]{Negahban-2012-Unified}
S.~Negahban, P.~Ravikumar, M.~J. Wainwright, and B.~Yu.
\newblock A unified framework for high-dimensional analysis of $m$-estimators
  with decomposable regularizers.
\newblock \emph{Statistical Science}, 27\penalty0 (4):\penalty0 538--557, 2012.

\bibitem[Nesterov(2018)]{Nesterov-2018-Lectures}
Y.~Nesterov.
\newblock \emph{Lectures on Convex Optimization}, volume 137.
\newblock Springer, 2018.

\bibitem[Padilla et~al.(2018{\natexlab{a}})Padilla, Sharpnack, Chen, and
  Witten]{Padilla-2018-adaptive}
O.~H.~M. Padilla, J.~Sharpnack, Y.~Chen, and D.~M. Witten.
\newblock Adaptive non-parametric regression with the {K-NN} fused {L}asso.
\newblock \emph{Arxiv Preprint: 1807.11641}, 2018{\natexlab{a}}.

\bibitem[Padilla et~al.(2018{\natexlab{b}})Padilla, Sharpnack, Scott, and
  Tibshirani]{Padilla-2018-DFS}
O.~H.~M. Padilla, J.~Sharpnack, J.~G. Scott, and R.~J. Tibshirani.
\newblock The {DFS} fused lasso: {L}inear-time denoising over general graphs.
\newblock \emph{Journal of Machine Learning Research}, 18\penalty0
  (1):\penalty0 1–36, 2018{\natexlab{b}}.

\bibitem[Palaniappan and Bach(2016)]{Palaniappan-2016-Stochastic}
B.~Palaniappan and F.~Bach.
\newblock Stochastic variance reduction methods for saddle-point problems.
\newblock In \emph{NeurIPS}, pages 1416--1424, 2016.

\bibitem[Pang(1987)]{Pang-1987-Posteriori}
J-S. Pang.
\newblock A posteriori error bounds for the linearly-constrained variational
  inequality problem.
\newblock \emph{Mathematics of Operations Research}, 12\penalty0 (3):\penalty0
  474--484, 1987.

\bibitem[Pang(1997)]{Pang-1997-Error}
J-S. Pang.
\newblock Error bounds in mathematical programming.
\newblock \emph{Mathematical Programming}, 79\penalty0 (1):\penalty0 299--332,
  1997.

\bibitem[Parikh et~al.(2014)Parikh, Boyd, et~al.]{Parikh-2014-Proximal}
N.~Parikh, S.~Boyd, et~al.
\newblock Proximal algorithms.
\newblock \emph{Foundations and Trends{\textregistered} in Optimization},
  1\penalty0 (3):\penalty0 127--239, 2014.

\bibitem[Pena et~al.(2018)Pena, Vera, and Zuluaga]{Pena-2018-Algorithm}
J.~Pena, J.~Vera, and L.~Zuluaga.
\newblock An algorithm to compute the hoffman constant of a system of linear
  constraints.
\newblock \emph{ArXiv Preprint: 1804.08418}, 2018.

\bibitem[Radchenko and Mukherjee(2017)]{Radchenko-2017-Convex}
P.~Radchenko and G.~Mukherjee.
\newblock Convex clustering via $\ell_1$ fusion penalization.
\newblock \emph{Journal of the Royal Statistical Society: Series B},
  79\penalty0 (5):\penalty0 1527–1546, 2017.

\bibitem[Rakhlin et~al.(2012)Rakhlin, Shamir, and
  Sridharan]{Rakhlin-2012-Making}
A.~Rakhlin, O.~Shamir, and K.~Sridharan.
\newblock Making gradient descent optimal for strongly convex stochastic
  optimization.
\newblock In \emph{ICML}, pages 1571--1578, 2012.

\bibitem[Ramdas and Tibshirani(2016)]{Ramdas-2016-Fast}
A.~Ramdas and R.~J. Tibshirani.
\newblock Fast and flexible {ADMM} algorithms for trend filtering.
\newblock \emph{Journal of Computational and Graphical Statistics}, 25\penalty0
  (3):\penalty0 839--858, 2016.

\bibitem[Rockafellar(2015)]{Rockafellar-2015-Convex}
R.~T. Rockafellar.
\newblock \emph{Convex Analysis}.
\newblock Princeton University Press, 2015.

\bibitem[Rudin et~al.(1992)Rudin, Osher, and Faterni]{Rudin-1992-Nonlinear}
L.~I. Rudin, S.~Osher, and E.~Faterni.
\newblock Nonlinear total variation based noise removal algorithms.
\newblock \emph{Physica D: Nonlinear Phenomena}, 60:\penalty0 259--268, 1992.

\bibitem[Schmidt et~al.(2011)Schmidt, Roux, and Bach]{Schmidt-2011-Convergence}
M.~Schmidt, N.~L. Roux, and F.~Bach.
\newblock Convergence rates of inexact proximal-gradient methods for convex
  optimization.
\newblock In \emph{NIPS}, pages 1458--1466, 2011.

\bibitem[Sun et~al.(2021)Sun, Toh, and Yuan]{Sun-2021-Convex}
D.~Sun, K-C. Toh, and Y.~Yuan.
\newblock Convex clustering: Model, theoretical guarantee and efficient
  algorithm.
\newblock \emph{The Journal of Machine Learning Research}, 22:\penalty0 1--32,
  2021.

\bibitem[Tan and Witten(2015)]{Tan-2015-Statistical}
K.~M. Tan and D.~Witten.
\newblock Statistical properties of convex clustering.
\newblock \emph{Electronic Journal of Statistics}, 9\penalty0 (2):\penalty0
  2324--2347, 2015.

\bibitem[Tibshirani et~al.(2005)Tibshirani, Saunders, Rosset, Zhu, and
  Knight]{Tibshirani-2005-Sparsity}
R.~Tibshirani, M.~Saunders, S.~Rosset, J.~Zhu, and K.~Knight.
\newblock Sparsity and smoothness via the fused lasso.
\newblock \emph{Journal of the Royal Statistical Society: Series B},
  67\penalty0 (1):\penalty0 91--108, 2005.

\bibitem[Tibshirani(2014)]{Ryan-2014-Adaptive}
R.~J. Tibshirani.
\newblock Adaptive piecewise polynomial estimation via trend filtering.
\newblock \emph{Annals of Statistics}, 42\penalty0 (1):\penalty0 285--323,
  2014.

\bibitem[Tibshirani and Taylor(2011)]{Ryan-2011-The}
R.~J. Tibshirani and J.~Taylor.
\newblock The solution path of the generalized lasso.
\newblock \emph{Annals of Statistics}, 39\penalty0 (3):\penalty0 1335–1371,
  2011.

\bibitem[Tseng(1991)]{Tseng-1991-Applications}
P.~Tseng.
\newblock Applications of a splitting algorithm to decomposition in convex
  programming and variational inequalities.
\newblock \emph{SIAM Journal on Control and Optimization}, 29\penalty0
  (1):\penalty0 119--138, 1991.

\bibitem[Tseng(2000)]{Tseng-2000-Modified}
P.~Tseng.
\newblock A modified forward-backward splitting method for maximal monotone
  mappings.
\newblock \emph{SIAM Journal on Control and Optimization}, 38\penalty0
  (2):\penalty0 431--446, 2000.

\bibitem[Tseng(2010)]{Tseng-2010-Approximation}
P.~Tseng.
\newblock Approximation accuracy, gradient methods, and error bound for
  structured convex optimization.
\newblock \emph{Mathematical Programming}, 125\penalty0 (2):\penalty0 263--295,
  2010.

\bibitem[Wang and Xiao(2017)]{Wang-2017-Exploiting}
J.~Wang and L.~Xiao.
\newblock Exploiting strong convexity from data with primal-dual first-order
  algorithms.
\newblock In \emph{ICML}, pages 3694--3702, 2017.

\bibitem[Wang and Lin(2014)]{Wang-2014-Iteration}
P-W. Wang and C-J. Lin.
\newblock Iteration complexity of feasible descent methods for convex
  optimization.
\newblock \emph{The Journal of Machine Learning Research}, 15\penalty0
  (1):\penalty0 1523--1548, 2014.

\bibitem[Wang et~al.(2016{\natexlab{a}})Wang, Gong, Chang, Huang, and
  Zhou]{Wang-2016-Robust}
Q.~Wang, P.~Gong, S.~Chang, T.~S. Huang, and J.~Zhou.
\newblock Robust convex clustering analysis.
\newblock In \emph{ICDM}, pages 1263--1268. IEEE, 2016{\natexlab{a}}.

\bibitem[Wang et~al.(2016{\natexlab{b}})Wang, Sharpnack, Smola, and
  Tibshirani]{Wang-2016-Trend}
Y.~Wang, J.~Sharpnack, A.~J. Smola, and R.~J. Tibshirani.
\newblock Trend filtering on graphs.
\newblock \emph{The Journal of Machine Learning Research}, 17\penalty0
  (1):\penalty0 3651--3691, 2016{\natexlab{b}}.

\bibitem[Wang et~al.(2014)Wang, Liu, and Zhang]{Wang-2014-Optimal}
Z.~Wang, H.~Liu, and T.~Zhang.
\newblock Optimal computational and statistical rates of convergence for sparse
  nonconvex learning problems.
\newblock \emph{Annals of Statistics}, 42\penalty0 (6):\penalty0 2164--2201,
  2014.

\bibitem[Wu et~al.(2016)Wu, Kwon, Shen, and Pan]{Wu-2017-New}
C.~Wu, S.~Kwon, X.~Shen, and W.~Pan.
\newblock A new algorithm and theory for penalized regression-based clustering.
\newblock \emph{The Journal of Machine Learning Research}, 17:\penalty0 1–25,
  2016.

\bibitem[Zhang and Xiao(2017)]{Zhang-2017-Stochastic}
Y.~Zhang and L.~Xiao.
\newblock Stochastic primal-dual coordinate method for regularized empirical
  risk minimization.
\newblock \emph{The Journal of Machine Learning Research}, 18\penalty0
  (1):\penalty0 2939--2980, 2017.

\bibitem[Zhou and So(2017)]{Zhou-2017-Unified}
Z.~Zhou and A.~M-C. So.
\newblock A unified approach to error bounds for structured convex optimization
  problems.
\newblock \emph{Mathematical Programming}, 165\penalty0 (2):\penalty0 689--728,
  2017.

\bibitem[Zhou et~al.(2015)Zhou, Zhang, and So]{Zhou-2015-Error}
Z.~Zhou, Q.~Zhang, and A.~M-C. So.
\newblock $\ell_{1,p}$-norm regularization: error bounds and convergence rate
  analysis of first-order methods.
\newblock In \emph{ICML}, pages 1501--1510. JMLR. org, 2015.

\bibitem[Zhu et~al.(2014)Zhu, Xu, Leng, and Yan]{Zhu-2014-Convex}
C.~Zhu, H.~Xu, C.~Leng, and S.~Yan.
\newblock Convex optimization procedure for clustering: Theoretical revisit.
\newblock In \emph{NIPS}, 2014.

\end{thebibliography}

\newpage\onecolumn
\appendix
\section{Postponed Proofs in Section~\ref{sec:GEB}}
This section first lays out the detailed proofs for Lemma~\ref{Lemma:conjugate}, Lemma~\ref{Lemma:minimizer} and Lemma~\ref{Lemma:dual-objective}.  Then, we introduce an \textit{upper Lipschitz continuity} (ULC) property of a set-valued mapping which suffices to guarantee the GEB condition via appeal to the techniques in~\citet{Zhou-2015-Error}. Finally, we provide the detailed proofs for Theorem~\ref{Theorem:GEB-polyhedron} and Theorem~\ref{Theorem:GEB-other}. 

\subsection{Proof of Lemma~\ref{Lemma:conjugate}}
We first show that $\alpha \in \partial f(\beta) \Leftrightarrow \beta \in \partial f^\star(\alpha)$. Indeed, if $\alpha \in \partial f(\beta)$, we have $f^\star(\alpha) = \alpha^\top \beta - f(\beta)$. This implies that $\beta \in \partial f^\star(\alpha)$ since, for all $\alpha' \in \br^m$, the following inequality holds true, 
\begin{equation*}
f^\star(\alpha') - f^\star(\alpha) \geq (\alpha')^\top \beta - f(\beta) - (\alpha^\top \beta - f(\beta)) = (\alpha' - \alpha)^\top \beta. 
\end{equation*}
Conversely, if $\beta \in \partial f^\star(\alpha)$, the above argument implies that $\alpha \in \partial f^{\star\star}(\beta)$. Note that $f$ is proper and convex.~\citet[Theorem~12.2]{Rockafellar-2015-Convex} shows that $f = f^{\star\star}$ and $\alpha \in \partial f(\beta)$. 

We are ready to prove that $f^\star$ is $\frac{1}{\mu}$-gradient Lipschitz. Indeed, since $f$ is $\mu$-strongly convex and differentiable, the gradient mapping $\nabla f$ is one-to-one and $\partial f(\beta)$ is a singleton. This implies that $\partial f^\star(\alpha) = \{\nabla f^\star(\alpha)\}$ and
\begin{equation*}
\left\|\nabla f^\star(\alpha_1) - \nabla f^\star(\alpha_2)\right\| = \|\beta_1 - \beta_2\| \leq \frac{\|\nabla f(\beta_1) - \nabla f(\beta_2)\|}{\mu} = \frac{\|\alpha_1 - \alpha_2\|}{\mu}.  
\end{equation*} 
By similar arguments, we obtain that $f^\star$ is $\frac{1}{\ell}$-strongly convex. Putting these pieces yields the desired result. 

\subsection{Proof of Lemma~\ref{Lemma:minimizer}}
Since $f$ is $\mu$-strongly convex, $\beta^\star(\alpha)$ is uniquely determined given $\alpha \in \BB_q^n$ and hence well-defined. We see from the optimality condition that $\nabla f(\beta^\star(\alpha)) = \lambda D^\top\alpha$. Therefore, the following inequality holds true for any $\alpha', \alpha \in \BB_q^n$:
\begin{equation*}
\|\beta^\star(\alpha') - \beta^\star(\alpha)\| \leq \frac{\|\nabla f(\beta^\star(\alpha')) - \nabla f(\beta^\star(\alpha))\|}{\mu} \leq \frac{\lambda\|D^\top\alpha' - D^\top\alpha\|}{\mu} \leq \left(\frac{\lambda\|D\|}{\mu}\right)\|\alpha' - \alpha\|. 
\end{equation*}
Putting these pieces together yields the desired result.  

\subsection{Proof of Lemma~\ref{Lemma:dual-objective}}
We see from the definition of $\bar{f}$ that $\bar{f}(\alpha) = f^\star(\lambda D^\top\alpha)$. Lemma~\ref{Lemma:conjugate} shows that $f^\star$ is smooth and strongly convex under Assumption~\ref{Assumption:main}. Therefore, we obtain that $\bar{f}$ is differentiable and $\nabla \bar{f}(\alpha) = \lambda D \nabla f^\star(\lambda D^\top\alpha)$ using the chain rule. Since $\nabla f(\beta^\star(\alpha)) = \lambda D^\top\alpha$ and $\alpha \in \partial f(\beta) \Leftrightarrow \beta \in \partial f^\star(\alpha)$ (cf. the proof of Lemma~\ref{Lemma:conjugate}), we have $\beta^\star(\alpha) = \nabla f^\star(\lambda D^\top\alpha)$. Putting these pieces together yields that $\nabla \bar{f}(\alpha) = \lambda D\beta^\star(\alpha)$. For all $\alpha', \alpha \in \BB_q$, we have
\begin{equation*}
\|\nabla\bar{f}(\alpha') - \nabla\bar{f}(\alpha)\| \leq \|\lambda D(\nabla f^\star(\lambda D^\top\alpha') - \nabla f^\star(\lambda D^\top\alpha_2))\| \leq \left(\frac{\lambda^2\|D\|^2}{\mu}\right)\|\alpha_1 - \alpha_2\|. 
\end{equation*}
Putting these pieces together yields the desired result.  

\subsection{ULC property}
We start with some mathematical notions. Indeed, let $\YCal$ and $\ZCal$ be Euclidean spaces. A mapping $\Gamma: \YCal \mapsto \ZCal$ is said to be a \textit{set-valued mapping}, or equivalently, a multifunction if for any $y \in \YCal$, we have $\Gamma(y) \subseteq \ZCal$. The graph of $\Gamma$ is defined by $\{(y, z) \in \YCal \times \ZCal \mid z \in \Gamma(y)\}$. We now define a notion of upper Lipschitz continuity (ULC). 
\begin{definition}
A set-valued mapping $\Gamma: \YCal \mapsto \ZCal$ has the ULC property at $y \in \YCal$ if $\Gamma(y)$ is nonempty and closed, and there exist constants $\kappa > 0$ and $\delta > 0$ such that for any $y \in \YCal$ with $\|y' - y\| \leq \delta$, we have $\Gamma(y') \subseteq \Gamma(y) + \kappa\|y' - y\|\BB_2$ where $\BB_2$ is the unit $\ell_2$-norm ball in $\ZCal$ and ``+" is the Minkowski sum of two sets. 
\end{definition}
Proceeding a further step, we opt to prove a sufficient condition which guarantees that the dual filtering-clustering model in Eq.~\eqref{prob:dual-main} satisfies a GEB condition. Let $\Sigma(\cdot, \cdot)$ be the set-valued mapping given by 
\begin{equation}\label{def:mapping-opt}
\Sigma(\s, \g) \mydefn \{\alpha \in \BB_q^n \mid D^\top \alpha = \s, -\g \in \NCal_{\BB_q^n}(\alpha)\}, \quad \textnormal{for all } (\s, \g) \in \br^d \times \br^{mn}. 
\end{equation}
Before proceeding to our main result, we summarize the relationship between the set-valued mapping $\Sigma$ and the optimal set $\Omega^\star$ in the following proposition. 
\begin{proposition}\label{Prop:set-map}
Under Assumption~\ref{Assumption:main}, we have $\Omega^\star = \Sigma(\s^\star, \g^\star)$ for a pair $(\s^\star, \g^\star)$ satisfying that $\s^\star = D^\top\alpha^\star$ and $\g^\star = \nabla\bar{f}(\alpha^\star)$ for all $\alpha^\star \in \Omega^\star$. 
\end{proposition}
\begin{proof}
Since the dual filtering-clustering model in Eq.~\eqref{prob:dual-main} is a convex optimization problem, the first-order optimality condition is both necessary and sufficient. Therefore, we have
\begin{equation}\label{condition:set-map}
\Omega^\star = \{\alpha^\star \in \br^{mn} \mid 0 \in \nabla \bar{f}(\alpha^\star) + \NCal_{\BB_q^n}(\alpha^\star)\}. 
\end{equation}
We first show that $\Omega^\star \subseteq \Sigma(\s^\star, \g^\star)$. Let $\alpha^\star \in \Omega^\star$, we notice that $\bar{f}(\alpha)=f^\star(\lambda D^\top\alpha)$ and $f^\star$ is strongly convex (cf. Lemma~\ref{Lemma:conjugate}). This implies that $\lambda D^\top\alpha^\star$ remains the same for all $\alpha^\star \in \Omega^\star$ and $\s^\star = D^\top\alpha^\star$ is well defined. In addition, $\g^\star = \nabla\bar{f}(\alpha^\star) = \lambda D\nabla f^\star(\lambda D^\top\alpha) = \lambda D\nabla f^\star(\lambda\s^\star)$. This implies that $\g^\star$ is well defined. Since $0 \in \nabla \bar{f}(\alpha^\star) + \NCal_{\BB_q^n}(\alpha^\star)$, we have $-\g^\star \in \NCal_{\BB_q^n}(\alpha^\star)$. Therefore, we have $\alpha^\star \in \Sigma(\s^\star, \g^\star)$. 

Conversely, let $\alpha^\star \in \Sigma(\s^\star, \g^\star)$, we have $\s^\star = D^\top \alpha^\star$ and $-\g^\star \in \NCal_{\BB_q^n}(\alpha^\star)$. In addition, $\g^\star = \nabla \bar{f}(\alpha^\star)$. Therefore, we conclude from Eq.~\eqref{condition:set-map} that $\alpha^\star \in \Omega^\star$.
\end{proof}
Given the result of Proposition~\ref{Prop:set-map}, we prove that the ULC property of $\Sigma$ guarantees that the GEB condition holds for the dual filtering-clustering model in Eq.~\eqref{prob:dual-main}. 
\begin{theorem}\label{Theorem:GEB-sufficient}
Under Assumption~\ref{Assumption:main}, the GEB condition holds for the dual filtering-clustering model in Eq.~\eqref{prob:dual-main} if the mapping $\Sigma$ in Eq.~\eqref{def:mapping-opt} has the ULC property at $(\s^\star, \g^\star)$ where a pair $(\s^\star, \g^\star)$ is given in Proposition~\ref{Prop:set-map}.
\end{theorem}
\begin{proof}
Since $\Sigma$ has the ULC property at $(\s^\star, \g^\star)$, there exist constants $\kappa > 0$ and $\delta > 0$ such that for all $(\s, \g) \in \br^d \times \br^{mn}$ with $\|(\s, \g) - (\s^\star, \g^\star)\| \leq \delta$, we have
\begin{equation}\label{inequality:GEBsuff-first}
\Sigma(\s, \g) \subseteq \Sigma(\s^\star, \g^\star) + \kappa\|(\s, \g) - (\s^\star, \g^\star)\|\BB_2. 
\end{equation}
We recall the residual function $R(\alpha) = \PCal_{\BB_q^n}(\alpha - \nabla \bar{f}(\alpha)) - \alpha$ in Eq.~\eqref{def:residue} and define two functions $\s^+: \BB_q^n \rightarrow \br^d$ and $\g^+: \BB_q^n \rightarrow \br^{mn}$ by 
\begin{equation}\label{inequality:GEBsuff-second}
\s^+(\alpha) \mydefn D^\top(\alpha + R(\alpha)), \quad \g^+(\alpha) \mydefn \nabla \bar{f}(\alpha) + R(\alpha). 
\end{equation}
Since $\BB_q^n$ is convex and bounded, $R(\cdot)$ is Lipschitz continuous~\citep{Rockafellar-2015-Convex}. Combining it with the Lipschitz continuity of $\nabla \bar{f}$ implies that $\s^+(\cdot)$ and $\g^+(\cdot)$ are Lipschitz continuous. This together with Proposition~\ref{Prop:set-map} implies that there exists a constant $\rho > 0$ such that, 
\begin{equation}\label{inequality:GEBsuff-third}
\|(\s^+(\alpha), \g^+(\alpha)) - (\s^\star, \g^\star)\| \leq \delta, \quad \textnormal{for all } \alpha \in \BB_q^n \cap \{d(\alpha, \Omega^\star) \leq \rho\}.  
\end{equation}
By the definition of the residual function $R$, we have
\begin{equation*}
\alpha + R(\alpha) = \argmin_{z \in \BB_q^n} \ z^\top \nabla \bar{f}(\alpha) + \frac{1}{2}\|z - \alpha\|^2. 
\end{equation*}
By the optimality condition, we have
\begin{equation}\label{inequality:GEBsuff-fourth}
- \nabla \bar{f}(\alpha) - R(\alpha) \in \NCal_{\BB_q^n}(\alpha + R(\alpha)). 
\end{equation}
Combining Eq.~\eqref{inequality:GEBsuff-second} and Eq.~\eqref{inequality:GEBsuff-fourth} yields that $\alpha + R(\alpha) \in \Sigma(\s^+(\alpha), \g^+(\alpha))$ for all $\alpha \in \BB_q^n$. This together with Eq.~\eqref{inequality:GEBsuff-first} and Eq.~\eqref{inequality:GEBsuff-third} yields that 
\begin{equation*}
d(\alpha + R(\alpha), \Sigma(\s^\star, \g^\star)) \leq \kappa\|(\s^+(\alpha), \g^+(\alpha)) - (\s^\star, \g^\star)\|, \quad \textnormal{for all } \alpha \in \BB_q^n \cap \{d(\alpha, \Omega^\star) \leq \rho\}. 
\end{equation*}
Recalling that $\nabla\bar{f}(\alpha) = \lambda D\nabla f^\star(\lambda D^\top \alpha)$ and $\g^\star = \lambda D\nabla f^\star(\lambda \s^\star)$, we obtain from the definition of $\s^+(\cdot)$ and $\g^+(\cdot)$ in Eq.~\eqref{inequality:GEBsuff-second} that 
\begin{eqnarray*}
\|\s^+(\alpha) - \s^\star\| & \leq & \|D^\top \alpha - \s^\star\| + \|D\|\|R(\alpha)\|, \\
\|\g^+(\alpha) - \g^\star\| & \leq & \left(\frac{\lambda^2\|D\|}{\mu}\right)\|D^\top \alpha - \s^\star\| + \|R(\alpha)\|. 
\end{eqnarray*}
In view of Proposition~\ref{Prop:set-map}, we have $d(\alpha, \Omega^\star) \leq d(\alpha + R(\alpha), \Sigma(\s^\star, \g^\star)) + \|R(\alpha)\|$. Putting these pieces together yields that, for all $\alpha \in \BB_q^n \cap \{d(\alpha, \Omega^\star) \leq \rho\}$, we have
\begin{equation*}
d(\alpha, \Omega^\star) \leq \left(\kappa + \frac{\lambda^2\|D\|\kappa}{\mu}\right)\|D^\top \alpha - \s^\star\| + (\kappa\|D\| + \kappa + 1)\|R(\alpha)\|. 
\end{equation*}
Letting $\kappa_0 = \max\{\kappa + \frac{\lambda^2\|D\|\kappa}{\mu}, \kappa\|D\| + \kappa + 1\}$ and using the inequality $(a+b)^2 \leq 2(a^2 + b^2)$ yields that, for all $\alpha \in \BB_q^n \cap \{d(\alpha, \Omega^\star) \leq \rho\}$, we have
\begin{equation}\label{inequality:GEBsuff-fifth}
d^2(\alpha, \Omega^\star) \leq 2\kappa_0^2(\|D^\top \alpha - \s^\star\|^2 + \|R(\alpha)\|^2). 
\end{equation}
Since $f^\star$ is $\frac{1}{\ell}$-strongly convex, we have
\begin{equation*}
\|D^\top \alpha - \s^\star\|^2 = \frac{\|\lambda D^\top \alpha - \lambda\s^\star\|^2}{\lambda^2} \leq \frac{\ell}{\lambda^2}(\lambda D^\top \alpha - \lambda\s^\star)^\top(\nabla f^\star(\lambda D^\top \alpha) - \nabla f^\star(\lambda\s^\star)). 
\end{equation*}
Let $\alpha^\star$ be the projection of $\alpha$ onto $\Omega^\star$, we have $\s^\star = D^\top \alpha^\star$. It also follows from the definition that $\g^\star = \nabla\bar{f}(\alpha^\star) = \lambda D\nabla f^\star(\lambda\s^\star)$. Then, we have 
\begin{equation}\label{inequality:GEBsuff-sixth}
\|D^\top \alpha - \s^\star\|^2 \leq \frac{\ell}{\lambda^2} (\alpha - \alpha^\star)^\top(\nabla \bar{f}(\alpha) - \g^\star). 
\end{equation}
For all $u \in \NCal_{\BB_q^n}(\alpha + R(\alpha))$ and $v \in \NCal_{\BB_q^n}(\alpha^\star)$, we obtain from the definition of the normal cone that $(u - v)^\top(\alpha + R(\alpha) - \alpha^\star) \geq 0$. Plugging $u = - \nabla\bar{f}(\alpha) - R(\alpha)$ and $v = -\g^\star$ into the above inequality yields that 
\begin{equation*}
(\alpha - \alpha^\star)^\top(\nabla\bar{f}(\alpha) - \g^\star) + \|R(\alpha)\|^2 \leq (\g^\star - \nabla\bar{f}(\alpha) + \alpha^\star - \alpha)^\top R(\alpha). 
\end{equation*}
Since $\g^\star = \nabla\bar{f}(\alpha^\star)$ and Lemma~\ref{Lemma:dual-objective} implies that $\nabla\bar{f}$ is Lipschitz continuous, there exists a constant $\kappa_1 > 0$ such that
\begin{equation*}
(\g^\star - \nabla\bar{f}(\alpha) + \alpha^\star - \alpha)^\top R(\alpha) \leq \kappa_1\|\alpha - \alpha^\star\|\|R(\alpha)\|. 
\end{equation*}
Since $\|R(\alpha)\|^2 \geq 0$, we have 
\begin{equation}\label{inequality:GEBsuff-seventh}
(\alpha - \alpha^\star)^\top(\nabla\bar{f}(\alpha) - \g^\star) \leq \kappa_1\|\alpha - \alpha^\star\|\|R(\alpha)\|. 
\end{equation}
Combining Eq.~\eqref{inequality:GEBsuff-fifth}, Eq.~\eqref{inequality:GEBsuff-sixth} and Eq.~\eqref{inequality:GEBsuff-seventh} yields that there exists a constant $\kappa_2 > 0$ such that the following inequality holds true, 
\begin{equation*}
d^2(\alpha, \Omega^\star) \leq \kappa_2(\|\alpha - \alpha^\star\|\|R(\alpha)\| + \|R(\alpha)\|^2), \quad \textnormal{for all } \alpha \in \BB_q^n \cap \{d(\alpha, \Omega^\star) \leq \rho\}.
\end{equation*}
Notice that $d(\alpha, \Omega^\star) = \|\alpha - \alpha^\star\|$. Thus, by solving the above quadratic inequality, we obtain that there exists a constant $\kappa_3 > 0$ such that 
\begin{equation*}
d(\alpha, \Omega^\star) \leq \kappa_3\|R(\alpha)\|, \quad \textnormal{for all } \alpha \in \BB_q^n \cap \{d(\alpha, \Omega^\star) \leq \rho\}.
\end{equation*}
Since $R(\alpha) = 0$ if and only if $d(\alpha, \Omega^\star) = 0$, the function $h(\alpha) = \frac{d(\alpha, \Omega^\star)}{\|R(\alpha)\|}$ is well defined and continuous over the domain $\BB_q^n \cap \{d(\alpha, \Omega^\star) > \rho\}$. Since $\BB_q^n$ is convex and bounded, there exists a constant $\kappa_4> 0$ such that $h(\alpha) \leq \kappa_4$ for all $\alpha \in \BB_q^n \cap \{d(\alpha, \Omega^\star) > \rho\}$. Equivalently, we have 
\begin{equation*}
d(\alpha, \Omega^\star) \leq \kappa_4\|R(\alpha)\|, \quad \textnormal{for all } \alpha \in \BB_q^n \cap \{d(\alpha, \Omega^\star) > \rho\}.
\end{equation*}
Setting $\tau = \max\{\kappa_3, \kappa_4\} > 0$, we obtain that $d(\alpha, \Omega^\star) \leq \tau\|R(\alpha)\|$ for all $\alpha \in \BB_q^n$. Therefore, the GEB condition holds for the dual filtering-clustering model in Eq.~\eqref{prob:dual-main}. 
\end{proof}
Equipped with the result of Theorem~\ref{Theorem:GEB-sufficient}, we will prove that the GEB condition holds for the dual filtering-clustering model in Eq.~\eqref{prob:dual-main} when $q \in [1, 2] \cup \{+\infty\}$. 

\subsection{Proof of Theorem~\ref{Theorem:GEB-polyhedron}}
We show that $\Sigma$ has the ULC property when $q=1$ or $q=+\infty$. 
\begin{lemma}\label{Lemma:GEB-polyhedron}
Under Assumption~\ref{Assumption:main} and let $q \in \{1, +\infty\}$, the set-valued mapping $\Sigma$ is a polyhedral multifunction. 
\end{lemma}
\begin{proof}
Since $q \in \left\{1, +\infty\right\}$, we have $\BB_q^n$ is a polyhedron which implies that its indicator function has a polyhedral epigraph. By the definition of the normal cone, we have $\NCal_{\BB_q^n}$ is the subdifferential of an indicator function of $\BB_q^n$. Putting these pieces together with~\citet[Lemma~2]{Zhou-2015-Error} yields that $\Sigma$ is a polyhedral multifunction. 
\end{proof}
With Theorem~\ref{Theorem:GEB-sufficient} and Lemma~\ref{Lemma:GEB-polyhedron} at hand, we can prove that the GEB condition holds for the dual filtering-clustering model in Eq.~\eqref{prob:dual-main}. Indeed, by Assumption~\ref{Assumption:main}, we have $\Omega^\star$ is an nonempty set. Proposition~\ref{Prop:set-map} guarantees that there exists a pair $(\s^\star, \g^\star) \in \br^d \times \br^{mn}$ such that $\Omega^\star = \Sigma(\s^\star, \g^\star)$ where the set-valued mapping $\Sigma$ is given by Eq.~\eqref{def:mapping-opt}. By Lemma~\ref{Lemma:GEB-polyhedron}, the set-valued mapping $\Sigma$ is a polyhedral multifunction. Putting these pieces together with~\citet[Lemma~1]{Zhou-2015-Error} implies that $\Sigma$ has the ULC property at $(\s^\star, \g^\star)$. Therefore, we conclude from Theorem~\ref{Theorem:GEB-sufficient} that the GEB condition holds for the dual filtering-clustering model in Eq.~\eqref{prob:dual-main}. 

\subsection{Proof of Theorem~\ref{Theorem:GEB-other}}
It suffices to show that $\Sigma$ has the ULC property when $q \in (1, 2]$ under certain conditions. Our first lemma provides the linear regularity of a collection of polyhedral sets. It has been stated as~\citet[Corollary~5.26]{Bauschke-1996-Projection} and we omit the proof here. 
\begin{lemma}\label{Lemma:regularity}
Suppose that $\SCal_1, \ldots, \SCal_M$ are a collection of polyhedra in $\br^{mn}$. Then, there exists a constant $\bar{\kappa} > 0$ such that $d(\alpha, \cap_{i=1}^M \SCal_i) \leq \bar{\kappa} \sum_{i=1}^M d(\alpha, \SCal_i)$ for all $\alpha \in \br^{mn}$. 
\end{lemma}
The next proposition provides a concrete representation of $\Omega^\star = \Sigma(\s^\star, \g^\star)$. In particular, we let $\JCal = \{j \in [n] \mid \g_j^\star \neq 0\}$ be the set of indices of nonzero coordinates. 
\begin{proposition}\label{Prop:opt-structure}
Under Assumption~\ref{Assumption:main} and let the mapping $\Sigma$ be defined in Eq.~\eqref{def:mapping-opt} and $(\s^\star, \g^\star) \in \br^d \times \br^{mn}$ be given in Proposition~\ref{Prop:set-map}. Then, we have
\begin{equation*}
\Sigma(\s^\star, \g^\star) = \left\{\alpha^\star = \begin{bmatrix} \alpha_1^\star \\ \vdots \\ \alpha_n^\star \end{bmatrix} \in \br^{mn} \left| \begin{array}{l} D^\top\alpha^\star = \s^\star, \\ \alpha_j^\star = -\frac{v(\g_j^\star)}{\|v(\g_j^\star)\|_q}, \textnormal{ for all } j \in \JCal, \\ \alpha_j^\star \in \textnormal{int}(\BB_q^n), \textnormal{ for all } j \notin \JCal.
\end{array} \right. \right\},   
\end{equation*}
where the function $v: \br^m \mapsto \br^m$ is defined by $v(\g) = (\sign(g_1)|g_1|^{\frac{p}{q}}, \ldots, \sign(g_m)|g_m|^{\frac{p}{q}})$. Fixing any $q \in (1, 2]$, there exists some constants $\delta > 0$ and $\nu > 0$ such that, for all $\g \in \br^m$ satisfying that $\|\g - \g^\star\| \leq \delta$, we have $\|v(\g) - v(\g^\star)\| \leq \nu\|\g - \g^\star\|$. 
\end{proposition}
\begin{proof}
By the definition of $(\s^\star, \g^\star) \in \br^d \times \br^{mn}$ in Proposition~\ref{Prop:set-map}, we have $D^\top\alpha^\star = \s^\star$, $\nabla\bar{f}(\alpha^\star) = \g^\star$ and $\alpha^\star \in \BB_q^n$. Then, we consider two different cases: $\g_j^\star = 0$ and $\g_j^\star \neq 0$. 

If $\g_j^\star = 0$, the $j$-th block of $\alpha^\star$ is in the interior of $\BB_q^n$ and $\alpha_j^\star \in \textnormal{int}(\BB_q^n)$ for all $j \notin \JCal$. 

If $\g_j^\star \neq \textbf{0}$,  the $j$-th block of $\alpha^\star$ is in the interior of $\BB_q^n$ and $\|\alpha_j^\star\|_q = 1$ for all $j \in \JCal$. By the KKT condition in constrained optimization, $\alpha^\star \in \Omega^\star$ if and only if there exists a multiplier $\mu \geq 0$ such that the following statement holds true, 
\begin{eqnarray*}
1 - \|\alpha_j^\star\|_q & \geq 	& 0, \\
\s^\star - D^\top\alpha^\star & = & 0, \\ 
\g_j^\star + \mu\cdot\frac{(\sign((\alpha_j^\star)_1)|(\alpha_j^\star)_1|^{q-1}, \ldots, \sign((\alpha_j^\star)_m)|(\alpha_j^\star)_m|^{q-1})}{\|\alpha_j^\star\|_q^{q/p}} & = & 0, \\
\mu(1 - \|\alpha_j^\star\|_q) & = & 0. 
\end{eqnarray*}
Since $\g_j^* \neq \textbf{0}$, we have $\mu > 0$ and again $\|\alpha_j^\star\|_q = 1$. Thus, we can solve $\alpha_j^\star$ in terms of $\g_j^\star$ by using the above equations and obtain that 
\begin{equation*}
\s^\star - D^\top\alpha^\star = 0, \quad \alpha_j^\star + \frac{v(\g_j^\star)}{\|v(\g_j^\star)\|_q} = 0. 
\end{equation*}
Putting these pieces together yields that 
\begin{equation*}
\Sigma(\s^\star, \g^\star) = \left\{\alpha^\star = \begin{bmatrix} \alpha_1^\star \\ \vdots \\ \alpha_n^\star \end{bmatrix} \in \br^{mn} \left| \begin{array}{l} D^\top\alpha^\star = \s^\star, \\ \alpha_j^\star = -\frac{v(\g_j^\star)}{\|v(\g_j^\star)\|_q}, \textnormal{ for all } j \in \JCal, \\ \alpha_j^\star \in \textnormal{int}(\BB_q^n), \textnormal{ for all } j \notin \JCal.
\end{array} \right. \right\}.    
\end{equation*}
Fixing any $q \in (1, 2]$, we have $\frac{p}{q} \geq 1$. Then, the function $s \mapsto \sign(s)|s|^{\frac{p}{q}}$ is continuously differentiable and hence locally Lipschitz~\citep{Rockafellar-2015-Convex}. So there exist some constants $\delta > 0$ and $\nu > 0$ such that, for all $s', s \in \br$ satisfying that $|s' - s| \leq \delta$, we have
\begin{equation*}
|\sign(s')|s'|^{\frac{p}{q}} - \sign(s)|s|^{\frac{p}{q}}| \leq \nu|s_1 - s_2|. 
\end{equation*}
By the definition of $v$, we have $\|v(\g) - v(\g^\star)\| \leq \nu\|\g - \g^\star\|$ for all $g \in \br^m$. As a consequence, we reach the conclusion of the proposition.
\end{proof}
Proposition~\ref{Prop:opt-structure} shows that $\Sigma(\s^\star, \g^\star)$ is closed. Since $\Sigma(\s^\star, \g^\star) \subseteq \BB_q^n$, we have $\Sigma(\s^\star, \g^\star)$ is bounded where $(\s^\star, \g^\star) \in \br^d \times \br^{mn}$ is given in Proposition~\ref{Prop:set-map}.  Now we are ready to prove that $\Sigma$ has the ULC property when $q \in (1, 2]$ and $\JCal = [n]$. 

By Assumption~\ref{Assumption:main}, we have $\Omega^\star$ is an nonempty set. Proposition~\ref{Prop:set-map} guarantees that there exists a pair $(\s^\star, \g^\star) \in \br^d \times \br^{mn}$ such that $\Omega^\star = \Sigma(\s^\star, \g^\star)$ where the set-valued mapping $\Sigma$ is given by Eq.~\eqref{def:mapping-opt}. 

Define the sets $\SCal$ and $\SCal_1, \SCal_2, \ldots, \SCal_n$ by 
\begin{equation*}
\SCal \mydefn \{\alpha^\star \in \br^{mn} \mid D^\top\alpha^\star = \s^\star\}, \quad \SCal_j \mydefn \left\{\alpha^\star = \begin{bmatrix} \alpha_1^\star \\ \vdots \\ \alpha_n^\star \end{bmatrix} \in \br^{mn} \Big\vert \begin{array}{l} \alpha_j^\star = -\frac{v(\g_j^\star)}{\|v(\g_j^\star)\|_q} \end{array} \right\}. 
\end{equation*}
By Proposition~\ref{Prop:opt-structure} and using the assumption that $\JCal = [n]$, we have
\begin{equation*}
\Sigma(\s^\star, \g^\star) = \SCal \cap (\cap_{j=1}^n \SCal_j).
\end{equation*}
Moreover, $\SCal, \SCal_1, \ldots, \SCal_n$ are all polyhedral subsets of $\br^{mn}$. By Lemma~\ref{Lemma:regularity}, there exists a constant $\bar{\kappa} > 0$ such that, for any $\alpha \in \br^{mn}$, we have
\begin{equation}\label{inequality:GEB-other-first}
d(\alpha, \Sigma(\s^\star, \g^\star)) \leq \bar{\kappa}\left(d(\alpha, \SCal) + \sum_{j=1}^n d(\alpha, \SCal_j)\right). 
\end{equation}
Thus, it suffices to bound the right-hand side of the above inequality for all $\alpha \in \Sigma(\s, \g)$ in which $(\s, \g)$ lies in the neighborhood of $(\s^\star, \g^\star) \in \br^d \times \br^{mn}$ and $\Sigma(\s, \g)$ is nonempty. Towards that end, we discuss the bound on $d(\alpha, \SCal)$ and $d(\alpha, \SCal_j)$ separately. 

The bound on $d(\alpha, \SCal)$ follows from the well-known Hoffman bound (see~\citet{Jourani-2000-Hoffman} for example). In particular, there exists a constant $\theta > 0$ such that $d(\alpha, \SCal) \leq \theta\|D^\top\alpha - \s^\star\|$ for all $\alpha \in \br^{mn}$. In addition, for all $\alpha \in \Sigma(\s, \g)$ with $\g \neq 0$, we have $D^\top\alpha = \s$. Putting these pieces together yields that 
\begin{equation}\label{inequality:GEB-other-second}
d(\alpha, \SCal) \leq \theta\|\s - \s^\star\|, \quad \textnormal{for all } \alpha \in \Sigma(\s, \g). 
\end{equation}
Proceeding a further step, we bound $d(\alpha, \SCal_j)$ via appeal to H\"{o}lder inequality. In particular, we notice that $\g_j^\star \neq 0$ for all $j \in [n]$. Therefore, there exists a constant $\delta_j > 0$ such that $\|(\s, \g) - (\s^\star, \g^\star)\| \leq \delta_j$ and $\g_j \neq 0$ for all $j \in [n]$. For any $\alpha \in \Sigma(\s, \g)$ satisfying that $\|(\s, \g) - (\s^\star, \g^\star)\| \leq \delta_j$, we have
\begin{equation*}
\s - D^\top\alpha = 0, \quad \alpha_j + \frac{v(\g_j)}{\|v(\g_j)\|_q} = 0. 
\end{equation*}
Since $\g_j, \g_j^\star \neq 0$, we have $\|v(\g_j)\|_q > 0$ and $\|v(\g_j^\star)\|_q > 0$. Putting these pieces together with Proposition~\ref{Prop:opt-structure}, we have
\begin{equation*}
d(\alpha, \SCal_j) \leq \left\|\frac{v(\g_j)}{\|v(\g_j)\|_q} - \frac{v(\g_j^\star)}{\|v(\g_j^\star)\|_q}\right\|,  
\end{equation*}
which implies that 
\begin{equation*}
\left\|\frac{v(\g_j)}{\|v(\g_j)\|_q} - \frac{v(\g_j^\star)}{\|v(\g_j^\star)\|_q}\right\| = \frac{\|v(\g_j) - v(\g_j^\star)\|_q\|v(\g_j)\| + \|v(\g_j) - v(\g_j^\star)\|\|v(\g_j)\|_q}{\|v(\g_j)\|_q\|v(\g_j^\star)\|_q}.
\end{equation*}
Since $\|(\s, \g) - (\s^\star, \g^\star)\| \leq \delta_j$, we have $\|v(\g_j)\|$ and $\|\|v(\g_j)\|_q$ are bounded. Since $q \in (1, 2]$, the H\"{o}lder inequality implies that $\|v(\g_j) - v(\g_j^\star)\|_q \leq m^{\frac{2-q}{2q}}\|v(\g_j) - v(\g_j^\star)\|$. Putting these pieces together yields that 
\begin{equation}\label{inequality:GEB-other-third}
d(\alpha, \SCal_j) \leq C\left(m^{\frac{2-q}{2q}} + 1\right)\|v(\g_j) - v(\g_j^\star)\| \overset{\text{Proposition~\ref{Prop:opt-structure}}}{\leq} C\nu_j\left(m^{\frac{2-q}{2q}} + 1\right)\|\g_j - \g_j^\star\|. 
\end{equation} 
Plugging Eq.~\eqref{inequality:GEB-other-second} and Eq.~\eqref{inequality:GEB-other-third} into Eq.~\eqref{inequality:GEB-other-first} yields that 
\begin{equation*}
d(\alpha, \Sigma(\s^\star, \g^\star)) \leq \left(\bar{\kappa}\max\left\{\theta, C\left(m^{\frac{2-q}{2q}} + 1\right)\left(\sum_{j \in \JCal} \nu_j\right)\right\}\right)\|(\s, \g) - (\s^\star, \g^\star)\|,  
\end{equation*}
for any $\alpha \in \Sigma(\s, \g)$ with $\|(\s, \g) - (\s^\star, \g^\star)\| \leq \min_{j=1}^n \delta_j$. This implies that $\Sigma$ has the ULC property at $(\s^\star, \g^\star) \in \br^d \times \br^{mn}$. Therefore, we conclude from Theorem~\ref{Theorem:GEB-sufficient} that the GEB condition holds for the dual filtering-clustering model in Eq.~\eqref{prob:dual-main}. 

\subsection{Counterexample to the case without $\JCal = [n]$}\label{app:counterexample}
The following example, which is presented in~\citet[Remark 2]{Jane-2021-Variational}, shows that the GEB condition fails in general when $q \in (1, 2]$. In particular, we consider 
\begin{equation}\label{prob:counterexample}
\min_{\x \in \BB_2} \ \bar{f}(\alpha) = \frac{1}{2}(\alpha_2 - 1)^2. 
\end{equation}
It is clear that $(0, 1)$ is the only point in $\Omega^\star$ and $\g^\star = 0$. Define the sequence
\begin{equation*}
\alpha_t = (\cos \theta_t, \sin\theta_t), \quad \theta_t \in (0, \tfrac{\pi}{2}), \quad \theta_t \rightarrow \tfrac{\pi}{2},  
\end{equation*}
we have $\alpha_t \in \BB_2$ and $\alpha_t \rightarrow (0, 1)$. Furthermore,
\begin{equation*}
R(\alpha_t) = \frac{((1 - \sqrt{1 + \cos^2\theta_t})\cos\theta_t, 1 - \sin\theta_t\sqrt{1 + \cos^2\theta_t})}{\sqrt{1 + \cos^2\theta_t}}. 
\end{equation*}
Let $z_t = \sin\theta_t$, we have
\begin{equation*}
\|R(\alpha_t)\| = \frac{\sqrt{2 + 2\cos^2\theta_t - 2\sqrt{1 + \cos^2\theta_t}((\cos^2\theta_t + \sin\theta_t)}}{\sqrt{1 + \cos^2\theta_t}} = \sqrt{2 - \frac{2 + 2z_t - 2z_t^2}{\sqrt{2 - z_t^2}}}. 
\end{equation*}
Therefore, we conclude that 
\begin{equation*}
\frac{d(\alpha_t, \Omega^\star)}{\|R(\alpha_t)\|} = \sqrt{\frac{(1-z_t)(2-z_t^2)}{2 - z_t^2 - (1 + z_t - z_t^2)\sqrt{2 - z_t^2}}}.
\end{equation*}
Letting $t \rightarrow +\infty$ and using the L'Hospital's rule, we have
\begin{equation*}
\lim_{t \rightarrow +\infty} \left(\frac{d(\alpha_t, \Omega^\star)}{\|R(\alpha_t)\|}\right)^2 = \lim_{z \rightarrow 1} \frac{3z^2 - 2z - 2}{-2z - (1-2z)\sqrt{2-z^2} + z(1 + z - z^2)(2 - z^2)^{-1/2}} = +\infty. 
\end{equation*}
This implies that the GEB condition does not hold true for the model in Eq.~\eqref{prob:counterexample}. 

\section{Postponed Proofs in Section~\ref{sec:framework}}
This section lays out the detailed proofs for Theorem~\ref{Theorem:GDGA-Framework}, Corollary~\ref{Corollary:GDGA-Stochastic-Framework}, Theorem~\ref{Theorem:GDGA-deterministic}, Theorem~\ref{Theorem:GDGA-finite-sum} as well as Theorem~\ref{Theorem:GDGA-stochastic}. 

\subsection{Technical lemmas}
We prove several useful technical lemmas. The key quantity here is the distance between $\beta_t$ and $\beta^\star(\alpha_t) = \argmin_{\beta \in \br^d} f(\beta) - \lambda\alpha_t^\top D\beta$ given by 
\begin{equation*}
\delta_t \mydefn \|\beta_t - \beta^\star(\alpha_t)\|. 
\end{equation*}
We also denote $\bar{\alpha}_t$ as the projection of $\alpha_t$ onto the optimal set $\Omega^\star$ of the dual filtering-clustering model in Eq.~\eqref{prob:dual-main}, and define the objective gap between $\bar{f}(\alpha_t)$ and $\bar{f}(\bar{\alpha}_t)$ by
\begin{equation*}
\Delta_t \mydefn \bar{f}(\alpha_t) - \bar{f}(\bar{\alpha}_t). 
\end{equation*}
Our first lemma provides a key lower bound for the iterative objective gap $\Delta_t - \Delta_{t+1}$.  
\begin{lemma}\label{Lemma:GDGA-objgap}
Under Assumption~\ref{Assumption:main} and let $(\alpha_t, \beta_t)_{t \geq 0}$ be generated by Algorithm~\ref{Algorithm:GDGA} with the stepsize $\eta \in (0, \min\{1, \frac{\mu}{4\|D\|^2\max\{1, \lambda^2\}}\})$. Then, for any $t \geq 0$, we have
\begin{equation*}
\Delta_t - \Delta_{t+1} \geq \left(\frac{\lambda^2}{4\eta}\right)\|\alpha_t - \alpha_{t+1}\|^2 - \left(\frac{\eta\|D\|^2}{2}\right)\delta_t^2. 
\end{equation*}
\end{lemma}
\begin{proof}
Since $\alpha_{t+1} = \PCal_{\BB_q^n}(\alpha_t - \eta\lambda D\beta_t)$, we have
\begin{equation*}
(\alpha - \alpha_{t+1})^\top(\alpha_{t+1} - \alpha_t + \eta\lambda D\beta_t) \geq 0, \quad \textnormal{for all } \alpha \in \BB_q^n. 
\end{equation*}
Plugging $\alpha = \alpha_t$ into the above inequality and rearranging yields that 
\begin{equation}\label{inequality:GDGA-objgap-first}
\lambda(\alpha_t - \alpha_{t+1})^\top D\beta_t \geq \frac{\|\alpha_t - \alpha_{t+1}\|^2}{\eta}. 
\end{equation}
Since $\bar{f}$ is $\frac{\lambda^2\|D\|^2}{\mu}$-gradient Lipschiz (cf. Lemma~\ref{Lemma:dual-objective}), we have
\begin{eqnarray}\label{inequality:GDGA-objgap-second}
\lefteqn{(\alpha_t - \alpha_{t+1})^\top \lambda D\beta_t \ = \ (\alpha_t - \alpha_{t+1})^\top\nabla \bar{f}(\alpha_t) + (\alpha_t - \alpha_{t+1})^\top(\lambda D\beta_t - \nabla \bar{f}(\alpha_t))} \\
& \leq & \bar{f}(\alpha_t) - \bar{f}(\alpha_{t+1}) + \left(\frac{\lambda^2\|D\|^2}{\mu}\right)\|\alpha_t - \alpha_{t+1}\|^2 + (\alpha_t - \alpha_{t+1})^\top(\lambda D\beta_t - \nabla \bar{f}(\alpha_t)). \nonumber
\end{eqnarray}
Using $\nabla \bar{f}(\alpha_t) = \lambda D\beta^\star(\alpha_t)$ and the Young's inequality, we have
\begin{eqnarray}\label{inequality:GDGA-objgap-third}
\lefteqn{(\alpha_t - \alpha_{t+1})^\top(\lambda D\beta_t - \nabla \bar{f}(\alpha_t)) \ = \ (\alpha_t - \alpha_{t+1})^\top(\lambda D\beta_t - \lambda D\beta^\star(\alpha_t))} \\
& \leq & \left(\frac{\lambda^2}{2\eta}\right)\|\alpha_t - \alpha_{t+1}\|^2 + \left(\frac{\eta}{2}\right)\|D\beta_t - D\beta^\star(\alpha_t)\|^2 \ \leq \ \left(\frac{\lambda^2}{2\eta}\right)\|\alpha_t - \alpha_{t+1}\|^2 + \left(\frac{\eta\|D\|^2}{2}\right)\delta_t^2. \nonumber
\end{eqnarray}
Plugging Eq.~\eqref{inequality:GDGA-objgap-second} and Eq.~\eqref{inequality:GDGA-objgap-third} into Eq.~\eqref{inequality:GDGA-objgap-first} yields that 
\begin{eqnarray*}
\bar{f}(\alpha_t) - \bar{f}(\alpha_{t+1}) & \geq & \left(\frac{\lambda^2}{2\eta} - \frac{\lambda^2\|D\|^2}{\mu}\right)\|\alpha_t - \alpha_{t+1}\|^2 - \left(\frac{\eta\|D\|^2}{2}\right)\delta_t^2 \\
& \geq & \left(\frac{\lambda^2}{4\eta}\right)\|\alpha_t - \alpha_{t+1}\|^2 - \left(\frac{\eta\|D\|^2}{2}\right)\delta_t^2. 
\end{eqnarray*}
This completes the proof. 
\end{proof}
Our second lemma presents an upper bound for $\Delta_{t+1}$ based on $\|\alpha_t - \alpha_{t+1}\|^2$ using a global error bound for the dual filtering-clustering model in Theorem~\ref{Theorem:GEB-polyhedron} and~\ref{Theorem:GEB-other}. 
\begin{lemma}\label{Lemma:GDGA-err}
Under Assumption~\ref{Assumption:main} and let $(\alpha_t, \beta_t)_{t \geq 0}$ be generated by Algorithm~\ref{Algorithm:GDGA} with the stepsize $\eta \in (0, \min\{1, \frac{\mu}{4\|D\|^2\max\{1, \lambda^2\}}\})$. Then, for any $t \geq 0$, we have
\begin{equation*}
\|\alpha_{t+1} - \alpha_t\|^2 \geq \left(\frac{4\tau\eta^2}{17\tau^2 + 14\tau + 1}\right)\Delta_{t+1} - \left(\frac{8\tau^2\lambda^2\|D\|^2\eta^2}{17\tau^2 + 14\tau + 1}\right)\delta_t^2. 
\end{equation*}
\end{lemma}
\begin{proof}
By the definition of $\nabla \bar{f}$ and using the triangle inequality, we have
\begin{eqnarray*}
\lefteqn{\|\alpha_t - \PCal_{\BB_q^n}(\alpha_t - \eta \nabla \bar{f}(\alpha_t))\| \ = \ \|\alpha_t - \PCal_{\BB_q^n}(\alpha_t - \eta \lambda D\beta^\star(\alpha_t))\|} \\
& \leq & \|\alpha_t - \alpha_{t+1}\| + \|\alpha_{t+1} - \PCal_{\BB_q^n}(\alpha_t - \eta \lambda D\beta^\star(\alpha_t))\|. 
\end{eqnarray*}
Since $\alpha_{t+1}=\PCal_{\BB_q^n}(\alpha_t - \eta \lambda D\beta_t)$ and the projection operator is nonexpansive, we have
\begin{equation*}
\|\alpha_{t+1} - \PCal_{\BB_q^n}(\alpha_t - \eta \lambda D\beta^\star(\alpha_t))\| \leq \eta\lambda\|D\|\delta_t.
\end{equation*}
We see from~\citet[Lemma~1]{Gafni-1984-Two} that $(1/\eta)\|\alpha_t - \PCal_{\BB_q^n}(\alpha_t - \eta \nabla \bar{f}(\alpha_t))\|$ is nonincreasing as a function of $\eta>0$. Since $\eta \in (0, 1)$, we have
\begin{equation*}
\eta\|\alpha_t - \PCal_{\BB_q^n}(\alpha_t - \nabla \bar{f}(\alpha_t))\| \leq \|\alpha_t - \PCal_{\BB_q^n}(\alpha_t - \eta \nabla \bar{f}(\alpha_t))\|. 
\end{equation*}
Putting these pieces together yields that 
\begin{equation}\label{inequality:GDGA-err-first}
\|\alpha_t - \PCal_{\BB_q^n}(\alpha_t - \nabla \bar{f}(\alpha_t))\| \leq \frac{\|\alpha_t - \alpha_{t+1}\|}{\eta} + \lambda\|D\|\delta_t. 
\end{equation}
Since $\bar{\alpha}_t$ is the projection of $\alpha_t$ onto $\Omega^\star$, we derive from Theorem~\ref{Theorem:GEB-polyhedron} and~\ref{Theorem:GEB-other} that 
\begin{equation*}
\|\alpha_t - \bar{\alpha}_t\| \leq \tau\|\alpha - \PCal_{\BB_q^n}(\alpha_t - \nabla \bar{f}(\alpha_t))\|. 
\end{equation*}
Plugging Eq.~\eqref{inequality:GDGA-err-first} into the above inequality yields that 
\begin{equation}\label{inequality:GDGA-err-second}
\|\alpha_t - \bar{\alpha}_t\| \leq \left(\frac{\tau}{\eta}\right)\|\alpha_t - \alpha_{t+1}\| + \tau\lambda\|D\|\delta_t. 
\end{equation}
It suffices to lower bound $\|\alpha_t - \bar{\alpha}_t\|$ using $\Delta_{t+1}$. Since $\alpha_{t+1} = \PCal_{\BB_q^n}(\alpha_t - \eta \lambda D\beta_t)$, we have
\begin{equation*}
(\alpha - \alpha_{t+1})^\top(\alpha_{t+1} - \alpha_t + \eta \lambda D\beta_t) \geq 0, \quad \textnormal{for all } \alpha \in \BB_q^n. 
\end{equation*}
Letting $\alpha=\bar{\alpha}_t$ in the above inequality and rearranging the inequality yields that 
\begin{equation}\label{inequality:GDGA-err-third}
\lambda(\alpha_{t+1} - \bar{\alpha}_t)^\top D\beta_t \leq \left(\frac{1}{\eta}\right)(\alpha_{t+1} - \bar{\alpha}_t)^\top(\alpha_t - \alpha_{t+1}). 
\end{equation}
Since $\bar{\alpha}_t, \bar{\alpha}_{t+1} \in \Omega^\star$, we have $\bar{f}(\bar{\alpha}_t) = \bar{f}(\bar{\alpha}_{t+1})$. By the convexity of $\bar{f}$, we have
\begin{equation}\label{inequality:GDGA-err-fourth}
\Delta_{t+1} = \bar{f}(\alpha_{t+1}) - \bar{f}(\bar{\alpha}_{t+1}) = \bar{f}(\alpha_{t+1}) - \bar{f}(\bar{\alpha}_t) \leq (\alpha_{t+1} - \bar{\alpha}_t)^\top\nabla\bar{f}(\alpha_{t+1}). 
\end{equation}
Since $\nabla\bar{f}(\alpha_t) = \lambda D\beta^\star(\alpha_t)$, we have
\begin{eqnarray*}
\lefteqn{(\alpha_{t+1} - \bar{\alpha}_t)^\top\nabla\bar{f}(\alpha_{t+1}) \ \leq \ \underbrace{(\alpha_{t+1} - \bar{\alpha}_t)^\top(\nabla\bar{f}(\alpha_{t+1}) - \nabla\bar{f}(\alpha_t))}_{\textbf{I}}} \\
& & + \underbrace{(\alpha_{t+1} - \bar{\alpha}_t)^\top(\lambda D\beta^\star(\alpha_t) - \lambda D\beta_t)}_{\textbf{II}} + \underbrace{\lambda (\alpha_{t+1} - \bar{\alpha}_t)^\top D\beta_t}_{\textbf{III}}. 
\end{eqnarray*}
Since $\bar{f}$ is $\frac{\lambda^2\|D\|^2}{\mu}$-gradient Lipschiz (cf. Lemma~\ref{Lemma:dual-objective}), we have
\begin{equation*}
\textbf{I} \leq \left(\frac{\lambda^2\|D\|^2}{\mu}\right)\|\alpha_{t+1} - \alpha_t\|\|\alpha_{t+1} - \bar{\alpha}_t\|. 
\end{equation*}
By the definition of $\delta_t$, we have
\begin{equation*}
\textbf{II} \leq \lambda\delta_t\|D\|\|\alpha_{t+1} - \bar{\alpha}_t\|. 
\end{equation*}
In addition, Eq.~\eqref{inequality:GDGA-err-third} implies that 
\begin{equation*}
\textbf{III} \leq \left(\frac{1}{\eta}\right)\|\alpha_t - \alpha_{t+1}\|\|\alpha_{t+1} - \bar{\alpha}_t\|. 
\end{equation*}
Therefore, we conclude that 
\begin{equation*}
(\alpha_{t+1} - \bar{\alpha}_t)^\top\nabla\bar{f}(\alpha_{t+1}) \leq \left(\lambda\delta_t\|D\| + \left(\frac{\lambda^2\|D\|^2}{\mu} + \frac{1}{\eta}\right)\left\|\alpha_{t+1} - \alpha_t\right\|\right)\|\alpha_{t+1} - \bar{\alpha}_t\|. 
\end{equation*}
By the triangle inequality, we have
\begin{equation*}
\|\alpha_{t+1} - \bar{\alpha}_t\| \leq \|\alpha_{t+1} - \alpha_t\| + \|\alpha_t - \bar{\alpha}_t\| \overset{\textnormal{Eq.}~\eqref{inequality:GDGA-err-second}}{\leq} \left(1 + \frac{\tau}{\eta}\right)\|\alpha_t - \alpha_{t+1}\| + \tau\lambda\|D\|\delta_t.
\end{equation*}
Putting these pieces together yields that 
\begin{equation*}
(\alpha_{t+1} - \bar{\alpha}_t)^\top\nabla\bar{f}(\alpha_{t+1}) \leq \left(\lambda\delta_t\|D\| + \left(\frac{\lambda^2\|D\|^2}{\mu} + \frac{1}{\eta}\right)\left\|\alpha_{t+1} - \alpha_t\right\|\right)\left(\left(1 + \frac{\tau}{\eta}\right)\|\alpha_t - \alpha_{t+1}\| + \tau\lambda\|D\|\delta_t\right). 
\end{equation*}
By the definition of $\eta$, we have $\eta \leq 1$ and $\frac{\lambda^2\|D\|^2}{\mu} < \frac{1}{\eta}$. Combining these two inequalities with the above inequality yields that 
\begin{equation*}
(\alpha_{t+1} - \bar{\alpha}_t)^\top\nabla\bar{f}(\alpha_{t+1}) \leq \left(\lambda\delta_t\|D\| + \left(\frac{2}{\eta}\right)\left\|\alpha_{t+1} - \alpha_t\right\|\right)\left(\left(\frac{\tau+1}{\eta}\right)\|\alpha_t - \alpha_{t+1}\| + \tau\lambda\|D\|\delta_t\right). 
\end{equation*}
which implies that 
\begin{equation*}
(\alpha_{t+1} - \bar{\alpha}_t)^\top\nabla\bar{f}(\alpha_{t+1}) \leq \left(\frac{2\tau + 2}{\eta^2}\right)\|\alpha_{t+1} - \alpha_t\|^2 + \left(\frac{(3\tau + 1)\lambda\|D\|}{\eta}\right)\delta_t\|\alpha_{t+1} - \alpha_t\| + \tau\lambda^2\|D\|^2\delta_t^2. 
\end{equation*}
By the Young's inequality, we have
\begin{equation*}
\delta_t\|\alpha_{t+1} - \alpha_t\| \leq \left(\frac{\tau \eta \lambda\|D\|}{3\tau + 1}\right)\delta_t^2 + \left(\frac{3\tau + 1}{4\tau \eta \lambda\|D\|}\right)\|\alpha_{t+1} - \alpha_t\|^2.
\end{equation*} 
Combining the above two inequalities, we have
\begin{equation*}
(\alpha_{t+1} - \bar{\alpha}_t)^\top\nabla\bar{f}(\alpha_{t+1}) \leq \left(\frac{17\tau^2 + 14 \tau + 1}{4\tau\eta^2}\right)\|\alpha_{t+1} - \alpha_t\|^2 + 2\tau\lambda^2\|D\|^2\delta_t^2. 
\end{equation*}
This together with Eq.~\eqref{inequality:GDGA-err-fourth} yields that 
\begin{equation*}
\|\alpha_{t+1} - \alpha_t\|^2 \geq \frac{4\tau\eta^2\Delta_{t+1}}{17\tau^2 + 14\tau + 1} - \left(\frac{8\tau^2\lambda^2\|D\|^2\eta^2}{17\tau^2 + 14\tau + 1}\right)\delta_t^2. 
\end{equation*}
This completes the proof. 
\end{proof} 
Equipped with the bounds of iterative objective gap $\Delta_t - \Delta_{t + 1}$ and objective gap $\Delta_{t + 1}$ in Lemma~\ref{Lemma:GDGA-objgap} and~\ref{Lemma:GDGA-err}, we prove the main lemma on the number of iterations required by Algorithm~\ref{Algorithm:GDGA} to reach a certain threshold with objective gap $\Delta_t$. Before stating that result, we assume the key technical assumption with an error $\hat{\varepsilon}$ for each inner. 
\begin{equation}\label{condition:err_inner_loop}
\hat{\varepsilon} = \frac{\sqrt{\varepsilon}}{2} \cdot \min\left\{1, \frac{\lambda\mu}{2\sqrt{\ell}}\sqrt{\frac{\tau}{C(17\tau^2 + (14 + \eta\lambda^2) \tau + 1)}}\right\}, 
\end{equation}
where $C>0$ is defined as 
\begin{equation*}
C \mydefn \frac{2\tau^2\lambda^4\|D\|^2}{17\tau^2 + 14\tau + 1} + \frac{\|D\|^2}{2}. 
\end{equation*}
\begin{lemma}\label{Lemma:GDGA-convergence}
Under Assumption~\ref{Assumption:main} and let $(\alpha_t, \beta_t)_{t \geq 0}$ be generated by Algorithm~\ref{Algorithm:GDGA} with the stepsize $\eta \in (0, \min\{1, \frac{\mu}{4\|D\|^2\max\{1, \lambda^2\}}\})$ and $\hat{\varepsilon} > 0$ satisfies Eq.~\eqref{condition:err_inner_loop}. Then, the number of iterations $T > 0$ required to reach $\Delta_T \leq (\frac{\mu^2}{8\ell})\varepsilon$ satisfies that 
\begin{equation*}
T \leq 1 + \left(\frac{17\tau^2 + 14\tau + 1}{\tau\lambda^2\eta}\right)\log\left(\frac{16\ell\Delta_1}{\mu^2\varepsilon}\right). 
\end{equation*}
\end{lemma}
\begin{proof}
Invoking the results from Lemma~\ref{Lemma:GDGA-objgap} and Lemma~\ref{Lemma:GDGA-err}, we have
\begin{eqnarray*}
\Delta_t - \Delta_{t+1} & \geq & \frac{\lambda^2}{4\eta}\left(\left(\frac{4\tau\eta^2}{17\tau^2 + 14\tau + 1}\right)\Delta_{t+1} - \left(\frac{8\tau^2\lambda^2\|D\|^2\eta^2}{17\tau^2 + 14\tau + 1}\right)\delta_t^2\right) - \left(\frac{\eta\|D\|^2}{2}\right)\delta_t^2 \\
& = & \left(\frac{\tau\lambda^2\eta}{17\tau^2 + 14\tau + 1}\right)\Delta_{t+1} - \left(\frac{2\tau^2\lambda^4\|D\|^2}{17\tau^2 + 14\tau + 1} + \frac{\|D\|^2}{2}\right)\eta\delta_t^2. 
\end{eqnarray*}
Since $\beta_t \leftarrow \textsc{InnerLoop}(f, \lambda, D, \alpha_t, \beta_{t-1}, \hat{\varepsilon})$, we have $\delta_t \leq \hat{\varepsilon}$. In addition, we define $\rho > 0$ by
\begin{eqnarray*}
\rho \mydefn  \left(1 + \frac{\tau\lambda^2\eta}{17\tau^2 + 14\tau + 1}\right)^{-1}.
\end{eqnarray*}
Then, for any $t \geq 0$, we find that
\begin{equation*}
\Delta_t - \Delta_{t+1} \geq \left(\frac{1}{\rho} - 1\right)\Delta_{t+1} - C\eta\hat{\varepsilon}^2, 
\end{equation*}
which implies that 
\begin{equation*}
\Delta_{t+1} \leq \rho \Delta_t + \rho C\eta\hat{\varepsilon}^2.   
\end{equation*}
Recursively performing the above inequality yields that 
\begin{equation*}
\Delta_t \leq \rho^{t-1} \Delta_1 + \left(\sum_{j=1}^{t-1} \rho^{t-j}\right) \cdot C\eta\hat{\varepsilon}^2 \leq \rho^{t-1} \Delta_1 + \left(\frac{C\eta}{1-\rho}\right)\hat{\varepsilon}^2. 
\end{equation*}
By the definition of $\rho$, we have
\begin{equation*}
\frac{C\eta}{1-\rho} = C\eta + \frac{C\left(17\tau^2 + 14\tau + 1\right)}{\tau\lambda^2}. 
\end{equation*}
By the definition of $\hat{\varepsilon}$ in Eq.~\eqref{condition:err_inner_loop}, we have $\Delta_t \leq \rho^{t-1} \Delta_1 + (\frac{\mu^2}{16\ell})\varepsilon$. Therefore, the number of iterations $T > 0$ required to reach $\Delta_T \leq (\frac{\mu^2}{8\ell})\varepsilon$ is 
\begin{equation*}
T \leq 1 + \left(\frac{17\tau^2 + 14\tau + 1}{\tau\lambda^2\eta}\right)\log\left(\frac{16\ell\Delta_1}{\mu^2\varepsilon}\right). 
\end{equation*}
This completes the proof. 
\end{proof}
Finally, we present the lemmas which will be used for the finite-sum setting and the online setting. Indeed, we consider the stochastic variant of the GDGA algorithm in these two settings which can be much more efficient than the deterministic counterpart in real applications~\citep{Lan-2020-First}. The stochastic GDGA algorithms are associated with the subroutines which are implemented using stochastic first-order optimization algorithms, e.g., Katyusha~\citep{Allen-2017-Katyusha} and optimal SGD~\citep{Rakhlin-2012-Making}, and the stopping criterion $\EE[\delta_t] \leq \hat{\varepsilon}$. We omit the proofs since they resemble the proofs for Lemma~\ref{Lemma:GDGA-objgap},~\ref{Lemma:GDGA-err} and~\ref{Lemma:GDGA-convergence} in the deterministic setting. 
\begin{lemma}
Under Assumption~\ref{Assumption:main} and let $(\alpha_t, \beta_t)_{t \geq 0}$ be generated by Algorithm~\ref{Algorithm:GDGA} with the stepsize $\eta \in (0, \min\{1, \frac{\mu}{4\|D\|^2\max\{1, \lambda^2\}}\})$. Then, for any $t \geq 0$, we have
\begin{equation*}
\EE[\Delta_t] - \EE[\Delta_{t+1}] \geq \left(\frac{\lambda^2}{4\eta}\right)\EE[\|\alpha_t - \alpha_{t+1}\|^2] - \left(\frac{\eta\|D\|^2}{2}\right) \EE[\delta_t^2]. 
\end{equation*}
\end{lemma}
\begin{lemma}
Under Assumption~\ref{Assumption:main} and let $(\alpha_t, \beta_t)_{t \geq 0}$ be generated by Algorithm~\ref{Algorithm:GDGA} with the stepsize $\eta \in (0, \min\{1, \frac{\mu}{4\|D\|^2\max\{1, \lambda^2\}}\})$. Then, for any $t \geq 0$, we have
\begin{equation*}
\EE[\|\alpha_{t+1} - \alpha_t\|^2] \geq \left(\frac{4\tau\eta^2}{17\tau^2 + 14\tau + 1}\right)\EE[\Delta_{t+1}] - \left(\frac{8\tau^2\lambda^2\|D\|^2\eta^2}{17\tau^2 + 14\tau + 1}\right)\EE[\delta_t^2]. 
\end{equation*}
\end{lemma}
\begin{lemma}\label{Lemma:SGDGA-convergence}
Under Assumption~\ref{Assumption:main} and let $(\alpha_t, \beta_t)_{t \geq 0}$ be generated by Algorithm~\ref{Algorithm:GDGA} with the stepsize $\eta \in (0, \min\{1, \frac{\mu}{4\|D\|^2\max\{1, \lambda^2\}}\})$ and $\hat{\varepsilon} > 0$ satisfies Eq.~\eqref{condition:err_inner_loop}. Then, the number of iterations $T > 0$ required to reach $\EE[\Delta_T] \leq (\frac{\mu^2}{8\ell})\varepsilon$ satisfies that 
\begin{equation*}
T \leq 1 + \left(\frac{17\tau^2 + 14\tau + 1}{\tau\lambda^2\eta}\right)\log\left(\frac{16\ell\Delta_1}{\mu^2\varepsilon}\right). 
\end{equation*}
\end{lemma}

\subsection{Proof of Theorem~\ref{Theorem:GDGA-Framework}}
We recall that $\bar{\alpha}_t$ is the projection of $\alpha_t$ onto the optimal set of the dual filtering-clustering model in Eq.~\eqref{prob:dual-main} and $\Delta_t = \bar{f}(\alpha_t) - \bar{f}(\bar{\alpha}_t)$ is the objective gap.  The following is the full version of Theorem~\ref{Theorem:GDGA-Framework}. 
\begin{theorem}
Under Assumption~\ref{Assumption:main} and set $\eta \in (0, \min\{1, \frac{\mu}{4\|D\|^2\max\{1, \lambda^2\}}\})$ in Algorithm~\ref{Algorithm:GDGA} and $\tau > 0$ be defined in Theorem~\ref{Theorem:GEB-polyhedron} and~\ref{Theorem:GEB-other}. Given a tolerance $\varepsilon > 0$, we let $\hat{\varepsilon} > 0$ be
\begin{equation*}
\hat{\varepsilon} = \frac{\sqrt{\varepsilon}}{2} \cdot \min\left\{1, \frac{\lambda\mu}{2\sqrt{\ell}}\sqrt{\frac{\tau}{C(17\tau^2 + (14 + \eta\lambda^2) \tau + 1)}}\right\}, 
\end{equation*}
where $C>0$ is defined as 
\begin{equation*}
C \mydefn \frac{2\tau^2\lambda^4\|D\|^2}{17\tau^2 + 14\tau + 1} + \frac{\|D\|^2}{2}. 
\end{equation*}
Then, the number of iterations required by Algorithm~\ref{Algorithm:GDGA} to return an $\varepsilon$-optimal solution is bounded by
\begin{equation*}
T \leq 1 + \left(\frac{17\tau^2 + 14\tau + 1}{\tau\lambda^2\eta}\right)\log\left(\frac{16\ell\Delta_1}{\mu^2\varepsilon}\right).  
\end{equation*}
\end{theorem}
\begin{proof}
Suppose that $\beta^\star \in \br^d$ is an optimal solution of the filtering-clustering model in Eq.~\eqref{prob:main}, we have
\begin{equation*}
\|\beta_t - \beta^\star\|^2 \leq 2(\underbrace{\|\beta_t - \beta^\star(\alpha_t)\|^2}_{\textbf{I}} + \underbrace{\|\beta^\star(\alpha_t) - \beta^\star\|^2}_{\textbf{II}}). 
\end{equation*}
Since $\beta_t \leftarrow \textsc{InnerLoop}(f, \lambda, D, \alpha_t, \beta_{t-1}, \hat{\varepsilon})$ and $\hat{\varepsilon} \leq \frac{\sqrt{\varepsilon}}{2}$, we have $\textbf{I} \leq \hat{\varepsilon}^2 \leq \frac{\varepsilon^2}{4}$. 

Recall that $\bar{\alpha}_t$ is the projection of $\alpha_t$ onto the optimal set of the dual filtering-clustering model in Eq.~\eqref{prob:dual-main}. Since the objective of the filtering-clustering model is strongly convex, its optimal solution is unique. Then, by the definition of $\beta^\star(\cdot)$, we have $\beta^\star = \beta^\star(\bar{\alpha}_t)$. Therefore, we conclude that 
\begin{equation*}
\textbf{II} \overset{\text{Lemma}~\ref{Lemma:dual-objective}}{\leq} \frac{\|\lambda D^\top\alpha_t - \lambda D^\top\bar{\alpha}_t\|^2}{\mu^2} \overset{\text{Lemma}~\ref{Lemma:conjugate}}{\leq} \frac{2\ell(f^\star(\lambda D^\top\alpha_t) - f^\star(\lambda D^\top\bar{\alpha}_t))}{\mu^2} = \frac{2\ell\Delta_t}{\mu^2}. 
\end{equation*} 
If the number of iterations $T > 0$ satisfies that 
\begin{equation*}
T > 1 + \left(\frac{17\tau^2 + 14\tau + 1}{\tau\lambda^2\eta}\right)\log\left(\frac{16\ell\Delta_1}{\mu^2\varepsilon}\right),   
\end{equation*}
we obtain from Lemma~\ref{Lemma:GDGA-convergence} that $\Delta_t \leq (\frac{\mu^2}{8\ell})\varepsilon$.  This implies that 
\begin{equation*}
\|\beta^\star(\alpha_t) - \beta^\star\|^2 \leq \frac{\varepsilon}{4}. 
\end{equation*}
Putting these pieces together yields that $\|\beta_t - \beta^\star\|^2 \leq \varepsilon$.
\end{proof}

\subsection{Proof of Corollary~\ref{Corollary:GDGA-Stochastic-Framework}}
The proof is the same as that in Theorem~\ref{Theorem:GDGA-Framework}. In particular, it suffices to bound $\EE[\|\beta^\star(\alpha_t) - \beta^\star\|^2]$ via appeal to Lemma~\ref{Lemma:SGDGA-convergence} instead of Lemma~\ref{Lemma:GDGA-convergence}. 

\subsection{Proof of Theorem~\ref{Theorem:GDGA-deterministic}}
We first present the complexity bound of the AGD-based subroutine in the following proposition. 
\begin{proposition}\label{Prop:subroutine-AGD}
Under Assumption~\ref{Assumption:main}, the required number of gradient evaluations in the implementation of $\beta_t \leftarrow \textsc{InnerLoopAgd}(f(\cdot), \lambda, D, \alpha_t, \beta_{t-1}, \hat{\varepsilon})$ is bounded by
\begin{equation*}
N_t \leq \sqrt{\frac{\ell}{\mu}} \cdot \left\{ \begin{array}{ll} 
\log\left(\frac{\|\beta_0 - \beta^\star(\alpha_1)\|}{\hat{\varepsilon}}\right), & t = 1, \\
\log\left(1 + \frac{\lambda\|D\|\|\alpha_t - \alpha_{t-1}\|}{\mu\hat{\varepsilon}}\right), & t \geq 2. 
\end{array}\right.  
\end{equation*}
\end{proposition}
\begin{proof}
Since $f$ is $\mu$-strongly convex and $\ell$-gradient Lipschitz, we have $f(\beta) - \lambda \alpha_t^\top D\beta$ is also $\mu$-strongly convex and $\ell$-gradient Lipschiz. For $t=1$, the initial distance is $\|\beta_0 - \beta^\star(\alpha_1)\|$. By the convergence theory for AGD in~\citet{Nesterov-2018-Lectures}, we have
\begin{equation*}
N_1 \leq \sqrt{\frac{\ell}{\mu}}\log\left(\frac{\|\beta_0 - \beta^\star(\alpha_1)\|}{\hat{\varepsilon}}\right). 
\end{equation*}
For $t \geq 2$, the initial distance is $\|\beta_{t-1} - \beta^\star(\alpha_t)\|$. Therefore, we have
\begin{equation*}
N_t \leq \sqrt{\frac{\ell}{\mu}}\log\left(\frac{\|\beta_{t-1} - \beta^\star(\alpha_t)\|}{\hat{\varepsilon}}\right). 
\end{equation*}
By using the triangle inequality, we have 
\begin{equation*}
\|\beta_{t-1} - \beta^\star(\alpha_t)\| \leq \|\beta_{t-1} - \beta^\star(\alpha_{t-1})\| + \|\beta^\star(\alpha_{t-1}) - \beta^\star(\alpha_t)\| \leq \hat{\varepsilon} + \|\beta^\star(\alpha_{t-1}) - \beta^\star(\alpha_t)\|.  
\end{equation*}
Since $\beta^\star(\alpha)$ is $\frac{\lambda\|D\|}{\mu}$-Lipschitz over $\BB_q^n$ (cf. Lemma~\ref{Lemma:dual-objective}), we have
\begin{equation*}
\|\beta^\star(\alpha_{t-1}) - \beta^\star(\alpha_t)\| \leq \frac{\lambda\|D\|\|\alpha_t - \alpha_{t-1}\|}{\mu}. 
\end{equation*}
Therefore, we have
\begin{equation*}
N_t \leq \sqrt{\frac{\ell}{\mu}}\log\left(1 + \frac{\lambda\|D\|\|\alpha_t - \alpha_{t-1}\|}{\mu\hat{\varepsilon}}\right). 
\end{equation*}
This completes the proof.
\end{proof}
Equipped with the result in Proposition~\ref{Prop:subroutine-AGD}, we are ready to derive the complexity bound of deterministic GDGA in terms of the number of gradient evaluations. 

By the definition, we have $N = N_1 + \sum_{t=2}^T N_t$. Then, Proposition~\ref{Prop:subroutine-AGD} implies that 
\begin{equation*}
N \leq \sqrt{\frac{\ell}{\mu}}\left((T-1)\log\left(1 + \frac{\lambda\|D\|\bar{D}_q}{\mu\hat{\varepsilon}}\right) + \log\left(\frac{\|\beta_0 - \beta^\star(\alpha_1)\|}{\hat{\varepsilon}}\right)\right). 
\end{equation*}
By Theorem~\ref{Theorem:GDGA-Framework}, we have
\begin{equation*}
T - 1 \leq \left(\frac{17\tau^2 + 14\tau + 1}{\tau\lambda^2\eta}\right)\log\left(\frac{16\ell\Delta_1}{\mu^2\varepsilon}\right). 
\end{equation*}
Putting these pieces together yields the desired result. 

\subsection{Proof of Theorem~\ref{Theorem:GDGA-finite-sum}}
We first present the complexity bound of the Katyusha-based subroutine in the following proposition. 
\begin{proposition}\label{Prop:subroutine-Katyusha}
Under Assumption~\ref{Assumption:main}, the required number of gradient evaluations in the implementation of $\beta_t \leftarrow \textsc{InnerLoopKatyusha}(f(\cdot) = \frac{1}{M}(\sum_{i=1}^M f_i(\cdot)), \lambda, D, \alpha_t, \beta_{t-1}, \hat{\varepsilon})$ is bounded by
\begin{equation*}
N_t \leq C_{\textnormal{Katyusha}} \cdot \left\{ \begin{array}{ll} 
M + \sqrt{\frac{\ell M}{\mu}}\log\left(\frac{\ell\|\beta_0 - \beta^\star(\alpha_1)\|}{\mu\hat{\varepsilon}}\right), & t = 1, \\
M + \sqrt{\frac{\ell M}{\mu}}\log\left(\frac{\ell}{\mu} + \frac{\lambda\ell\|D\|\|\alpha_t - \alpha_{t-1}\|}{\mu^2\hat{\varepsilon}}\right), & t \geq 2, 
\end{array}\right.  
\end{equation*}
where $C_{\textnormal{Katyusha}} > 0$ is a universal constant defined in~\citet[Theorem~2.1]{Allen-2017-Katyusha} and independent of $\ell$, $\mu$, $\|D\|$ and $1/\varepsilon$. 
\end{proposition}
\begin{proof}
Since $f$ is $\mu$-strongly convex and $\ell$-gradient Lipschitz, we have $f(\beta) - \lambda \alpha_t^\top D\beta$ is $\mu$-strongly convex and $\ell$-gradient Lipschiz. For $t=1$, the initial distance is $\|\beta_0 - \beta^\star(\alpha_1)\|$. By the convergence theory for Katyusha~\citep[Theorem~2.1]{Allen-2017-Katyusha} with the suggested stepsize rule $\max\{\frac{2}{3\ell}, \frac{1}{\sqrt{3M\mu\ell}}\}$, we have
\begin{equation*}
N_1 \leq C_{\textnormal{Katyusha}} \cdot \left(M + \sqrt{\frac{\ell M}{\mu}}\log\left(\frac{\ell\|\beta_0 - \beta^\star(\alpha_1)\|}{\mu\hat{\varepsilon}}\right)\right). 
\end{equation*}
For $t \geq 2$, the initial distance is $\|\beta_{t-1} - \beta^\star(\alpha_t)\|$. Therefore, we have
\begin{equation*}
N_t \leq C_{\textnormal{Katyusha}} \cdot \left(M + \sqrt{\frac{\ell M}{\mu}}\log\left(\frac{\ell\|\beta_{t-1} - \beta^\star(\alpha_t)\|}{\mu\hat{\varepsilon}}\right)\right). 
\end{equation*}
Applying the similar argument as that in the proof of Proposition~\ref{Prop:subroutine-AGD}, we have
\begin{equation*}
N_t \leq C_{\textnormal{Katyusha}} \cdot \left(M + \sqrt{\frac{\ell M}{\mu}}\log\left(\frac{\ell}{\mu} + \frac{\lambda\ell\|D\|\|\alpha_t - \alpha_{t-1}\|}{\mu^2\hat{\varepsilon}}\right)\right). 
\end{equation*}
This completes the proof.
\end{proof}
Equipped with the result in Proposition~\ref{Prop:subroutine-Katyusha}, we are ready to derive the complexity bound of stochastic variance reduced GDGA in terms of the number of component gradient evaluations. 

By the definition, we have $N = N_1 + \sum_{2=1}^T N_t$. Then, Proposition~\ref{Prop:subroutine-Katyusha} implies that
\begin{equation*}
N \leq C_{\textnormal{Katyusha}} \cdot \left(TM + (T-1)\sqrt{\frac{\ell M}{\mu}}\log\left(\frac{\ell}{\mu} + \frac{\lambda\ell\|D\|\bar{D}_q}{\mu^2\hat{\varepsilon}}\right) + \sqrt{\frac{\ell M}{\mu}}\log\left(\frac{\ell\|\beta_0 - \beta^\star(\alpha_1)\|}{\mu\hat{\varepsilon}}\right)\right). 
\end{equation*}
By Corollary~\ref{Corollary:GDGA-Stochastic-Framework}, we have
\begin{equation*}
T - 1 \leq \left(\frac{17\tau^2 + 14\tau + 1}{\tau\lambda^2\eta}\right)\log\left(\frac{16\ell\Delta_1}{\mu^2\varepsilon}\right). 
\end{equation*}
Putting these pieces together yields the desired result. 

\subsection{Proof of Theorem~\ref{Theorem:GDGA-stochastic}}
We first present the complexity bound of the SGD-based subroutine in the following proposition. 
\begin{proposition}\label{Prop:subroutine-SGD}
Under Assumption~\ref{Assumption:main}, the required number of gradient evaluations in the implementation of $\beta_t \leftarrow \textsc{InnerLoopSgd}(f(\cdot) = \EE_P[F(\cdot, \xi)], \lambda, D, \alpha_t, \beta_{t-1}, \hat{\varepsilon})$ is bounded by
\begin{equation*}
N_t \leq \frac{4G^2}{\mu^2\hat{\varepsilon}^2}, \quad \textnormal{for all } t \geq 0.  
\end{equation*}
where $G > 0$ refers to the bound for stochastic gradient oracle in Assumption~\ref{Assumption:main}. 
\end{proposition}
\begin{proof}
Since $f$ is $\mu$-strongly convex and $\ell$-gradient Lipschitz, we have $f(\beta) - \lambda \alpha_t^\top D \beta$ is $\mu$-strongly convex and $\ell$-gradient Lipschiz. Under Assumption~\ref{Assumption:main}, we have the stochastic gradient oracle $g(\cdot, \xi)$ is unbiased and bounded: $\EE[\|g(\beta, \xi)\|^2] \leq G^2$ for some $G > 0$. By the convergence theory for optimal SGD~\citep[Lemma~1]{Rakhlin-2012-Making} with the suggested stepsize rule $\frac{1}{\mu k}$ ($k$ refers to SGD iteration count), we have
\begin{equation*}
N_t \leq \frac{4G}{\mu^2\hat{\varepsilon}^2}, \quad \textnormal{for all } t \geq 1. 
\end{equation*}
This completes the proof. 
\end{proof}
Equipped with the result in Proposition~\ref{Prop:subroutine-SGD}, we are ready to derive the complexity bound of stochastic GDGA in terms of the number of stochastic gradient evaluations. 

By the definition, we have $N = \sum_{t=1}^T N_t$. Then, Proposition~\ref{Prop:subroutine-SGD} implies that
\begin{equation*}
N \leq \left(\frac{4G^2}{\mu^2\hat{\varepsilon}^2}\right)T. 
\end{equation*}
By Corollary~\ref{Corollary:GDGA-Stochastic-Framework}, we have
\begin{equation*}
T \leq 1 + \left(\frac{17\tau^2 + 14\tau + 1}{\tau\lambda^2\eta}\right)\log\left(\frac{16\ell\Delta_1}{\mu^2\varepsilon}\right). 
\end{equation*}
Putting these pieces together yields the desired result. 
\begin{figure*}[!t]
\begin{minipage}[b]{.33\textwidth}
\includegraphics[width=60mm,height=45mm]{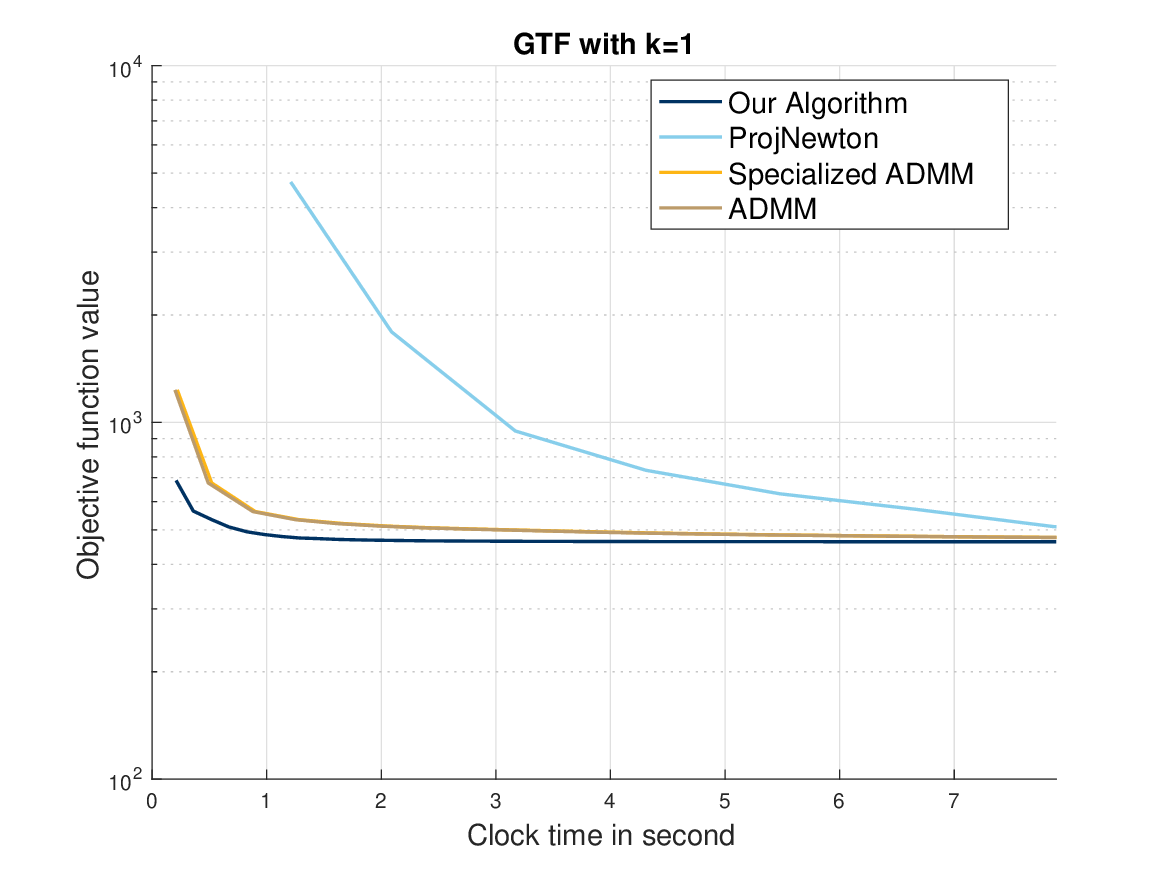}
\end{minipage}
\begin{minipage}[b]{.33\textwidth}
\includegraphics[width=60mm,height=45mm]{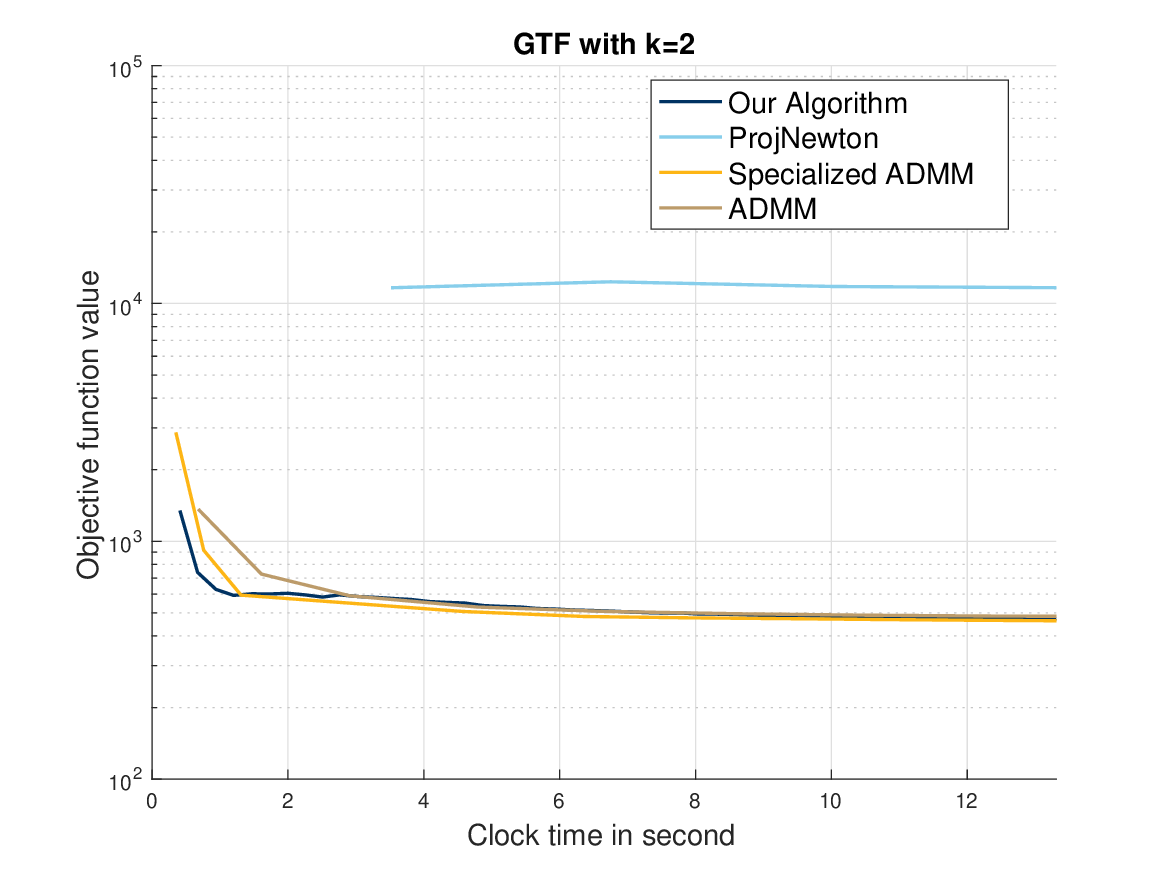}
\end{minipage}
\begin{minipage}[b]{.33\textwidth}
\includegraphics[width=60mm,height=45mm]{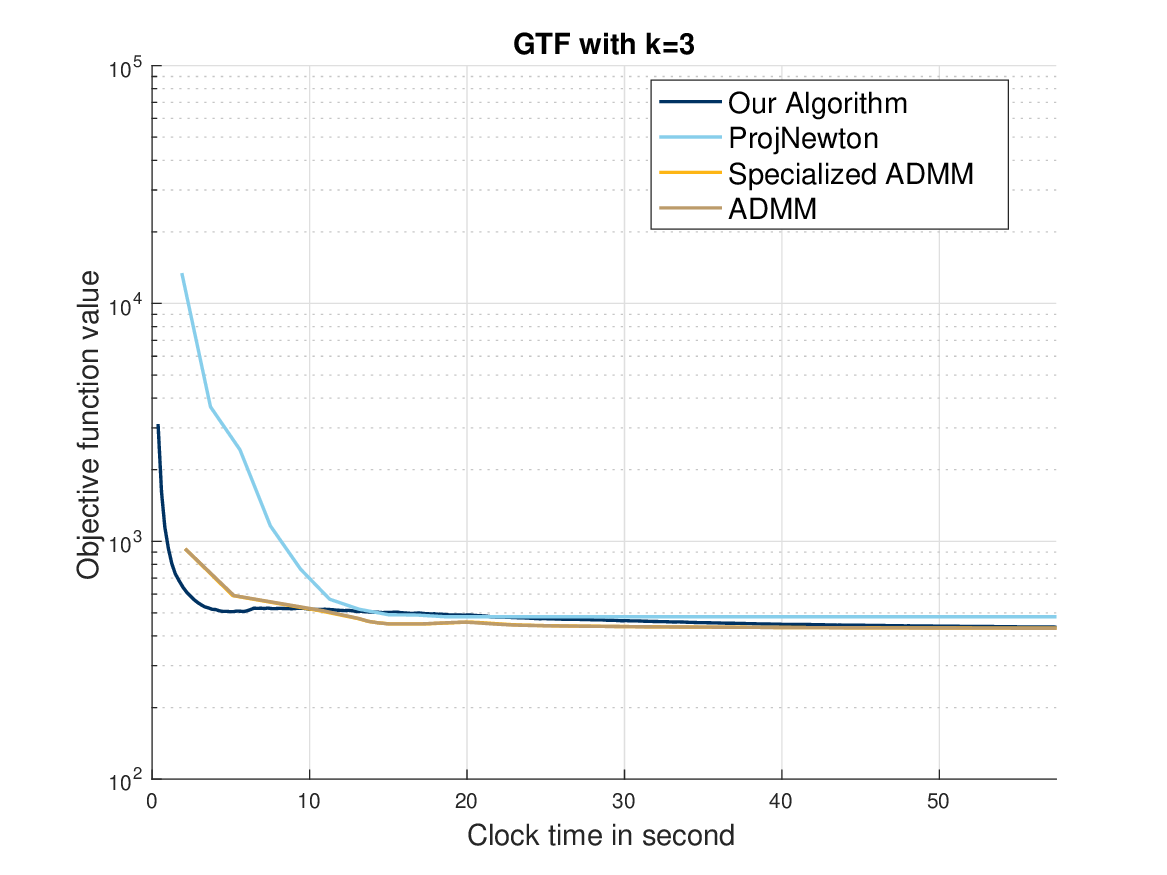}
\end{minipage} \\
\begin{minipage}[b]{.33\textwidth}
\includegraphics[width=60mm,height=45mm]{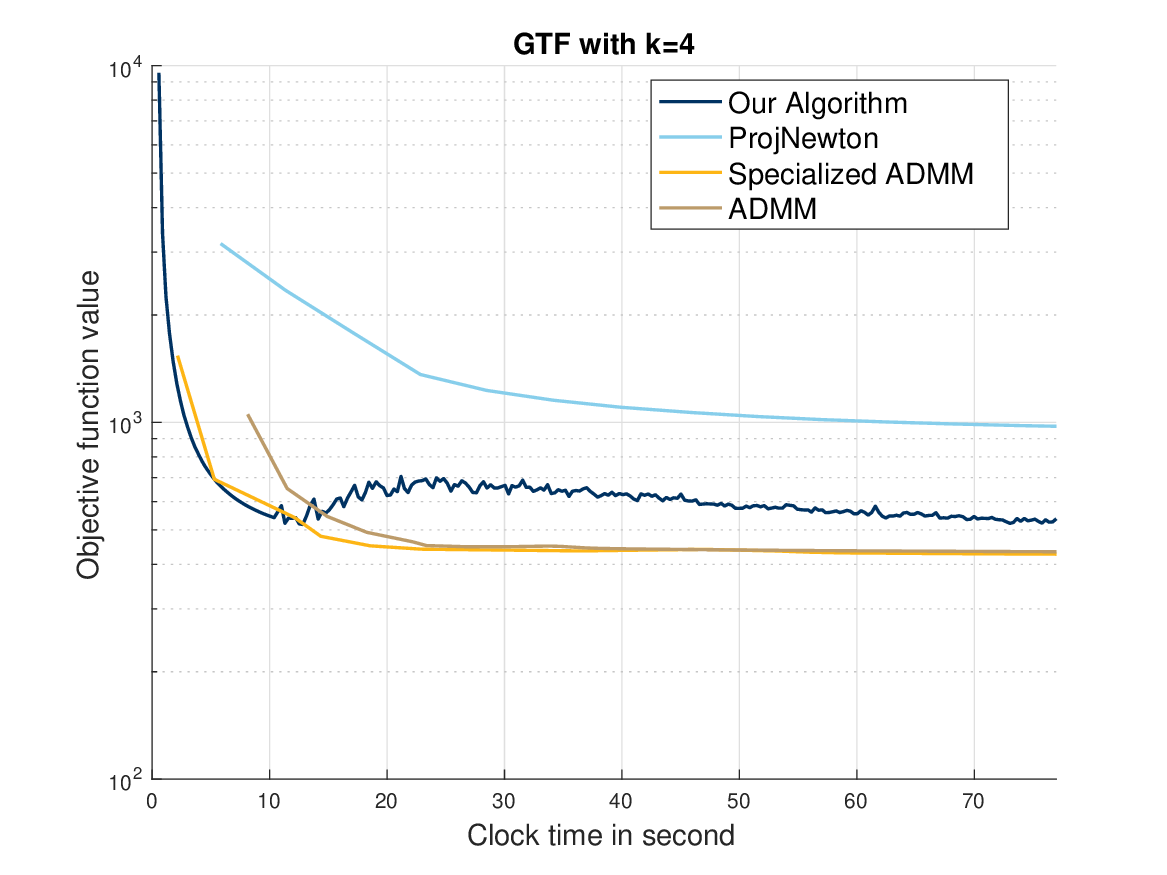}
\end{minipage}
\begin{minipage}[b]{.33\textwidth}
\includegraphics[width=60mm,height=45mm]{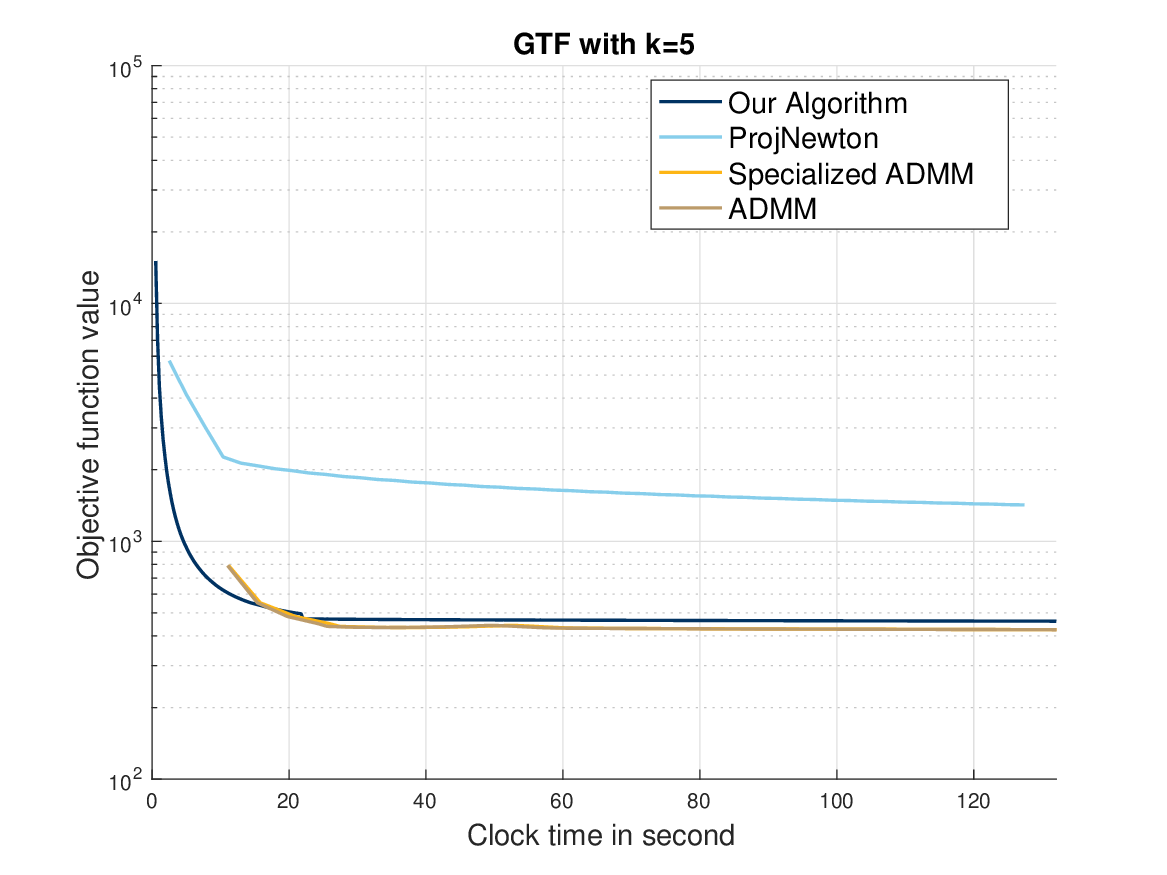}
\end{minipage}
\begin{minipage}[b]{.33\textwidth}
\includegraphics[width=60mm,height=45mm]{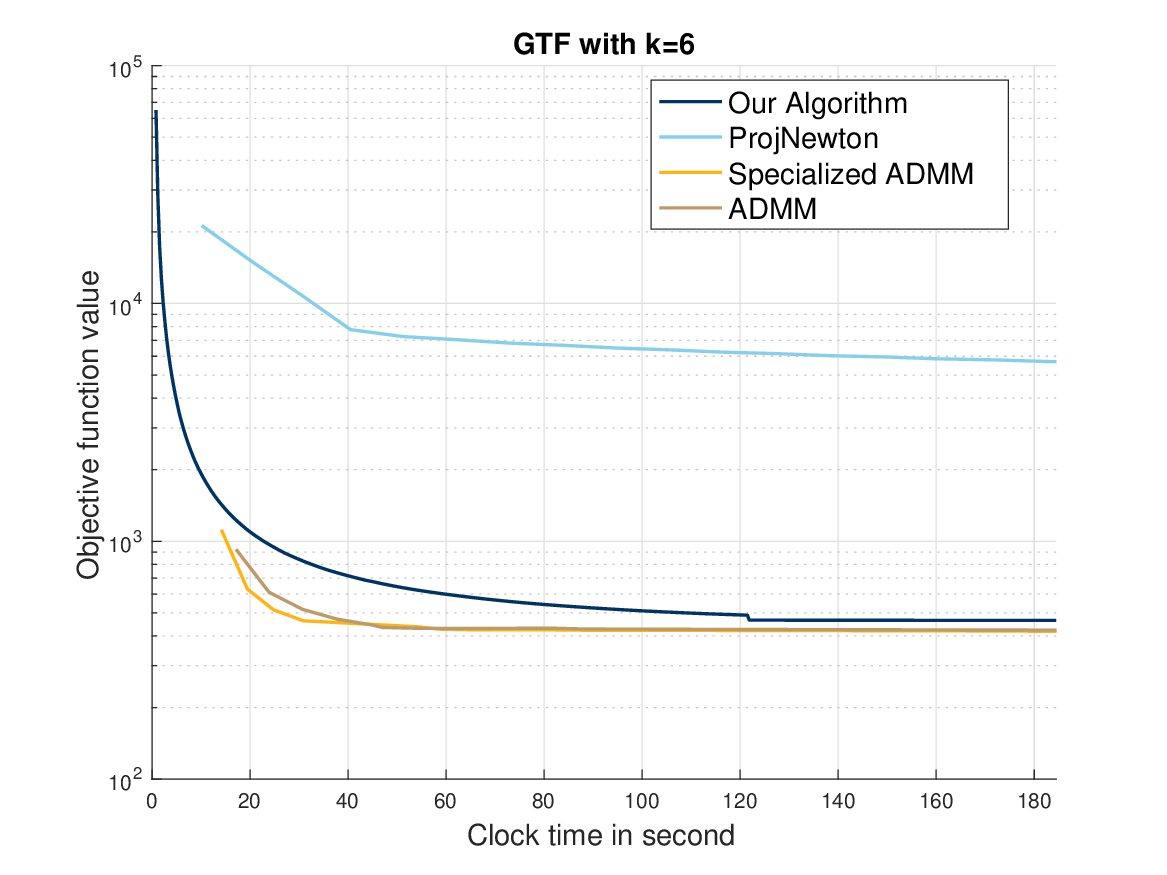}
\end{minipage}
\caption{\small{Comparison of Algorithm~\ref{Algorithm:GDGA-simplified}, ADMM, specialized ADMM, and projected Newton method on medium image.}} \label{fig:TF_medium}
\end{figure*}
\begin{figure*}[!t]
\begin{minipage}[b]{.33\textwidth}
\includegraphics[width=60mm,height=45mm]{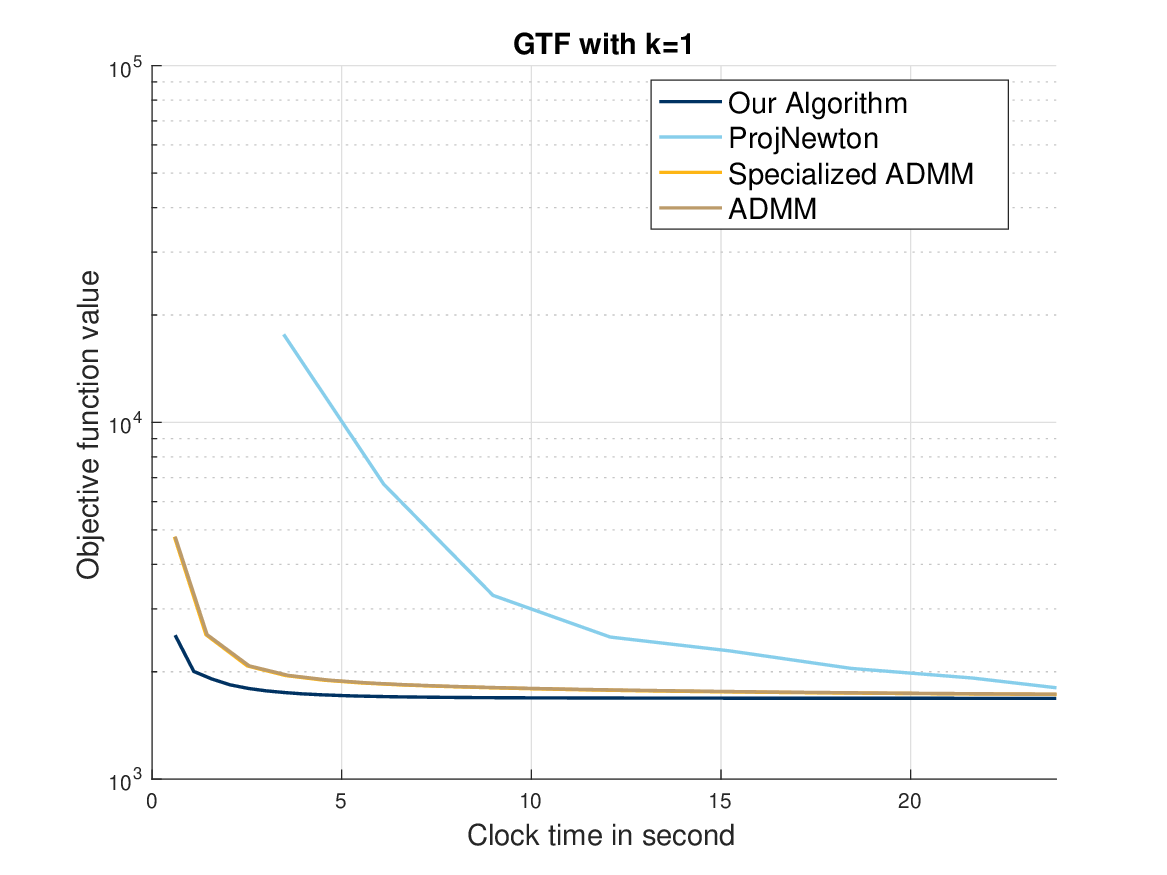}
\end{minipage}
\begin{minipage}[b]{.33\textwidth}
\includegraphics[width=60mm,height=45mm]{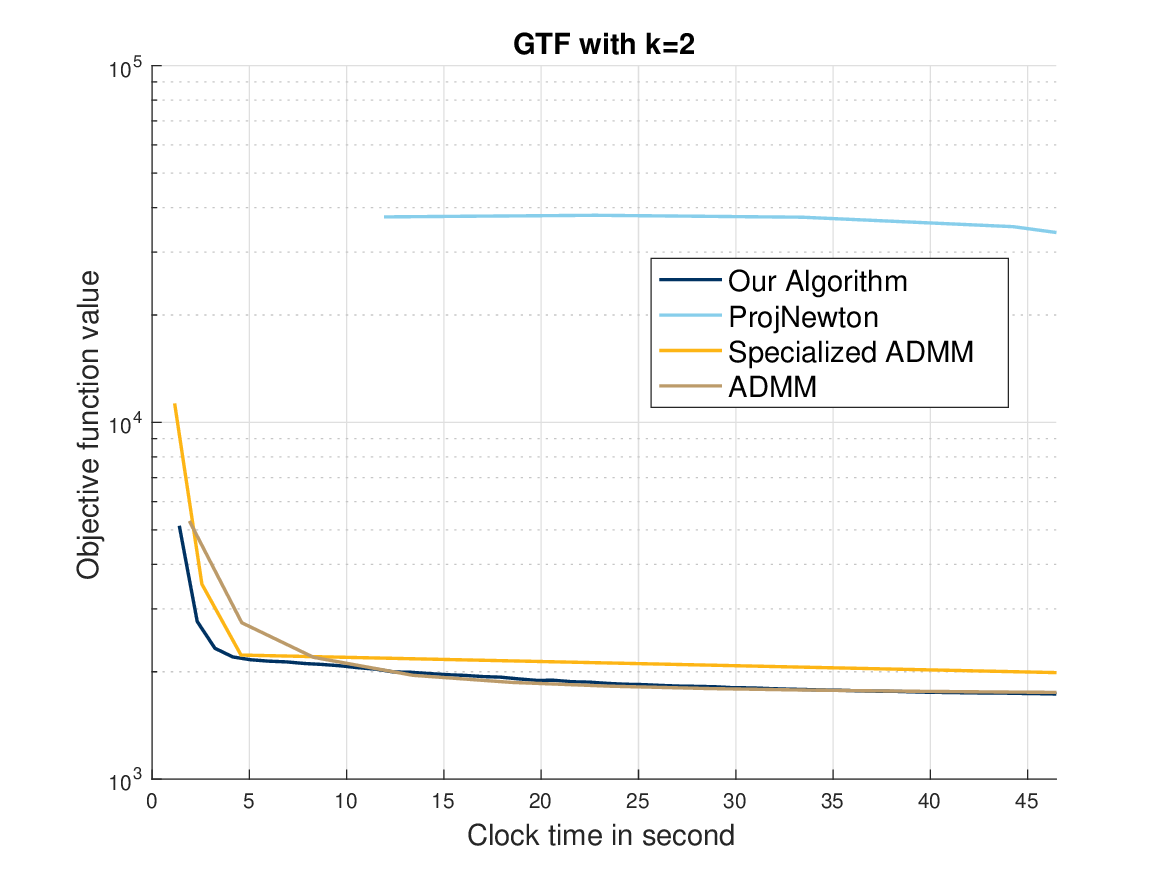}
\end{minipage}
\begin{minipage}[b]{.33\textwidth}
\includegraphics[width=60mm,height=45mm]{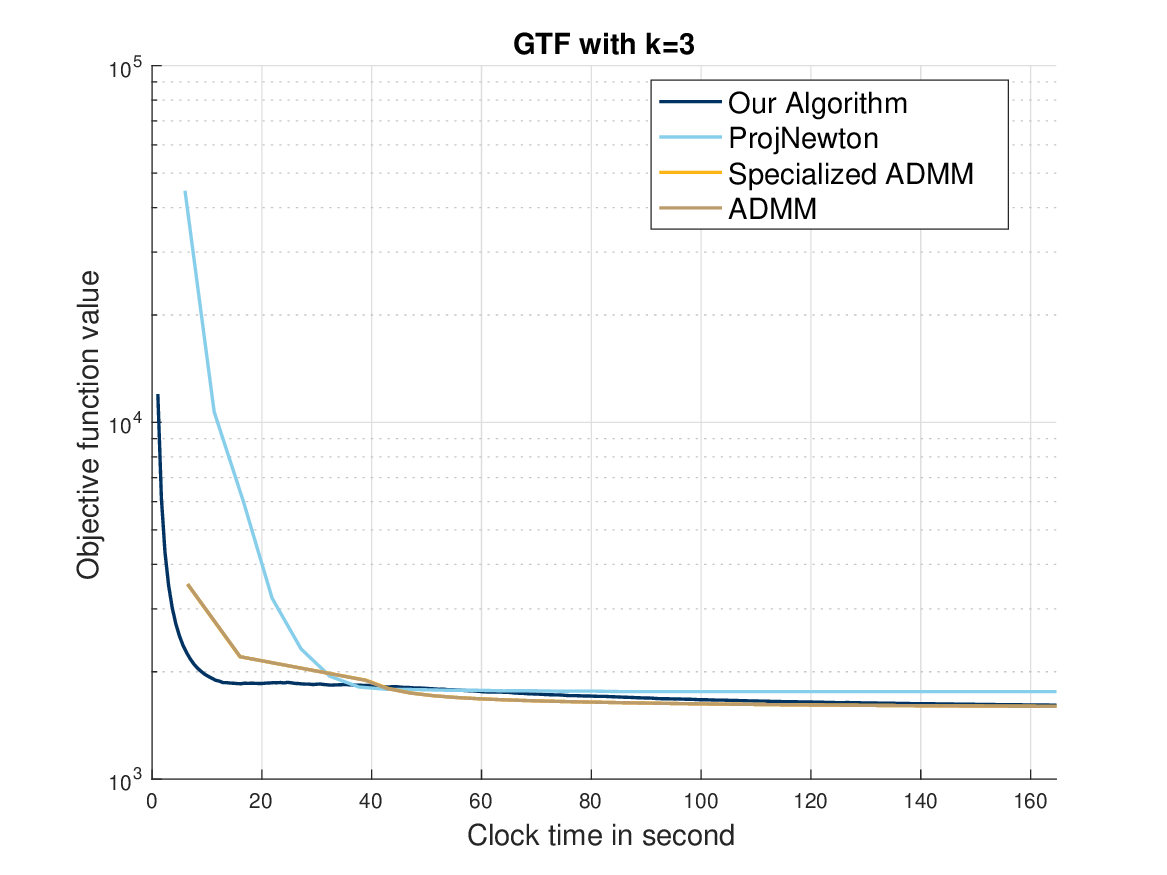}
\end{minipage} \\
\begin{minipage}[b]{.33\textwidth}
\includegraphics[width=60mm,height=45mm]{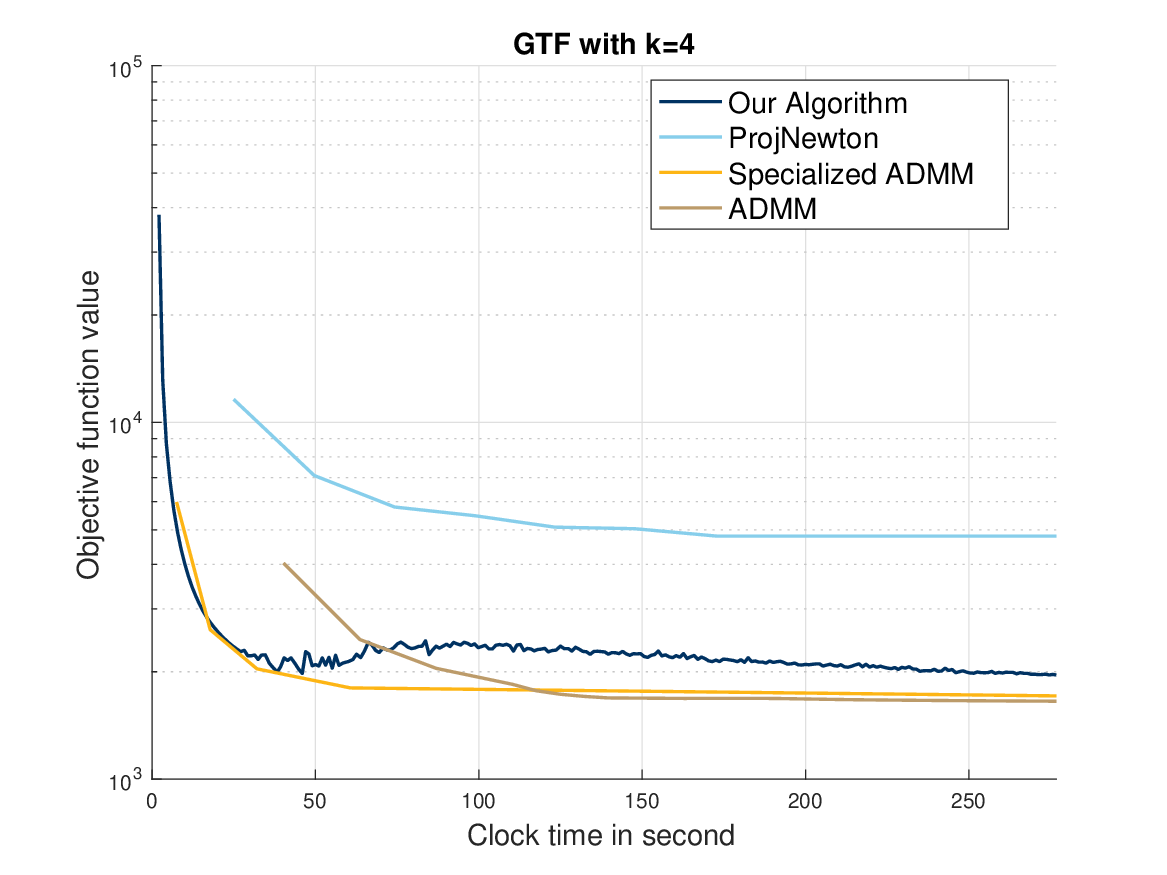}
\end{minipage}
\begin{minipage}[b]{.33\textwidth}
\includegraphics[width=60mm,height=45mm]{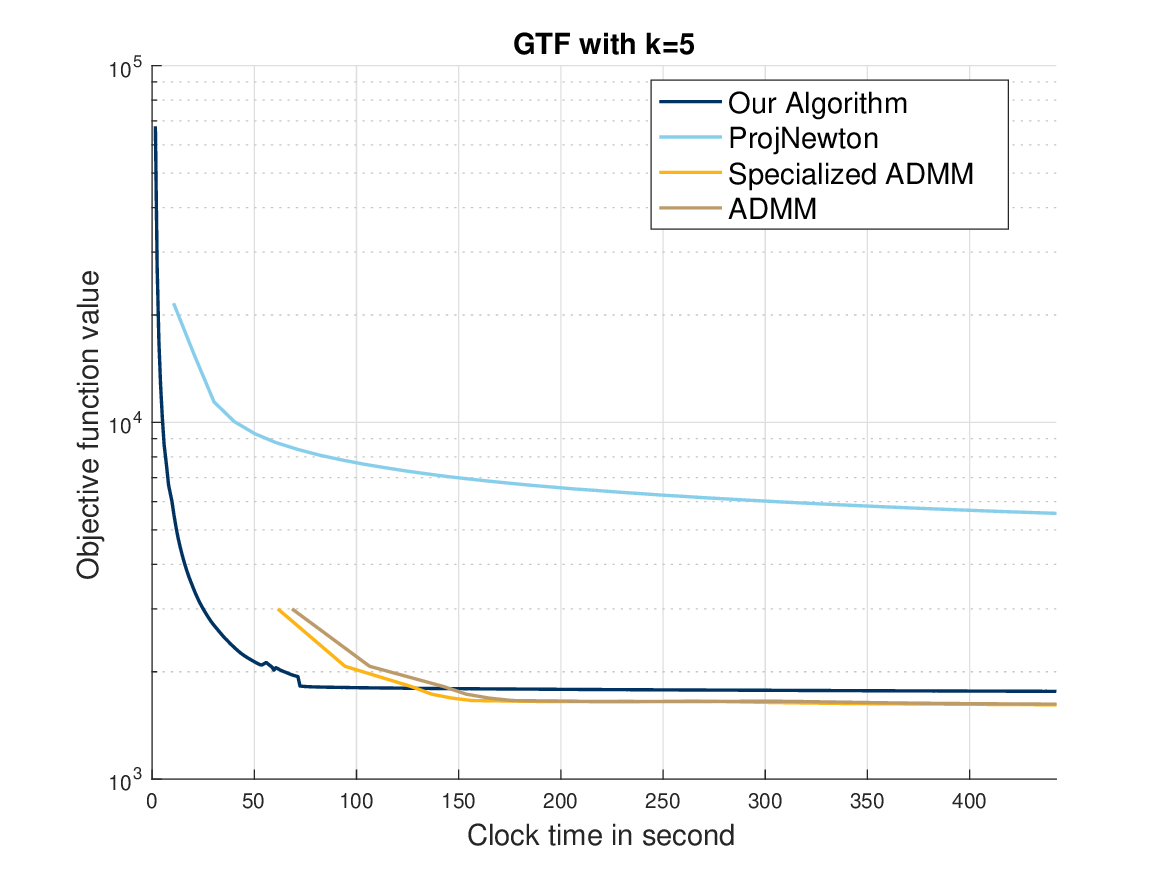}
\end{minipage}
\begin{minipage}[b]{.33\textwidth}
\includegraphics[width=60mm,height=45mm]{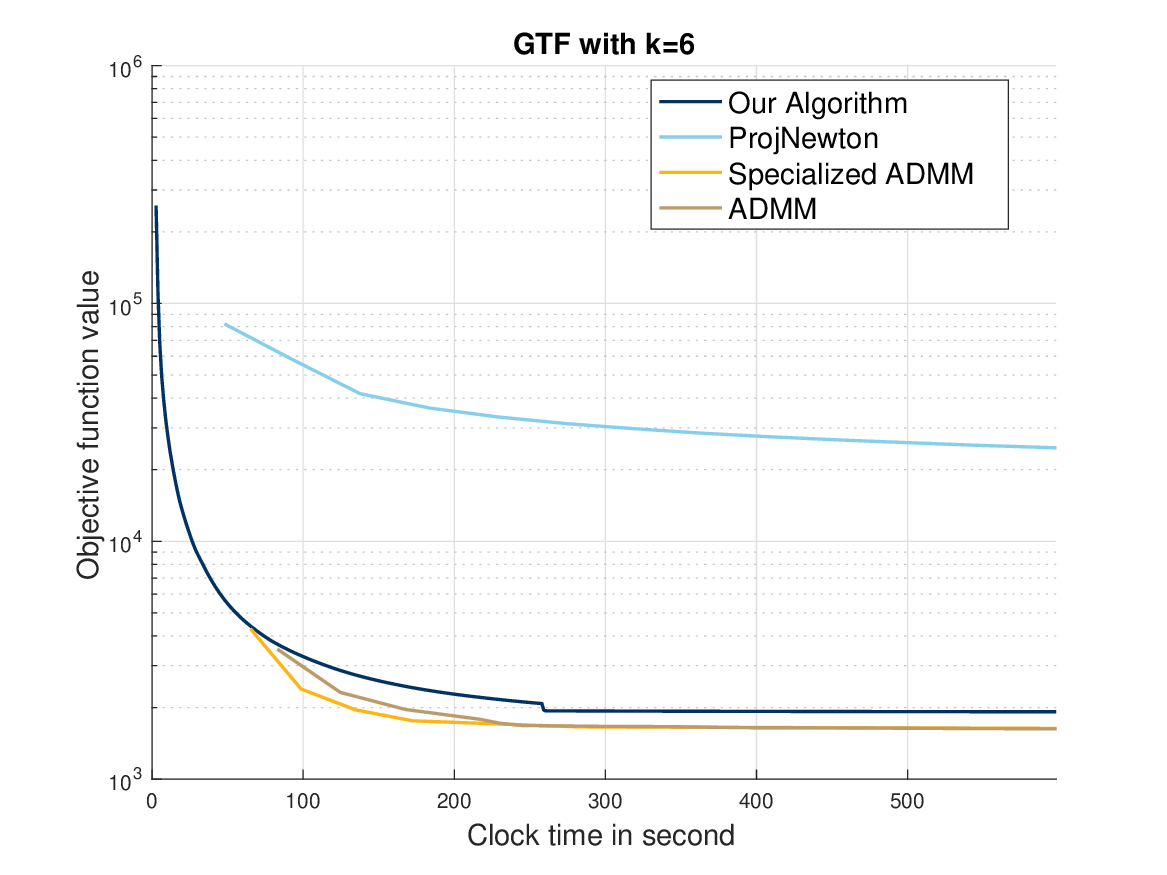}
\end{minipage}
\caption{\small{Comparison of Algorithm~\ref{Algorithm:GDGA-simplified}, ADMM, specialized ADMM, and projected Newton method on large image.}} \label{fig:TF_large}
\end{figure*}
\section{Additional Experimental Results}
\textbf{Setup.} We include ADMM, specialized ADMM~\citep{Ramdas-2016-Fast} and projected Newton method~\citep{Wang-2016-Trend} as the baseline approaches and consider three real images with various sizes: 128 by 128 pixels (small image), 256 by 256 pixels (medium image) and 512 by 512 pixels (large image)\footnote{These images can be found at: \url{http://sipi.usc.edu/database/database.php?volume=misc}}. All the algorithms are evaluated as the order $k$ varies in the discrete difference operator $D^{(k+1)}$ in which the evaluation metric is the objective function value. We choose $\lambda = 0.2$ in our experiment and consider the adaptive step-size $\eta_t > 0$ based on Barzilai-Borwein rule which is omitted due to space limit.

\end{document}